%% file: main.tex
\documentclass{article}

\usepackage{PRIMEarxiv}

\usepackage[utf8]{inputenc}
\usepackage[T1]{fontenc}
\usepackage[dvipsnames,table]{xcolor}
\usepackage{xspace}
\usepackage{graphicx}
\usepackage{amsmath}
\usepackage{amssymb}
\usepackage{amsfonts}
\usepackage{booktabs}
\usepackage{nicefrac}
\usepackage{microtype}
\usepackage[numbers,sort&compress]{natbib}
\usepackage[format=plain,labelformat=simple,labelsep=period,font=small,compatibility=false]{caption}
\usepackage[font=footnotesize,skip=3pt,subrefformat=parens]{subcaption}
\usepackage[hyphens]{url}
\usepackage[shortlabels,inline]{enumitem}

\input{preamble}

\definecolor{linkblue}{rgb}{0.21,0.49,0.74}
\usepackage[pagebackref,breaklinks,colorlinks,allcolors=linkblue]{hyperref}
\usepackage[capitalize]{cleveref}
\crefname{section}{Sec.}{Secs.}
\Crefname{section}{Section}{Sections}
\crefname{table}{Tab.}{Tabs.}
\Crefname{table}{Table}{Tables}

\makeatletter
\DeclareRobustCommand\onedot{\futurelet\@let@token\@onedot}
\def\@onedot{\ifx\@let@token.\else.\null\fi\xspace}
\def\eg{\emph{e.g}\onedot} 
\def\ie{\emph{i.e}\onedot} 
 
 \def\vs{\emph{vs}\onedot}

\makeatother

\title{ForeSay: Tell the Future in Words, Branch the Details in Noise for Precipitation Nowcasting}

\author{
  Xudong Ling$^{1*}$ \quad Wenjie Luo$^{1*}$ \quad Chaorong Li$^{2}$ \quad Tianxi Huang$^{3}$ \quad Tao He$^{1}$ \quad Guiduo Duan$^{1\dagger}$ \\[6pt]
  {\normalfont $^{1}$University of Electronic Science and Technology of China} \\
  {\normalfont $^{2}$Yibin University} \\
  {\normalfont $^{3}$Chengdu Textile College} \\[4pt]
  {\normalfont\small $^{*}$Equal contribution. \quad $^{\dagger}$Corresponding author.}
}

\begin{document}
\maketitle

\input{sec/0_abstract}


\input{sec/1_intro}

\input{sec/2_formatting}
\input{sec/3_method}

\input{sec/4_experiments}
\input{sec/5_ablation}

\input{sec/6_conclusion}

{
  \small
  \bibliographystyle{ieeenat_fullname}
  \bibliography{main}
}

\clearpage
\input{sec/X_appendix}

\end{document}

%% file: preamble.tex
\usepackage{algorithm}
\usepackage{algorithmicx}
\usepackage{algpseudocode}
\usepackage{listings}
\usepackage{subcaption}
\usepackage{float}  %
\usepackage{multirow}

\renewcommand{\topfraction}{0.85}
\renewcommand{\textfraction}{0.1}

\usepackage{colortbl}
\definecolor{oursrow}{RGB}{252,234,234}
\definecolor{secondblue}{RGB}{43,92,138}
\newcommand{\sbest}[1]{\underline{#1}}  %

\definecolor{branchteal}{RGB}{0,139,150}
\newcommand{\modline}[1]{\textcolor{branchteal}{#1}}

%% file: sec/0_abstract.tex
\begin{abstract}
Precipitation nowcasting involves two distinct scales of uncertainty: macroscopic storm evolution and fine-scale local convective variations. Current generative ensemble methods entangle both scales within independent random noise draws. This coupling causes two practical limitations: every ensemble member must execute the full sampling trajectory from scratch, so an $N$-member ensemble costs $N$ times a single forecast; moreover, divergent trajectories spatially misalign heavy-rain echoes across members, causing standard pixel-wise averaging to blur critical high-intensity extremes. We present \textbf{ForeSay}, a generative nowcasting framework that decouples macroscopic evolution prediction from local stochastic detail generation. Given historical radar frames, a vision--language model first predicts a textual scenario of future storm dynamics to condition a Rectified Flow generator. Guided by this shared macroscopic prior, ensemble members share early trunk sampling steps to establish the storm trajectory and branch with fresh noise only in later steps to explore local convective details. Because members then differ mainly in local details, self-anchored support-consensus decoding replaces pixel averaging with discrete support voting to preserve sharp heavy-rain cores. Across six datasets, ForeSay achieves the best CSI at the highest precipitation thresholds, with relative improvements over the strongest baseline on each dataset ranging from over $10\%$ to $84\%$. Our 8-member ensemble retains the forecasting skill and diversity of eight independently sampled runs while reducing computational cost from $8\times$ to $5.2\times$ that of a single forecast.
\end{abstract}

%% file: sec/1_intro.tex
\section{Introduction}
\label{sec:intro}
\begin{figure}[t]
  \centering
  \includegraphics[width=0.45\linewidth]{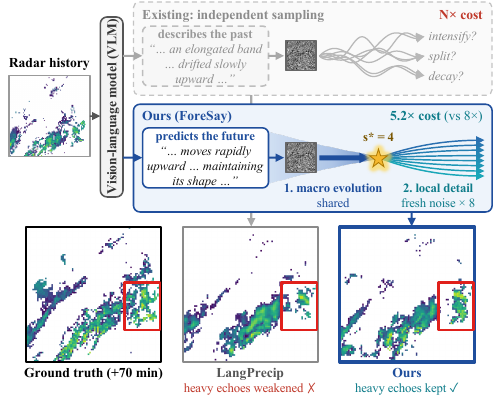}
  \caption{\textbf{Describing the past \vs predicting the future.} Conditioned only on the observed history, LangPrecip~\cite{ling2026langprecip} leaves the evolution to each member's noise (paths illustrative) and repeats every step. ForeSay conditions on the predicted evolution, so members share the early steps and then branch ($5.2\times$ \vs $8\times$ cost). Bottom: forecasts at $+70$\,min, with red boxes marking the same heavy-echo region.}
  \label{fig:concept}
\end{figure}

Precipitation nowcasting supports disaster mitigation, urban flood control and severe weather warning~\cite{ravuri2021skilful}, and is commonly formulated in computer vision as video prediction of radar echo sequences. Unlike natural video generation, which pursues overall visual realism, extreme precipitation is sparse in space: it covers only a tiny fraction of pixels yet decides whether a warning succeeds, so optimizing average fidelity tends to smooth these cores away. Driven by chaotic, nonlinear convection, future precipitation is also intrinsically uncertain at two levels: the \emph{macroscopic evolution} of the system, \ie where it moves, whether it intensifies or decays, and whether it splits or reorganizes; and the \emph{local detail} that remains uncertain once the evolution is fixed, such as the exact location, shape and intensity of convective cells.

Deterministic models~\cite{shi2015convolutional,wang2017predrnn,gao2022simvp,gao2022earthformer} regress toward the conditional mean and blur heavy echoes at long lead times. Generative models~\cite{voleti2022mcvd,ravuri2021skilful,gao2023prediff,yu2024diffcast,gong2024cascast} instead sample from the conditional distribution, preserving fine details, and are widely used to build ensemble nowcasts. Such ensembles, however, face two problems. First, an $N$-member ensemble costs $N$ times a single forecast. Second, members place heavy echoes at different locations, so pixel-wise averaging smooths them out and the aggregate loses the extremes. Both problems share one cause. Each member entangles the macroscopic evolution with the local detail in its own noise. Although this evolution can largely be inferred from the radar history, it is never separated out for members to share, so the early, high-noise steps that settle it are repeated across members, and the evolutions they settle diverge. LangPrecip~\cite{ling2026langprecip} introduced natural-language motion descriptions as an explicit condition, but its text summarizes past observations rather than forecasting how the system will evolve, so the future evolution is still left to noise. The question is thus how ensemble members can share a single forecast of the storm system's overall evolution, so that ensembles preserving local heavy-rain echoes can be generated efficiently.

We therefore decouple the prediction of the macroscopic evolution from the generation of local stochastic detail, and build ForeSay on this idea (\cref{fig:concept}). A vision--language model first predicts the future evolution in text, which conditions a Rectified Flow~\cite{liu2022flow,lipman2022flow} generator. Conditioned on this same text, all ensemble members share early trunk sampling steps along the forward trajectory and inject fresh noise only to explore local details. Differing mainly in local details, members are decoded by self-anchored consensus voting instead of pixel averaging, preserving sharp heavy echoes.

Our contributions are summarized as follows:
\begin{itemize}
\item \textbf{Shared forecast of macroscopic evolution:} Existing generative ensembles leave the storm system's evolution to each member's noise, or to text that only describes the past, with no explicit forecast of that evolution. We propose ForeSay, which uses a vision--language model to predict a textual scenario of the future evolution directly from radar history, giving all ensemble members a shared macroscopic prior.
\item \textbf{Efficient branched sampling and consensus decoding:} Independent ensemble sampling is costly, and pixel averaging smooths out heavy echoes. We propose training-free branched Rectified Flow sampling, in which members share early trunk steps and branch with fresh noise only in later steps, and self-anchored support-consensus decoding, which replaces pixel averaging with pixel-wise voting to avoid averaging blur. The branched ensemble matches the skill and diversity of 8 independent members at $5.2\times$ compute.
\item \textbf{Comprehensive benchmarks and validation:} On six datasets, ForeSay achieves the best CSI at the highest intensity threshold, improving over the strongest baseline by $11\%$--$84\%$ (\cref{tab:AvG,tab:radar-main}). Ablations, hyperparameter analyses, and qualitative results verify each component.
\end{itemize}

%% file: sec/2_formatting.tex
\section{Related work}
\label{sec:formatting}

\noindent\textbf{Precipitation nowcasting as video prediction.}
Deterministic nowcasters~\cite{shi2015convolutional,wang2017predrnn,gao2022simvp,gao2022earthformer,tan2023openstl,lin2025alphapre,song2026extreme,zhou2026sdir} regress toward the conditional mean and blur heavy echoes at long lead times. Following generative video models~\cite{denton2018stochastic,voleti2022mcvd,ho2022video,blattmann2023align}, adversarial~\cite{ravuri2021skilful}, diffusion~\cite{gao2023prediff,yu2024diffcast,gong2024cascast} and flow-based~\cite{fengperceptually,ribeiro2026flowcast,luo2026pdrf} nowcasters preserve sharp details and express uncertainty with multi-member ensembles, but each member entangles the macroscopic evolution and the local detail in one noise draw. Every member therefore recomputes the full trajectory, and when members diverge, averaging blurs the heavy echoes again. ForeSay predicts the evolution once as a shared text condition and leaves only the local detail to the noise.

\noindent\textbf{Language conditions for video and weather.}
\label{sec:rw-lang}
Text-conditioned video generators~\cite{singer2022make,lian2024llmgrounded,lin2024videodirectorgpt,du2024vlp} follow user prompts rather than physical dynamics, while meteorological language models mostly act as backbones~\cite{franch2025gptcast,she2024llmdiff}, foundation models~\cite{zhang2025weathergfm} or event predictors~\cite{li2025cllmate,tang2026meteorpred}. LangPrecip~\cite{ling2026langprecip} first conditioned a precipitation generator on motion text, yet its text only summarizes past observations and does not constrain future advection, intensity or reorganization. Omni-Weather~\cite{zhou2026omniweather} reasons via chain-of-thought in a unified backbone, improving LPIPS at the cost of CSI. ForeSay instead predicts the future evolution from radar history, checked against the ground truth (\cref{tab:scenario-acc}).

\noindent\textbf{Shared sampling and ensemble aggregation.}
\label{sec:rw-renoise}\label{sec:rw-pmm}
Existing trajectory-sharing and re-noising methods accelerate or edit single samples: DeepCache~\cite{ma2024deepcache} reuses features across adjacent steps, SAGE~\cite{sage2025shared} shares early steps across similar prompts, expert denoisers~\cite{balaji2022ediffi,wan2025wan22} split the noise stages across networks, and re-noising~\cite{meng2022sdedit,xu2023restart} serves editing or refinement. None is designed for ensemble forecasting, which needs members that share the macroscopic evolution and differ only in local details. Branched sampling shares a trunk under one predicted evolution and replaces only each member's noise endpoint. For aggregation, probability matching (PMM)~\cite{ebert2001pmm,clark2017lpm} maps member amplitudes onto the blurry ensemble-mean pattern, whereas support-consensus decoding avoids the mean field and corrects one anchor member by a pixel-wise vote (\cref{sec:agg-compare}).

%% file: sec/3_method.tex
\section{Method}
\label{sec:method}

  \begin{figure*}[!t]
    \centering
    \includegraphics[width=0.9\textwidth]{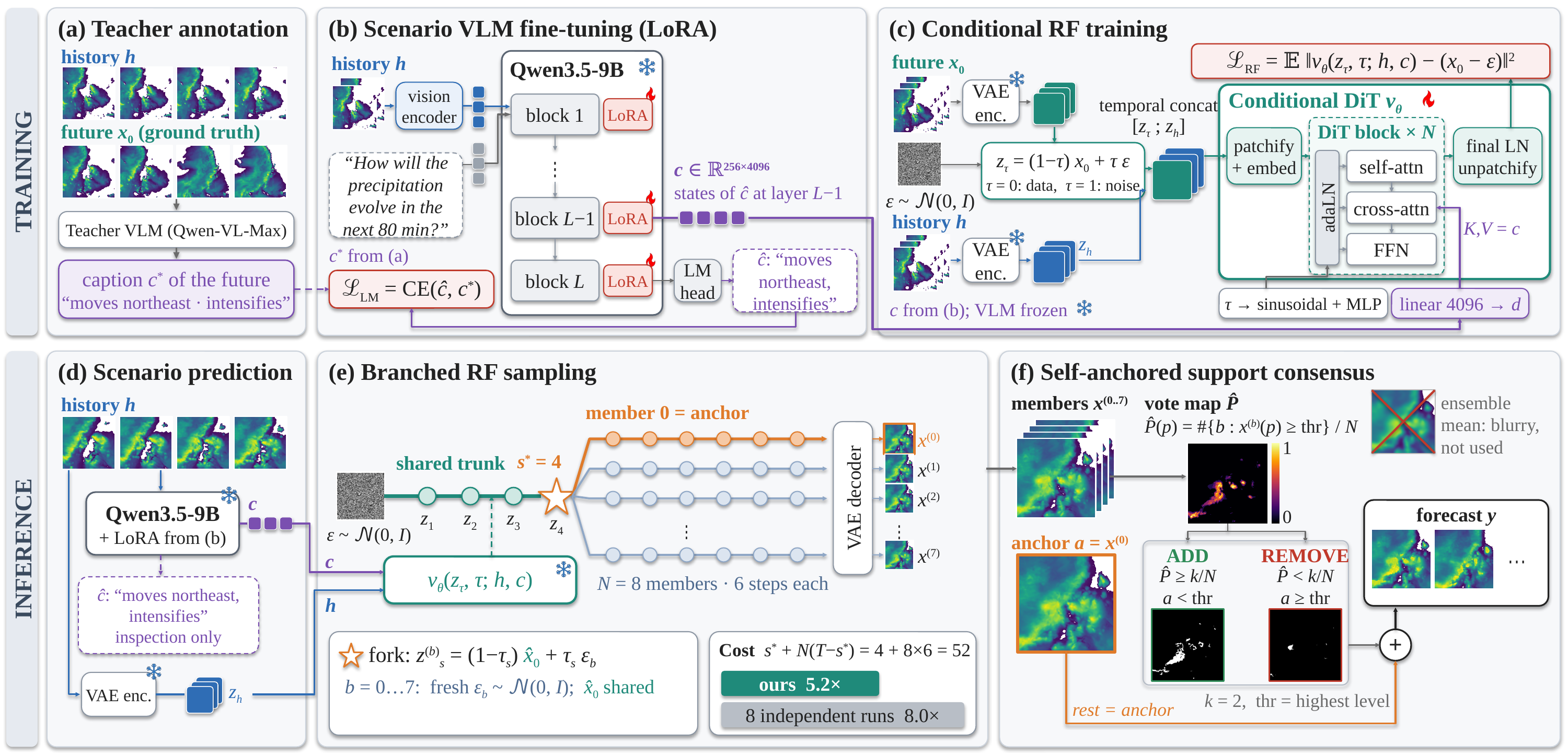}
\caption{\textbf{ForeSay overview.} \emph{Training} (top): (a) a teacher VLM that sees the true future writes the description $c^{*}$; (b) Qwen3.5-9B is LoRA-tuned to predict $c^{*}$ from the history $h$ alone, and $c$ is read from its penultimate layer; (c) with the VLM frozen, the DiT velocity field $v_\theta(z_\tau,\tau;h,c)$ is trained with Rectified Flow ($h$ concatenated along time, $c$ via cross-attention, $\tau$ via adaLN). \emph{Inference} (bottom): (d) the tuned VLM predicts $c$ ($\hat{c}$ for inspection only); (e) after a shared trunk of $s^{*}=4$ steps, $N=8$ members keep the shared $\hat{x}_0$ and differ only in the noise endpoint, costing $5.2\times$ instead of $8\times$; (f) the anchor's precipitation support is corrected by a $k$-of-$N$ vote.}
    \label{fig:start1}
\end{figure*}
A generative nowcaster draws each forecast from a random noise sample, and this sample entangles the macroscopic storm evolution with local convective details. In an ensemble, every member therefore iterates the full sampling chain from scratch to trace its own evolution path, so an $N$-member ensemble costs $N$ times a single forecast; members with divergent evolutions also misalign heavy echoes in space, which simple pixel averaging blurs away. ForeSay addresses these two problems with three coordinated components (\cref{fig:start1}): (i) \textbf{future scenario condition learning} (\cref{subsec:scenario}) uses a vision--language model (VLM) to predict the macroscopic evolution in text as a shared prior $c$; (ii) \textbf{branched ensemble sampling} (\cref{subsec:branch}) shares early trunk sampling steps across all members to cut compute; and (iii) \textbf{self-anchored support-consensus decoding} (\cref{subsec:decoding}) conducts pixel-wise consensus voting across members to avoid averaging blur.

\noindent\textbf{Problem setup.} Given $L$ observed radar frames $h$, nowcasting predicts the next $F$ frames $x$. A frozen VAE~\cite{rombach2022high,flux2} with encoder $\mathcal{E}$ and decoder $\mathcal{D}$ maps each $256\times256$ frame to a $32\times32\times32$ latent, so each frame $f$ of $x_0=\mathcal{E}(x)$ has $d_f=32768$ coordinates. The generator is a Rectified Flow~\cite{liu2022flow} in this latent space. Its velocity field $v_\theta(z_\tau,\tau;h,c)$ with parameters $\theta$ is evaluated at the latent state $z_\tau$ at time $\tau\in[0,1]$, is conditioned on the history latents $\mathcal{E}(h)$ and a text condition $c$ (\cref{subsec:scenario}), and transports Gaussian noise at $\tau=1$ to the data at $\tau=0$. Sampling takes $T$ Euler steps on a grid $1=\tau_0>\tau_1>\dots>\tau_T=0$, so the state after $s$ steps lies at time $\tau_s$, and $\mathcal{D}$ decodes the final state.

\subsection{Future Scenario Condition Learning}
\label{subsec:scenario}
To give all members a shared forecast of the macroscopic storm evolution, the future scenario condition learning module turns radar history into a textual scenario condition $c$. Specifically, a vision--language model (VLM) reads the observed sequence $h$ to predict a text scenario describing how radar echoes will evolve, covering advection direction, intensity trends and structural reorganization. The hidden states of this text condition the Rectified Flow generator. Branched ensemble sampling (\cref{subsec:branch}) relies on $c$ to share the early sampling steps across members, and support-consensus decoding (\cref{subsec:decoding}) relies on the resulting members differing mainly locally. Unlike prior approaches such as LangPrecip~\cite{ling2026langprecip} that only describe observed historical frames and leave future evolution unconstrained, ForeSay directly forecasts the future evolution itself, providing explicit guidance on the macroscopic evolution for downstream generation.

\noindent\textbf{Training.} Training has three stages (top row of Fig.~\ref{fig:start1}). (i) A teacher VLM (Qwen-VL-Max) sees $h$ and the true future $x$ and writes a reference description $c^{*}$ offline, used only as supervision. (ii) A student VLM, Qwen3.5-9B~\cite{qwen35} with LoRA~\cite{hu2022lora} on all linear layers, learns to generate $c^{*}$ from $h$ alone by minimizing the token-level cross-entropy
\begin{equation}
\mathcal{L}_{\mathrm{LM}}(\phi)=-\sum_{j=1}^{|c^{*}|}\log p_\phi\bigl(c^{*}_j\mid c^{*}_{<j},h\bigr),
\label{eq:ce}
\end{equation}
where $|c^{*}|$ is the number of tokens in $c^{*}$, $c^{*}_j$ is its $j$-th token, $c^{*}_{<j}$ are the tokens before it, $p_\phi$ is the next-token distribution of the student and $\phi$ are its LoRA parameters. From the tokens $\hat{c}$ that the student generates autoregressively given $h$, we take the penultimate-layer hidden states, padded to $256$ tokens, as $c\in\mathbb{R}^{256\times4096}$. (iii) With the weights of the student VLM and the VAE frozen, only the parameters $\theta$ of the DiT~\cite{peebles2023scalable} velocity field are trained, with the Rectified Flow loss
\begin{equation}
\mathcal{L}_{\mathrm{RF}}(\theta)=\mathbb{E}_{x_0,\varepsilon,\tau}\bigl\lVert v_\theta(z_\tau,\tau;h,c)-(x_0-\varepsilon)\bigr\rVert^{2},
\label{eq:rf}
\end{equation}
where $x_0=\mathcal{E}(x)$ is the latent of the true future, $\varepsilon\sim\mathcal{N}(0,I)$ is Gaussian noise of the same shape, $z_\tau=(1-\tau)x_0+\tau\varepsilon$ is the point at time $\tau$ on the straight path from $x_0$ to $\varepsilon$, and $c$ is the condition the student predicts for $h$, as at inference. The target $x_0-\varepsilon$ is the constant velocity of this path toward the data. $\mathcal{E}(h)$ is concatenated with $z_\tau$ along the time axis, $c$ enters through cross-attention and $\tau$ through adaLN. The generator never sees the teacher's $c^{*}$, and the text $\hat{c}$ serves only for inspection.

\subsection{Branched Ensemble Sampling}
\label{subsec:branch}
\begin{figure}[t]
  \centering
  \includegraphics[width=0.62\linewidth]{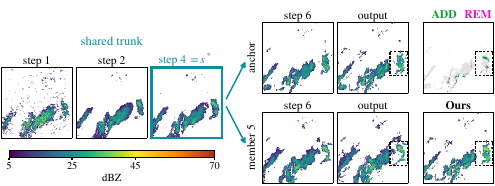}
  \caption{\textbf{Branched sampling on the Swedish case of Fig.~\ref{fig:qualitative}} ($+70$~min, $s^*=4$, $N=8$). Each panel is the endpoint estimate $\hat{x}_0$ at that step. The trunk fixes the rain band early; after branching, the members differ in the location and intensity of the heavy cores (dashed box), which the $k=2$ vote on members $\geq35$~dBZ ($\approx6.3$~mm/h) resolves by adding (ADD) or removing (REM) anchor pixels (right column). All eight members: App.~Fig.~\ref{fig:members-all}.}
  \label{fig:members}
\end{figure}

Given the evolution condition $c$, drawing $N$ members by independent sampling would cost $N$ times a single forecast. However, the early high-noise steps settle the coarse structure, as the trunk in the top row of Fig.~\ref{fig:members} shows. Under a shared scenario of the macroscopic evolution, independent members repeat nearly the same computation in these steps. To reduce this redundancy, we propose training-free \textbf{branched ensemble sampling}. Rather than rerunning the full trajectory, all members share the early steps as a single ``trunk'' to settle the macroscopic evolution, and then branch at step $s$ by injecting fresh random noise to explore local convective details independently. This strategy lowers the computational cost of an 8-member ensemble from $8\times$ to $5.2\times$ that of a single forecast, and the resulting members differ mainly in local details, as the subsequent consensus decoding requires (\cref{subsec:decoding}). Below, we first formalize the branching operator, and then address its two critical design choices: which evolution the trunk should convey, and at which step $s$ branching should occur.

\noindent\textbf{Branching operator.} Let $z$ be the trunk state after $s$ shared steps, $\tau_s \in (0, 1)$ the corresponding noise level (on the decreasing sampling grid $1=\tau_0>\tau_1>\dots>\tau_T=0$, where $\tau=1$ denotes pure noise and $\tau=0$ data), and $v$ the velocity already computed at the last trunk step. The estimates of the data endpoint $x_0$ and the noise endpoint $\varepsilon$ of the path through $z$, $\hat{x}_0=z+\tau_s v$ and $\hat{\varepsilon}=z-(1-\tau_s)v$, satisfy $z=(1-\tau_s)\hat{x}_0+\tau_s\hat{\varepsilon}$ exactly. Each of the $N$ members $b=0,\dots,N-1$ keeps $\hat{x}_0$ and mixes $\hat{\varepsilon}$ with fresh noise $\varepsilon_b\sim\mathcal{N}(0,I)$ at strength $\lambda\in[0,1]$,
\begin{equation}
z_b=(1-\tau_s)\hat{x}_0+\tau_s\bigl(\sqrt{1-\lambda^2}\,\hat{\varepsilon}+\lambda\,\varepsilon_b\bigr),
\label{eq:fork}
\end{equation}
where the two weights keep the noise term at unit variance. With $\lambda=1$, used throughout (\cref{subsec:lambda-main}),
\begin{equation}
z_b=(1-\tau_s)\hat{x}_0+\tau_s\varepsilon_b=z+\tau_s(\varepsilon_b-\hat{\varepsilon}).
\label{eq:branch}
\end{equation}
The branching operator replaces the noise component of the trunk and leaves its data component unchanged. Every member thus starts from the same $\hat{x}_0$ mixed with its own noise. The members share $\hat{x}_0$, finish the remaining $T-s$ steps independently, and are decoded by $\mathcal{D}$ into $x^{(b)}$. One ensemble costs $s+N(T-s)$ network evaluations, $52$ for $s=4$, $N=8$, $T=10$, \ie $5.2\times$ a single forecast against $8\times$ for $8$ independent samples (ens8). The two design choices appear in Eq.~\eqref{eq:branch} as the shared $\hat{x}_0$, setting the common evolution, and the weight $\tau_s$, setting the fresh noise level.

\begin{figure}[t]
  \centering
  \includegraphics[width=0.68\linewidth]{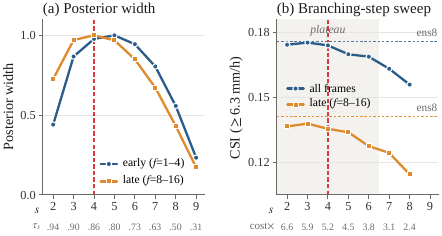}
  \caption{Selection of the branching step $s$ on the held-out set. (a) Peak-normalized posterior width of early and late frames. Below the axis, $\tau_s$ is the noise level of the sampler at step $s$ ($1$ is pure noise, $0$ is data). (b) Highest-threshold CSI when branching at step $s$ ($N=8$, $\lambda=1$, $k=2$), with the cost below the axis and the plateau shaded. Dashed lines: ens8 ($8.0\times$). Vertical line: $s^*=4$. Numbers in App.~Tab.~\ref{tab:s-sweep}.}
  \label{fig:posterior-width}
\end{figure}

\noindent\textbf{Trunk evolution guidance.} Since every member keeps the same $\hat{x}_0$, the whole ensemble inherits the evolution that the trunk has settled by step $s$. Without a future condition, this evolution comes from the noise drawn at the start of the trunk, and all members inherit one random choice. Under the predicted scenario $c$ of \cref{subsec:scenario}, the trunk is guided by the predicted evolution, and the members spend their fresh noise on local details. \Cref{subsec:abl-cond} ablates this effect.

\noindent\textbf{Optimal branching timing.}\label{sec:divergence-probe} With the evolution guided by $c$, the step $s$ sets the balance in Eq.~\eqref{eq:branch} between the shared $\hat{x}_0$, weighted by $1-\tau_s$, and the fresh noise, weighted by $\tau_s$. If the members branch after only a few shared steps ($s$ small, $\tau_s\to1$), each member restarts from nearly pure noise and stays close to an independent sample. Skill is then preserved, but the trunk saves little compute. If they branch after many shared steps ($s$ close to $T$, $\tau_s\to0$), $\hat{x}_0$ has already fixed local details that the few remaining steps cannot change. The members then collapse toward copies of one forecast, and the vote has little to act on. Branching too early thus saves little compute, whereas branching too late costs both diversity and skill, so skill stays on a plateau for early-to-mid $s$ and drops sharply once local details freeze (App.~\ref{app:operator}).

The best branching step is therefore the latest one at which the macroscopic evolution is settled while local details remain unconstrained. Rather than running full ensembles at every candidate step against ground truth, we locate it without labels from the unresolved posterior variance of each frame along a single sampling trajectory. Restricting the flow-matching Tweedie identity $\mathrm{Cov}(x_0\mid z_\tau)=\frac{\tau^2}{1-\tau}[I+\tau J_v]$, where $J_v=\partial v/\partial z$~\cite{xing2026divergence}, to frame $f$ yields its posterior width:
\begin{equation}
U_f(\tau)=\frac{\tau^2}{1-\tau}\left[d_f+\tau\,\mathrm{div}_f\, v\right],
\label{eq:uf}
\end{equation}
where $\mathrm{div}_f\,v$ is the trace of the frame-$f$ block of $J_v$. Since $U_f(\tau_s)$ quantifies the remaining stochastic exploration capacity, branching beyond the peak of $U_f$ rapidly reduces member diversity. We therefore branch where the mean width of the late frames ($f=8$--$16$) peaks:
\begin{equation}
s^{*}=\operatorname*{arg\,max}_{s}\ \bar{U}_{\mathrm{late}}(\tau_s).
\label{eq:sstar}
\end{equation}

\noindent\textbf{Posterior-width probe.} We estimate $\mathrm{div}_f\,v$ in Eq.~\eqref{eq:uf} with a Hutchinson trace estimator using a Rademacher probe on frame $f$, which costs one Jacobian--vector product (JVP). Locating $s^*$ thus takes only $F=16$ JVPs per step along a single trajectory, with neither ground truth nor branched members (App.~\ref{app:probe}). On the Swedish held-out set (\cref{subsec:setup}), this posterior width peaks at $s^*=4$ and falls to $0.43$ of its peak by $s=8$ (Fig.~\ref{fig:posterior-width}a). A supervised sweep over all $s$ (Fig.~\ref{fig:posterior-width}b) confirms the choice: CSI stays close to ens8 for $s\in[2,6]$ and drops from $s=7$, and $s^*=4$ captures the full 8-member gain over all frames ($94\%$ on frames $8$--$16$) at $5.2\times$ compute. The cheaper $s=6$ ($3.8\times$) suits compute-limited deployments (App.~\ref{app:s-sweep-stats}).

\begin{table*}[t]
  \centering
  \caption{Comparison on the four radar extrapolation benchmarks (official test sets, sample counts in parentheses, $128\times128$ protocol). CSI is given at the three highest thresholds of each benchmark (SEVIR in VIL $0$--$255$, the others in dBZ). $\uparrow$/$\downarrow$: higher/lower is better. Best in bold, second best underlined. Ours is the decoded ensemble with $\mathrm{thr}$ at the highest tier. CSI-M and CSI at all thresholds are in App.~Tab.~\ref{tab:app-csi}.}
  \label{tab:radar-main}
  \scriptsize
  \setlength{\tabcolsep}{1.1pt}
  \resizebox{\textwidth}{!}{%
  \begin{tabular}{l|cccccc|cccccc|cccccc|cccccc}
    \hline
     & \multicolumn{6}{c|}{\textbf{SEVIR} ($8096$)} & \multicolumn{6}{c|}{\textbf{MeteoNet} ($1310$)} & \multicolumn{6}{c|}{\textbf{Shanghai} ($526$)} & \multicolumn{6}{c}{\textbf{CIKM} ($4000$)} \\
     & \multicolumn{3}{c}{CSI$\uparrow$} & HSS$\uparrow$ & SSIM$\uparrow$ & LPIPS$\downarrow$ & \multicolumn{3}{c}{CSI$\uparrow$} & HSS$\uparrow$ & SSIM$\uparrow$ & LPIPS$\downarrow$ & \multicolumn{3}{c}{CSI$\uparrow$} & HSS$\uparrow$ & SSIM$\uparrow$ & LPIPS$\downarrow$ & \multicolumn{3}{c}{CSI$\uparrow$} & HSS$\uparrow$ & SSIM$\uparrow$ & LPIPS$\downarrow$ \\
    Method & 160 & 181 & 219 & & &  & 18 & 24 & 32 & & &  & 30 & 35 & 40 & & &  & 30 & 35 & 40 \\
    \hline
    MIMO-VP~\cite{ning2023mimo} & .1492 & .1102 & .0280 & .3792 & .6822 & .3576 & .3838 & .2469 & .0505 & .4016 & .8212 & .2732 & .3722 & .2586 & .1171 & .4332 & .7886 & .2532 & .1933 & .1351 & .0752 & .3198 & .6624 & .3853 \\
    SwinLSTM~\cite{tang2023swinlstm} & .1585 & .1188 & .0373 & .3892 & .6775 & .3309 & .4558 & .3631 & .1762 & .5129 & .8203 & .2404 & .4409 & .3460 & .2141 & .5212 & .7708 & .2831 & \sbest{.2990} & .2002 & .0799 & .3885 & .6058 & .3485 \\
    DiffCast~\cite{yu2024diffcast} & .1740 & .1435 & .0750 & .4017 & .6536 & .1983 & .4427 & .3496 & .1767 & .5004 & .8148 & .2364 & .4072 & .3254 & .2286 & .4986 & .7648 & \sbest{.1658} & .2792 & .2161 & .1261 & .4002 & .6676 & .3091 \\
    AlphaPre~\cite{lin2025alphapre} & .1754 & .1332 & .0535 & .4100 & .6818 & .3520 & .4315 & .3461 & .1807 & .4936 & .8065 & .2336 & .4514 & .3591 & .2287 & .5267 & .7463 & .3195 & .2604 & .1810 & .1070 & .3674 & .6359 & .2936 \\
    PercpCast~\cite{fengperceptually} & .2203 & .1929 & .1022 & .4538 & .6852 & \sbest{.1884} & .4737 & .3933 & .2207 & .5436 & .8266 & \sbest{.1238} & .4459 & .3628 & .2577 & .5404 & .7942 & .1918 & .2766 & .2223 & .1553 & .4107 & .6432 & \textbf{.2174} \\
    exPreCast~\cite{song2026extreme} & .2133 & .1649 & .0716 & .4355 & .6388 & .3237 & .4704 & .3833 & .2090 & .5346 & .8051 & .2375 & .4399 & .3471 & .2323 & .5250 & .7144 & .3224 & .2729 & .1873 & .0886 & .3733 & .5598 & .3794 \\
    FlowCast~\cite{ribeiro2026flowcast} & .2149 & .1661 & .0638 & .4454 & \sbest{.7254} & .2631 & .4705 & .3715 & .1803 & .5177 & .8345 & .1856 & .4815 & .3927 & .2622 & .5642 & .8003 & .2474 & .2954 & .2193 & .1324 & .4093 & .6652 & .3190 \\
    SDIR~\cite{zhou2026sdir} & .2201 & .1828 & .0934 & .4578 & \textbf{.7408} & .2869 & .5037 & .4220 & .2349 & .5665 & \sbest{.8423} & .1843 & \sbest{.4919} & .4040 & .2797 & .5768 & \sbest{.8213} & .2097 & .2988 & .2210 & .1524 & .4193 & \textbf{.6911} & .3052 \\
    PDRF~\cite{luo2026pdrf} & \sbest{.2408} & \sbest{.2110} & \sbest{.1199} & \sbest{.4678} & .6870 & .2068 & \sbest{.5053} & \sbest{.4285} & \sbest{.2679} & \sbest{.5746} & .8385 & .1408 & .4856 & \sbest{.4066} & \sbest{.3032} & \sbest{.5772} & .7938 & .1827 & .2938 & \sbest{.2381} & \sbest{.1720} & \sbest{.4312} & .6619 & .3161 \\
    \rowcolor{oursrow}Ours & \textbf{.2477} & \textbf{.2173} & \textbf{.1458} & \textbf{.4800} & .6978 & \textbf{.1670} & \textbf{.5058} & \textbf{.4369} & \textbf{.2971} & \textbf{.5860} & \textbf{.8451} & \textbf{.1000} & \textbf{.5166} & \textbf{.4369} & \textbf{.3387} & \textbf{.6098} & \textbf{.8222} & \textbf{.1302} & \textbf{.3172} & \textbf{.2598} & \textbf{.1941} & \textbf{.4546} & \sbest{.6754} & \sbest{.2691} \\
    \hline
  \end{tabular}%
  }
\end{table*}

\begin{figure*}[!t]
  \centering
  \includegraphics[width=\textwidth]{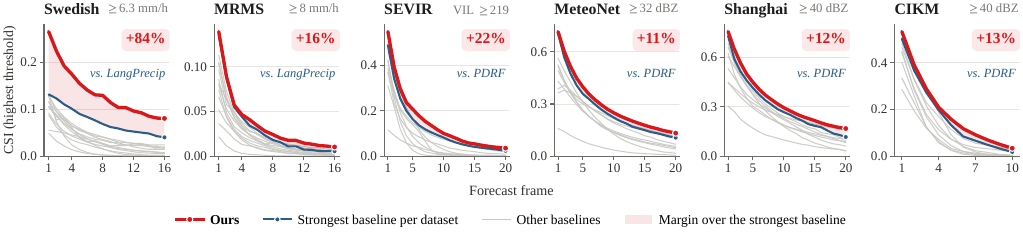}
  \caption{\textbf{Highest-threshold CSI versus forecast frame on all six datasets.} Ours is the decoded ensemble ($5.2\times$). The strongest baseline of each dataset is drawn in blue and named in the panel, with our relative gain in mean CSI (Tabs.~\ref{tab:radar-main} and~\ref{tab:AvG}). The shaded band is the margin between the two, and grey lines are the remaining baselines. Curves for every metric are in App.~Figs.~\ref{fig:sw_temporal}, \ref{fig:mrms_temporal} and~\ref{fig:lt-sevir}--\ref{fig:lt-cikm}.}
  \label{fig:leadtime}
\end{figure*}

\subsection{Self-Anchored Support-Consensus Decoding}
\label{subsec:decoding}
Averaging the $N$ members suppresses variance but smooths out their spatially misaligned heavy echoes, reintroducing the blur of deterministic models (\cref{sec:agg-compare}). To avoid this averaging blur, we propose \textbf{self-anchored support-consensus decoding}, which uses no pixel averaging. It takes one complete $T$-step member as the anchor, at no extra sampling cost, and adds or removes heavy precipitation by pixel-wise support voting across all members. Because all members share the scenario condition $c$ (\cref{subsec:scenario}) and branch from a shared trunk (\cref{subsec:branch}), their convective cores are already well aligned in space and differ primarily in local extent and intensity (Fig.~\ref{fig:members}). Resolving such local disagreements via consensus voting harnesses the diversity benefits of the ensemble while preserving the peak intensity and sharp boundaries of extreme echoes.

Let $a=x^{(0)}$ denote the anchor. For a pixel $p$ of any forecast frame and a given heavy-precipitation intensity threshold $\mathrm{thr}$, let $\hat{P}(p)=\frac{1}{N}\#\{b: x^{(b)}(p)\geq\mathrm{thr}\}$ denote the fraction of members at or above $\mathrm{thr}$. The delivered forecast is
\begin{equation}
y(p)=\begin{cases}
\max_b x^{(b)}(p), & \hat{P}(p)\geq k/N,\ a(p)<\mathrm{thr},\\
\mathrm{thr}-\delta, & \hat{P}(p)<k/N,\ a(p)\geq\mathrm{thr},\\
a(p), & \text{otherwise},
\end{cases}
\label{eq:vote}
\end{equation}
where $k$ is the required number of votes. The first case fills heavy rain that the anchor misses and at least $k$ members predict. The second removes an anchor false alarm supported by fewer than $k$ members by setting the pixel to $\mathrm{thr}-\delta$ ($\delta=10^{-4}$), removing the unsupported extreme-tier value with a minimal change to the anchor. All other pixels keep the anchor's values, so no pixel is averaged. We set $\mathrm{thr}$ to the highest intensity tier of each dataset and $k=2$, the smallest vote that forms a consensus (\cref{tab:k-main}). The vote edits only pixels where the anchor and the ensemble disagree about this tier (effect on other tiers and metrics in \cref{subsec:abl-main2}), and neither branching nor decoding changes the trained generator.

%% file: sec/4_experiments.tex
\section{Experiments}
\subsection{Experimental Setup}
\label{subsec:setup}

\begin{figure*}[t]
  \centering
  \includegraphics[width=0.72\textwidth]{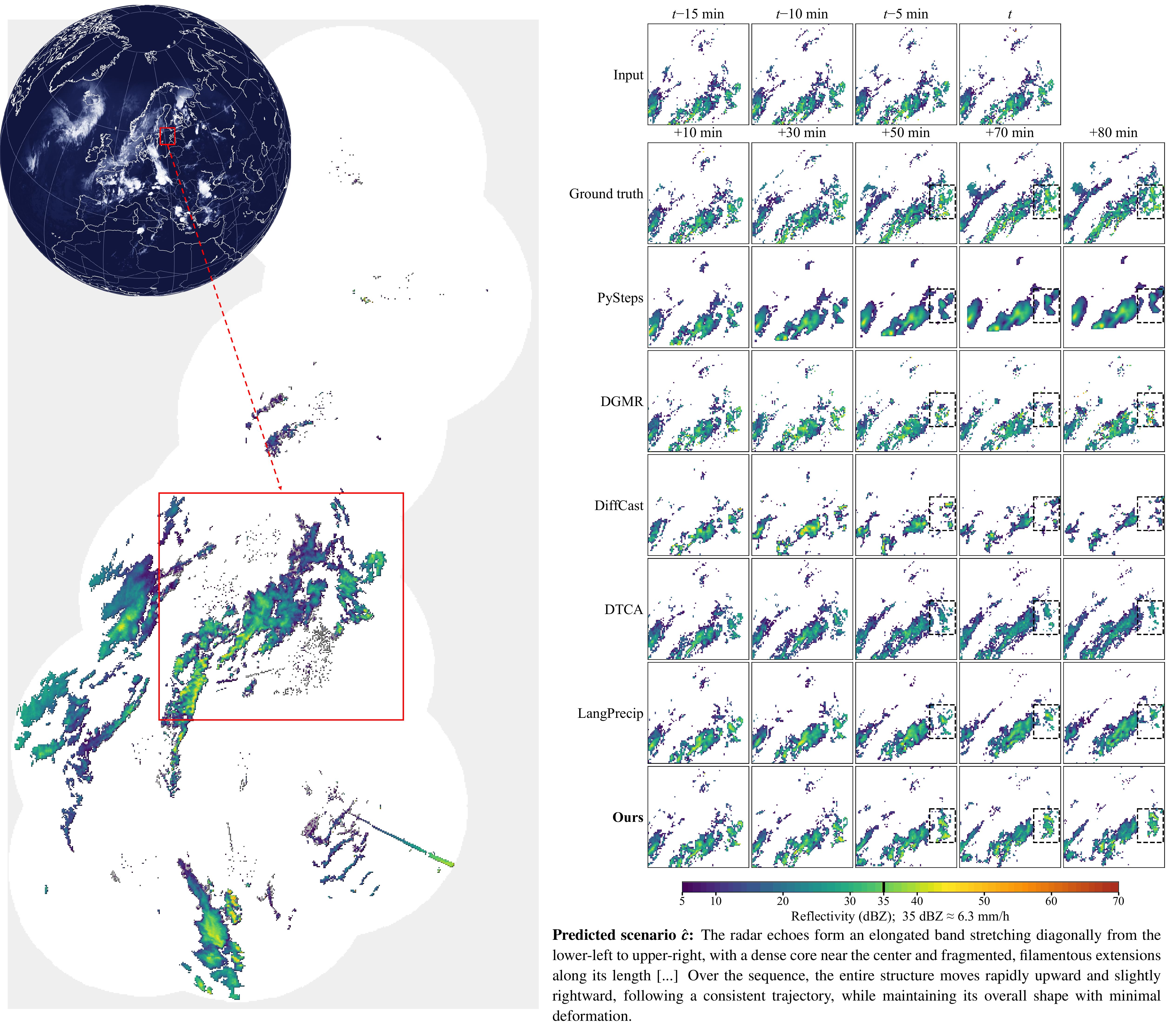}
\caption{\textbf{Case study of a precipitation event starting on 21 June 2021 at 18:05 UTC, showing a rain band with embedded heavy cores over central Sweden.} Left: the radar composite, with the red box marking the $256\times256$ crop ($58.5$--$62^\circ$N, $13$--$19^\circ$E), and the Meteosat IR10.8 image at 18:00 UTC. Right: the four input frames, the ground truth and forecasts at $+10$ to $+80$ min. Dashed boxes mark a region of heavy echoes at later lead times. The predicted scenario $\hat{c}$ for this case is shown at the bottom.}
  \label{fig:qualitative}
\end{figure*}

\noindent\textbf{Datasets.} We evaluate ForeSay on six datasets. The Swedish precipitation dataset and MRMS follow the evaluation setup of LangPrecip~\cite{ling2026langprecip} ($256\times256$, $5$-minute interval), forecasting $16$ future frames ($80$\,min) from $4$ historical observations on $1000$ test sequences each. SEVIR ($8096$ test sequences), MeteoNet ($1310$), Shanghai ($526$), and CIKM ($4000$) serve as four public radar benchmarks ($128\times128$ protocol), on which models predict $20$ future frames from $5$ inputs ($10$ frames for CIKM, App.~\ref{app:radar-benchmarks}). All inference hyperparameters ($s^*=4$, $\lambda=1$, $k=2$, $N=8$) are selected once on a $500$-sample held-out set of the Swedish dataset, disjoint from the training and test splits, and kept fixed on all six benchmarks.

\noindent\textbf{Metrics.} On the Swedish precipitation dataset and MRMS we follow LangPrecip~\cite{ling2026langprecip} and SSLDM-ISI~\cite{ling2024spacetime} and report CSI at a weak and a heavy threshold ($0.06$/$6.3~\mathrm{mm/h}$ for Sweden, $1$/$8~\mathrm{mm/h}$ for MRMS), FSS, CRPS, SSIM and LPIPS. On the four radar benchmarks we follow the protocol of DiffCast~\cite{yu2024diffcast}, AlphaPre~\cite{lin2025alphapre} and PercpCast~\cite{fengperceptually} with CSI-M, per-threshold CSI, HSS, SSIM and LPIPS.

\begin{table}[!t]
  \centering
  \caption{Comparison on the Swedish precipitation dataset and MRMS ($80$-minute lead time, averaged over $16$ frames). Ours is the decoded ensemble ($s=4$, $N=8$, $5.2\times$) with $\mathrm{thr}$ at the highest tier. LangPrecip on MRMS uses CFG scale $4$. Best in bold, second best underlined. FSS and CRPS are in App.~Tabs.~\ref{tab:sw-full} and~\ref{tab:mrms-full}.}
  \label{tab:AvG}\label{tab:AvGmrms}
  \setlength{\tabcolsep}{2.6pt}
  \resizebox{0.72\linewidth}{!}{%
  \begin{tabular}{l|cccc|cccc}
    \hline
     & \multicolumn{4}{c|}{\textbf{Swedish}} & \multicolumn{4}{c}{\textbf{MRMS}} \\
    Method & CSI$_{0.06}\uparrow$ & CSI$_{6.3}\uparrow$ & SSIM$\uparrow$ & LPIPS$\downarrow$
           & CSI$_{1}\uparrow$ & CSI$_{8}\uparrow$ & SSIM$\uparrow$ & LPIPS$\downarrow$ \\
    \hline
    PySteps~\cite{pulkkinen2019pysteps} & 0.473 & 0.049 & 0.589 & 0.284 & 0.338 & 0.024 & 0.871 & 0.187 \\
    DGMR~\cite{ravuri2021skilful} & 0.433 & 0.039 & 0.702 & 0.204 & 0.367 & 0.022 & 0.883 & 0.137 \\
    DiffCast~\cite{yu2024diffcast} & 0.439 & 0.029 & 0.660 & 0.292 & 0.328 & 0.018 & 0.869 & 0.183 \\
    SSLDM-ISI~\cite{ling2024spacetime} & 0.558 & 0.040 & 0.718 & 0.191 & 0.416 & 0.024 & 0.897 & \sbest{0.120} \\
    AlphaPre~\cite{lin2025alphapre} & 0.541 & 0.025 & 0.647 & 0.333 & 0.360 & 0.024 & 0.891 & 0.197 \\
    DTCA~\cite{li2024precipitation} & 0.572 & 0.046 & \sbest{0.732} & \textbf{0.180} & \sbest{0.423} & 0.026 & 0.899 & \textbf{0.118} \\
    RainHCNet~\cite{wang2025rainhcnet} & 0.499 & 0.025 & 0.332 & 0.320 & 0.320 & 0.002 & 0.877 & 0.207 \\
    SSRF-Net~\cite{luo2025ssrf} & 0.485 & 0.045 & 0.572 & 0.310 & 0.336 & 0.009 & 0.844 & 0.213 \\
    exPreCast~\cite{song2026extreme} & 0.489 & 0.028 & 0.520 & 0.348 & 0.394 & 0.015 & 0.828 & 0.203 \\
    FlowCast~\cite{ribeiro2026flowcast} & 0.482 & 0.009 & 0.680 & 0.324 & 0.393 & 0.013 & 0.900 & 0.192 \\
    SDIR~\cite{zhou2026sdir} & 0.424 & 0.019 & 0.718 & 0.307 & 0.349 & 0.013 & \textbf{0.907} & 0.189 \\
    PDRF~\cite{luo2026pdrf} & 0.474 & 0.037 & 0.571 & 0.314 & 0.383 & 0.020 & 0.875 & 0.190 \\
    LangPrecip~\cite{ling2026langprecip} & \sbest{0.586} & \sbest{0.074} & \sbest{0.732} & 0.193 & \textbf{0.435} & \sbest{0.031} & \sbest{0.905} & 0.142 \\
    \rowcolor{oursrow}Ours & \textbf{0.593} & \textbf{0.136} & \textbf{0.747} & \sbest{0.190} & \textbf{0.435} & \textbf{0.036} & 0.898 & 0.131 \\
    \hline
  \end{tabular}%
  }
\end{table}
\subsection{Results}
\label{sec:results}

\noindent\textbf{Quantitative evaluation and trade-off analysis.} On the four radar benchmarks (Tab.~\ref{tab:radar-main}), ForeSay ranks first on all three high-threshold CSIs and HSS. At the highest tier it outperforms the strongest baseline PDRF by $+21.6\%$ (SEVIR), $+10.9\%$ (MeteoNet), $+11.7\%$ (Shanghai) and $+12.8\%$ (CIKM), with the best multi-scale CSI-M (App.~Tab.~\ref{tab:app-csi}). On the Swedish and MRMS benchmarks (Tab.~\ref{tab:AvG}), heavy-rain CSI exceeds the text baseline LangPrecip by $+84\%$ ($0.074\to0.136$) and $+16\%$ ($0.031\to0.036$), and ours stays ahead at every lead time (Fig.~\ref{fig:leadtime}). Even our single-sample anchor ($1\times$) matches or beats the strongest baseline on all datasets but MRMS (Tab.~\ref{tab:decoding-gain}).

These results also expose the \emph{double-penalty} trade-off against metrics that favor blur (App.~\ref{app:sweden-extra}). Pixel-wise metrics (CRPS, MAE) penalize a displaced echo twice, as a false alarm and a miss, and the neighborhood metric FSS also tends to favor smooth rain fields, so smooth baselines (\eg FlowCast, SDIR) score well while omitting heavy echoes (CSI down to $0.009$). When displacement is negligible ($\leq10$~min), ForeSay has the lowest CRPS of all methods (App.~Fig.~\ref{fig:sw_temporal}). Over the full horizon, consensus decoding raises the Swedish CRPS by $6.8\%$ (at most $0.3\%$ on the other datasets, App.~Tabs.~\ref{tab:abl-main2} and~\ref{tab:abl-branch}) in exchange for a $+37.9\%$ gain in heavy-rain CSI.

\noindent\textbf{Qualitative results.} In Fig.~\ref{fig:qualitative}, at $+70$ and $+80$~min, PySteps and DTCA smooth the heavy echoes inside the dashed box, DiffCast nearly loses them, DGMR breaks them into speckles and LangPrecip weakens the core above $35$~dBZ ($\approx6.3$~mm/h). Guided by $\hat{c}$ (bottom), which describes the diagonal band and its rapid upward motion, our anchor keeps part of the heavy core, and decoding brings its extent and intensity closer to the ground truth without blurring. App.~\ref{app:sweden-extra} shows more examples on all six datasets.

%% file: sec/5_ablation.tex
\section{Ablation Studies}
\label{sec:ablation}

\subsection{Conditioning Components}
\label{subsec:abl-cond}
\begin{table}[t]
  \centering
  \small
  \caption{Conditioning ablation on the Swedish precipitation dataset. History text describes the observed frames as in LangPrecip, and future text is our predicted scenario. W/o branching: the anchor. W/ branching: the decoded ensemble.}
  \label{tab:ablation}
  \setlength{\tabcolsep}{1.6pt}
  \begin{tabular}{lccccc}
    \hline
     & \multicolumn{4}{c}{\shortstack{w/o branching\\($1\times$)}} & \shortstack{w/ branching\\($5.2\times$)}\\
    \cline{2-5}\cline{6-6}
    Condition & CSI$_{0.06}\uparrow$ & CSI$_{6.3}\uparrow$ & SSIM$\uparrow$ & FSS$\uparrow$ & CSI$_{6.3}\uparrow$\\ \hline
    Radar only & 0.578 & 0.0895 & 0.736 & 0.910 & 0.1273\\
    \quad + history text & 0.584 & 0.0949 & 0.743 & 0.914 & 0.1275\\
    \quad + future text & \textbf{0.593} & \textbf{0.0984} & \textbf{0.747} & \textbf{0.915} & \textbf{0.1357}\\
    \hline
  \end{tabular}
\end{table}

\begin{table}[t]
  \centering
  \small
  \caption{Agreement (\%, $\uparrow$) of the predicted $\hat{c}$ with the ground truth. Main: primary axis; L/R, U/D: left/right, up/down; 8-way: compass direction.}
  \label{tab:scenario-acc}
  \setlength{\tabcolsep}{3.2pt}
  \begin{tabular}{cccc|cccc}
    \hline
    \multicolumn{4}{c|}{Motion direction} & \multicolumn{4}{c}{Intensity and structure} \\
    Main & L/R & U/D & 8-way & Intens. & Decay & Split & Merge \\
    \hline
    95.1 & 90.9 & 90.8 & 84.2 & 76.7 & 73.6 & 75.4 & 90.8 \\
    \hline
  \end{tabular}
\end{table}

We first evaluate the role of predicted scenario text in guiding macroscopic storm evolution (Tab.~\ref{tab:ablation}). Under the unbranched setting, history text raises the heavy-rain CSI of radar alone from $0.0895$ to $0.0949$, and the predicted future scenario advances it to $0.0984$. With branching and consensus decoding, the two conditions diverge: history text gains almost nothing over radar alone ($0.1275$ \vs $0.1273$), whereas the future scenario reaches $0.1357$ ($+6.6\%$). While history text merely summarizes past observations, the predicted scenario guides the trunk toward the true evolution (Tab.~\ref{tab:scenario-acc}: $84.2\%$--$95.1\%$ on motion direction, $73.6\%$--$90.8\%$ on intensity and structure changes, App.~\ref{app:scenario-acc}). Members thus share the storm evolution and differ mainly locally, enabling consensus voting to recover heavy echoes (\cref{subsec:decoding}).

\subsection{Branched Sampling and Noise Strength}
\label{subsec:lambda-main}
\begin{figure}[t]
  \centering
  \includegraphics[width=0.85\linewidth]{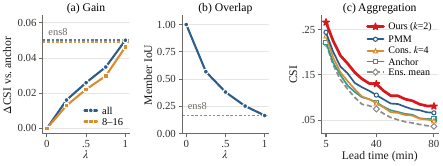}
  \caption{\textbf{Branching and aggregation ablations.} (a) CSI gain over the anchor and (b) member IoU at frame $12$ versus the noise strength $\lambda$ (held-out set), with ens8 dashed. (c) Heavy-rain CSI per lead time for different aggregations of the same $N=8$ members (test set, FB not comparable to Tab.~\ref{tab:k-main}). All-frame CSI (FB): ours $0.1357$ ($1.18$), PMM $0.1165$ ($0.73$), consensus $k=4$ $0.1002$ ($0.62$), anchor $0.0984$ ($0.73$), ensemble mean $0.0852$ ($0.50$).}
  \label{fig:ablation-curves}
\end{figure}

The noise strength $\lambda$ in Eq.~\eqref{eq:fork} balances member diversity and forecast skill (Fig.~\ref{fig:ablation-curves}a, b). At $\lambda=0$, members collapse to the anchor. As $\lambda$ increases, support IoU drops from $1.000$ to $0.166$ and paired CSI gains rise monotonically. At $\lambda=1$, the ensemble matches ens8 in both skill ($+0.0501$ \vs $+0.0503$) and diversity (IoU $0.166$ \vs $0.165$). Thus, late-stage branching retains skill and diversity while preserving macroscopic evolution (Fig.~\ref{fig:members}).

\subsection{Support-Consensus Decoding and Aggregation}
\label{subsec:abl-main2}
\label{sec:agg-compare}

\begin{table}[t]
  \centering
  \caption{\textbf{Effect of support-consensus decoding.} Anchor ($1.0\times$) \vs decoded ensemble ($5.2\times$), paired on the same anchor. Base.: the strongest baseline (Tabs.~\ref{tab:radar-main} and~\ref{tab:AvG}). $\Delta$: relative change (\%) of Ours over the anchor, with the sign set so that positive is better. Radar CSI uses the $128\times128$ protocol, and the other metrics use the native $256\times256$ output. Full values in App.~Tabs.~\ref{tab:abl-main2} and \ref{tab:abl-branch}.}
  \label{tab:decoding-gain}
  \footnotesize
  \setlength{\tabcolsep}{2.0pt}
  \begin{tabular}{l|cccc|cccc}
    \hline
     & \multicolumn{4}{c|}{CSI$\uparrow$, highest tier} & \multicolumn{4}{c}{$\Delta$ (\%), other metrics} \\
    Dataset & Base. & Anchor & Ours & $\Delta$ (\%) & CSI$_{\mathrm{2nd}}$ & SSIM & LPIPS & CRPS \\
    \hline
    Swedish & .074 & .0984 & .1357 & \textbf{+37.9} & 0.0 & 0.0 & +0.1 & $-$6.8 \\
    MRMS & .031 & .0278 & .0360 & \textbf{+29.5} & 0.0 & 0.0 & $-$0.2 & $-$0.3 \\
    SEVIR & .1199 & .1202 & .1458 & \textbf{+21.3} & +4.1 & +0.6 & +0.6 & +1.9 \\
    MeteoNet & .2679 & .2747 & .2971 & \textbf{+8.2} & +0.3 & +0.2 & +0.1 & +0.5 \\
    Shanghai & .3032 & .3217 & .3387 & \textbf{+5.3} & +0.4 & 0.0 & +0.1 & $-$0.1 \\
    CIKM & .1720 & .1744 & .1941 & \textbf{+11.3} & +0.9 & $-$0.1 & +0.1 & $-$0.3 \\
    \hline
  \end{tabular}
\end{table}

Aggregating ensemble members without blurring convective extremes is critical (Fig.~\ref{fig:ablation-curves}c). Pixel averaging drops heavy-rain CSI to $0.0852$ (FB $0.50$), below the anchor's $0.0984$. While PMM restores amplitudes to $0.1165$, it keeps the mean pattern. Support-consensus decoding avoids the mean field entirely, reaching $0.1357$ ($+16.5\%$ over PMM) and leading at every lead time. Because members differ mainly in heavy cores (Fig.~\ref{fig:members}), gains concentrate at the highest tier (Tab.~\ref{tab:decoding-gain}, $+5.3\%$ to $+37.9\%$), while second-tier CSI, SSIM, LPIPS, and CRPS mostly stay within $\pm1\%$ (exceptions: SEVIR second-tier CSI $+4.1\%$ and CRPS $+1.9\%$, Swedish CRPS $-6.8\%$), as voting only edits heavy-rain regions at $\mathrm{thr}$. Across the four radar benchmarks, $19$ of $22$ CSIs and all four HSS improve (App.~\ref{subsec:abl-branch}).

\begin{table}[!t]
  \centering
  \small
  \caption{Number of votes $k$ (top, $N=8$) and ensemble size $N$ (bottom, $k=2$) on the held-out set. CSI is at $\geq6.3$~mm/h and FB is the frequency bias ($1$ is unbiased). $N=1$ is the anchor. Full sweeps are in App.~Tabs.~\ref{tab:k-sweep} and~\ref{tab:n-sweep}.}
  \label{tab:k-main}
  \setlength{\tabcolsep}{2.5pt}
  \begin{tabular}{l|cccccccc}
    \hline
    $k$ & 1 & \textbf{2} & 3 & 4 & 5 & 6 & 7 & 8 \\
    \hline
    CSI$\uparrow$ & \textbf{.1760} & .1740 & .1518 & .1266 & .1036 & .0821 & .0608 & .0420 \\
    FB$\to1$ & 1.92 & \textbf{1.05} & 0.70 & 0.50 & 0.37 & 0.27 & 0.19 & 0.12 \\
    \hline
    $N$ & 1 & 2 & 4 & 6 & \textbf{8} & 16 & & \\
    \hline
    CSI$\uparrow$ & .1239 & .0893 & .1399 & .1649 & .1740 & \textbf{.1867} & & \\
    FB$\to1$ & 0.66 & 0.33 & 0.65 & 0.88 & \textbf{1.05} & 1.53 & & \\
    \hline
  \end{tabular}
\end{table}

\noindent\textbf{Vote count and ensemble size.} In \cref{tab:k-main} (top), union voting $k=1$ gives the highest raw CSI but doubles the rain area with an excessive FB of $1.92$. In contrast, $k=2$, the minimal consensus requirement, preserves nearly the same CSI at a near-unbiased FB of $1.05$. With $k=2$ (bottom), $N=2$ reduces to the intersection of two members and under-forecasts (FB $0.33$). Scaling to $N=16$ adds $0.0127$ CSI at $1.9\times$ compute cost, mainly inflating rain area (FB $1.53$). We therefore use $k=2$ and $N=8$.

%% file: sec/6_conclusion.tex
\section{Conclusion}
\label{sec:conclusion}

ForeSay separates the two scales of nowcasting uncertainty: the macroscopic evolution is forecast as scenario text, and only the local details are sampled. Members no longer need their own early sampling steps, so they share a trunk and match eight independent samples in skill and diversity at $5.2\times$ instead of $8\times$ compute. Their remaining differences are local, so pixel-wise voting can replace pixel averaging and keep the heavy echoes that averaging smooths out. The branching step is read from the generator's own posterior width. Across six datasets, ForeSay achieves the best heavy-precipitation CSI, $11\%$--$84\%$ above the strongest baseline, and the ablations show that text helps when it predicts the future rather than summarizes the past. These results suggest that a nowcasting ensemble can forecast the storm system once and spend its randomness on where heavy convective cores form.

\noindent\textbf{Limitations and future work.} Conditioning all members on a single predicted scenario leaves macroscopic uncertainty unmodeled, so any error in the text propagates across the entire ensemble---especially when radar history cannot resolve whether a storm system will intensify, decay, or split. Sampling several scenarios in such cases, each with its own branches, would capture uncertainty at both scales. Separately, because consensus voting acts pixel-wise, it keeps heavy cores sharp at their forecast locations and thus raises CRPS when those positions drift.
\clearpage

%% file: sec/X_appendix.tex
\raggedbottom
\renewcommand{\topfraction}{0.95}\renewcommand{\bottomfraction}{0.6}\renewcommand{\textfraction}{0.05}\renewcommand{\floatpagefraction}{0.8}
\setcounter{topnumber}{3}\setcounter{bottomnumber}{2}\setcounter{totalnumber}{5}
\appendix
\section{Details of the Branching Operator}
\label{app:operator}
\begin{algorithm}[!htbp]
\caption{Branched RF sampling. Lines in \modline{teal} depart from standard RF sampling.}
\label{alg:branched-rf}
\begin{algorithmic}[1]
\Require $h, c$, branching step $s$, members $N$, strength $\lambda$, RF grid $1=\tau_0>\cdots>\tau_T=0$
\Ensure members $x^{(0)},\dots,x^{(N-1)}$, anchor $x^{(0)}$
\State $z \leftarrow \varepsilon_0 \sim \mathcal{N}(0, I)$
\For{$i = 0 \ldots s-1$} \Comment{shared trunk}
    \State $v \leftarrow v_\theta(z, \tau_i; h, c)$; \quad $z \leftarrow z + v \cdot (\tau_i - \tau_{i+1})$
\EndFor
\State \modline{$\hat{x}_0 \leftarrow z + \tau_s v$; \quad $\hat{\varepsilon} \leftarrow z + (\tau_s - 1) v$} \Comment{preview}
\For{$b = 0 \ldots N-1$} \Comment{branch, $x^{(0)}$ is the anchor}
    \State \modline{$\varepsilon_b \sim \mathcal{N}(0, I)$} \Comment{re-noise at $\tau_s$}
    \State \modline{$z_b \leftarrow (1-\tau_s)\,\hat{x}_0 + \tau_s\bigl(\sqrt{1-\lambda^2}\,\hat{\varepsilon} + \lambda\,\varepsilon_b\bigr)$}
\EndFor
\For{$b = 0 \ldots N-1$}
    \For{$i = s \ldots T-1$}
        \State $v \leftarrow v_\theta(z_b, \tau_i; h, c)$; \quad $z_b \leftarrow z_b + v \cdot (\tau_i - \tau_{i+1})$
    \EndFor
    \State $x^{(b)} \leftarrow \mathcal{D}(z_b)$
\EndFor
\end{algorithmic}
\end{algorithm}
\begin{algorithm}[!htbp]
\caption{Support-consensus decoding}
\label{alg:support-consensus}
\begin{algorithmic}[1]
\Require members $x^{(0..N-1)}$, anchor $a = x^{(0)}$, threshold $\mathrm{thr}$, vote $k$
\Ensure delivered field $y$
\State $y \leftarrow a$
\State $\hat{P}(p) \leftarrow \frac{1}{N}\, \#\{b : x^{(b)}(p) \geq \mathrm{thr}\}$
\State $\mathrm{ADD} \leftarrow \{p : \hat{P}(p) \geq k/N \wedge a(p) < \mathrm{thr}\}$
\State $\mathrm{REM} \leftarrow \{p : \hat{P}(p) < k/N \wedge a(p) \geq \mathrm{thr}\}$
\State $y[\mathrm{ADD}] \leftarrow \max_b x^{(b)}[\mathrm{ADD}]$; \quad $y[\mathrm{REM}] \leftarrow \mathrm{thr} - 10^{-4}$
\end{algorithmic}
\end{algorithm}
Algorithms~\ref{alg:branched-rf} and~\ref{alg:support-consensus} give the full procedures of branched sampling (Eqs.~\eqref{eq:fork}--\eqref{eq:branch} of the main text) and support-consensus decoding (Eq.~\eqref{eq:vote}).

With $\tau=0$ at the data end and $\tau=1$ at the noise end, Rectified Flow uses the path $z_\tau=(1-\tau)x_0+\tau\varepsilon$, and the network predicts the velocity $v_\theta\approx x_0-\varepsilon$. At a branching step $\tau_s$ the current state can therefore be written as
\[
z\approx(1-\tau_s)\hat{x}_0+\tau_s\hat{\varepsilon},
\]
where $\hat{x}_0=z+\tau_s v_\theta$ is the estimated data endpoint and $\hat{\varepsilon}=z+(\tau_s-1)v_\theta$ the estimated noise endpoint. The branching step in Alg.~\ref{alg:branched-rf} keeps $\hat{x}_0$ and re-noises it to the same level $\tau_s$,
\[
z_b\approx(1-\tau_s)\hat{x}_0
+\tau_s\left(\sqrt{1-\lambda^2}\,\hat{\varepsilon}
+\lambda\varepsilon_b\right).
\]
The step $\tau_s$ sets the weight between the data and noise ends, and $\lambda$ sets how much of the old noise is replaced. At $\lambda=1$ the update reduces to $z_b = z + \tau_s(\varepsilon_b-\hat{\varepsilon})$, so that
\[
z_b\approx \bigl((1-\tau_s)\hat{x}_0+\tau_s\hat{\varepsilon}\bigr)+\tau_s(\varepsilon_b-\hat{\varepsilon})=(1-\tau_s)\hat{x}_0+\tau_s\varepsilon_b .
\]
Branching thus removes the old noise component and keeps the data component $(1-\tau_s)\hat{x}_0$. To match the base sampler, our experiments compute $\hat{x}_0$ and $\hat{\varepsilon}$ from the velocity of the last trunk step. An exact variant spends one more forward pass at the branching state, which changes the estimates of $\hat{x}_0$ and $\hat{\varepsilon}$ but not the operator. The members share only this endpoint. After branching, each of them evolves independently under the nonlinear velocity field.

\noindent\textbf{Why a useful branching step lies inside the trajectory.}
The same decomposition predicts where branching can work, independently of the dataset. In the branched state
\[
z_b\approx(1-\tau_s)\hat{x}_0+\tau_s\varepsilon_b
\]
the weight $1-\tau_s$ sets how strongly the shared endpoint binds the members to one macroscopic evolution, and $\tau_s$ sets how much independent randomness each member keeps.
\begin{itemize}
    \item \emph{Branching too early} ($\tau_s\to1$). The shared $\hat{x}_0$ carries almost no weight and every member restarts from nearly pure noise. The trunk saves little compute, and the ensemble approaches independent sampling.
    \item \emph{Branching too late} ($\tau_s\to0$). The fresh noise $\varepsilon_b$ carries almost no weight, all members start close to the same $\hat{x}_0$, and only a few steps remain for them to diverge. The ensemble collapses toward copies of one forecast and leaves the vote little to work with.
    \item \emph{In between}, $\hat{x}_0$ is strong enough to hold all members to one evolution, and $\varepsilon_b$ is strong enough to make them disagree on local detail.
\end{itemize}
A useful branching step therefore lies inside the trajectory. The two failures are asymmetric. Branching too early leaves skill intact and only loses the saving, whereas branching too late hurts both diversity and skill. As the branching step moves later, skill should stay flat and then drop. This argument shows that such a step exists but not where it is. Its location is given by the posterior-width criterion of Eq.~\eqref{eq:sstar} in the main text (Appendix~\ref{app:probe}) and checked by the sweep (Appendix~\ref{app:s-sweep-stats}).

\noindent\textbf{Anchor cost.} All experiments are self-anchored. The anchor $x^{(0)}$ is the first branch of the same run. Like every member it receives fresh noise at step $s$, and on its own it is a complete $T$-step sample ($1.0\times$). An external anchor would cost one more full sampling ($+T$ steps).

\section{Posterior-Width Probe}
\label{app:probe}

\noindent\textbf{Why a probe.} Appendix~\ref{app:operator} shows that skill is flat for early branching and drops for late branching, but it does not say where the drop begins. A sweep over $s$ needs ground truth and a full ensemble at every candidate step, and it locates the drop without explaining it. The posterior width $U_f(\tau)$ of Eq.~\eqref{eq:uf} measures how much of frame $f$ the current state still leaves undetermined, and hence how much members branched from that state can still differ. It is computed from the velocity field along one sampling trajectory, without ground truth and without branched members. The criterion of Eq.~\eqref{eq:sstar} branches at the last step before the late-frame width starts to contract.

\noindent\textbf{Frame-structured Hutchinson probe with JVPs.} The full Jacobian covers all $16\times32768$ latent coordinates, and even one per-frame block has $32768\times32768$ entries. To avoid forming either, we use the Hutchinson estimator $\mathbb{E}[n^\top J n] = \mathrm{tr}(J)$ and restrict each probe vector to the coordinates of one frame, so that the estimate counts that frame only:

\begin{lstlisting}[language=Python, basicstyle=\ttfamily\footnotesize, breaklines=true, columns=fullflexible, frame=tb, framesep=5pt, framerule=0.6pt, xleftmargin=6pt, xrightmargin=6pt, aboveskip=10pt, belowskip=8pt, captionpos=t, caption={Frame-structured Hutchinson probe.}, label={lst:divergence-probe}]
signs = randint(0, 2, z.shape) * 2 - 1       # Rademacher +-1, drawn once for z
for fr in range(16):                         # loop over frames
    n = zeros_like(z); n[:, :, fr] = signs[:, :, fr]   # probe on frame fr
    _, jv = jvp(lambda zz: cfg_velocity(zz, t), (z,), (n,))  # jv = J n
    divs[pi, fr, dr] = float((jv * n).sum())  # n^T J n, estimate of tr(J_ff)
\end{lstlisting}

The estimator follows three choices. First, each probe vector is nonzero only on the coordinates of one frame, so one Jacobian--vector product (JVP) gives the divergence of one frame. A $16$-frame forecast thus needs $16$ JVPs per probe sample, each costing about one forward pass. Second, the Rademacher signs \texttt{signs} are drawn once for the whole latent and sliced per frame, so all frames of one sample share the same signs. These common random numbers reduce the noise in the difference between early and late frames. Third, the probe is evaluated only at the grid points of the sampler, and between them $z$ follows the standard update $z \leftarrow z + v\,\Delta\tau$. The divergence is therefore measured at states that the sampler actually visits.

Fig.~\ref{fig:posterior-width}a of the main text compares the normalized widths of early and late frames across sampling steps. The late-frame width peaks at step $4$, where the early frames are also close to their peak, so Eq.~\eqref{eq:sstar} gives $s^*=4$. After this step the late-frame width falls monotonically, to about $0.97$, $0.85$, $0.67$ and $0.43$ of its peak at $s=5$--$8$. Over the same range the member IoU rises ($0.190$, $0.220$, $0.228$, $0.297$) and the CSI drops from $s=7$ on (Table~\ref{tab:s-sweep}).

\section{Accuracy of the Predicted Scenario}
\label{app:scenario-acc}
We compare the scenario $\hat{c}$ predicted from the history alone with labels computed from the ground-truth future radar, without involving the teacher. From $\hat{c}$ we extract the motion direction and the intensity and structural changes. The reference motion direction is obtained by optical flow on the true future frames, and the reference intensity and structural changes are computed from the true future radar fields. We report the fraction of the $1000$ Swedish precipitation test cases on which the two agree (Tab.~\ref{tab:scenario-acc} of the main text). The predicted scenario agrees with the ground truth on the main motion direction in $95.1\%$ of the cases and on the 8-way direction in $84.2\%$. Agreement on intensity and structural change ranges from $73.6\%$ to $90.8\%$.

\section{Hyperparameter Sweeps}
\label{app:sweeps}
All sweeps use the $500$-sample Swedish precipitation held-out set.

\subsection{Branching Step $s$}
\label{app:s-sweep-stats}
The posterior width selects $s^*$ without labels. To check the choice on the reported metric, we generate $N$ branches at every candidate $s$ and measure two quantities. The heavy-rain CSI tells whether branching still brings a forecast gain, and the IoU of the precipitation supports of all $N$ members tells whether the members remain diverse (lower is more diverse).

The results (Fig.~\ref{fig:posterior-width}b of the main text and Table~\ref{tab:s-sweep}) show a \emph{plateau} followed by a \emph{drop}. For $s\in[2,4]$ the all-frame CSI is $0.1740$--$0.1753$, within $0.002$ of the independent ensemble (ens8, $0.1757$). At $s=5$ and $6$ it is only $0.0041$ and $0.0052$ below $s=4$, still recovering $102\%$ and $93\%$ of the ens8 gain. From $s=7$ on the CSI drops clearly ($0.1632$ and $0.1559$ at $s=7$ and $8$, recovering $79\%$ and $66\%$), and the late frames ($f=8$--$16$) drop more. The member IoU rises from $0.166$ at $s=4$ to $0.297$ at $s=8$, so later branching gives more similar members and leaves the vote less to work with.

This shape matches Appendix~\ref{app:operator}. Early branching leaves skill intact and only costs compute, as $s=2$ and $3$ reach the CSI of $s=4$ at $6.6\times$ and $5.9\times$. Late branching hurts both diversity and skill.

\noindent\textbf{Why $s^*=4$.} Within the plateau later branching is cheaper, and $s=6$ still recovers $93\%$ of the ens8 gain at $3.8\times$. We keep $s=4$ for two reasons.
\begin{enumerate}
    \item \textbf{Margin to the drop.} $s=6$ lies next to the drop that starts at $s=7$, and its late frames ($f=8$--$16$) recover only $82\%$ of the ens8 gain against $94\%$ at $s=4$.
    \item \textbf{Unchanged diversity.} The member IoU at $s=4$ ($0.166$) equals that of independent sampling ($0.165$), while at $s=5$ and $s=6$ it has already risen to $0.190$ and $0.220$, as the late-frame width has started to contract ($0.97$ and $0.85$ of its peak).
\end{enumerate}
Deployments with a tighter compute budget may use $s=6$ ($3.8\times$) at a slightly lower gain at long lead times.

\begin{table}[H]
  \centering
  \small
  \caption{Sweep of the branching step $s$ on the Swedish precipitation held-out set ($N=8$, $T=10$, $\lambda=1$, $k=2$). Steps is the number of network evaluations per forecast, $s+N(T-s)$, and cost is relative to a single sample ($10$ steps). CSI is at the highest tier ($\geq 6.3~\mathrm{mm/h}$), computed per sample and frame and then averaged. ``Of $8\times$'' is the paired CSI gain of each setting over its own anchor (the first branch of the same run), as a fraction of the gain of ens8 over its own first member. The anchor is re-noised at its own branching step, so this fraction need not follow the order of the absolute CSI. Member IoU is the intersection over union of the precipitation supports of all $N$ members in the $12$th forecast frame, and lower means more diverse members.}
  \label{tab:s-sweep}
  \resizebox{\ifdim\width>\linewidth\linewidth\else\width\fi}{!}{%
  \begin{tabular*}{\textwidth}{@{\extracolsep{\fill}}lcccccc}
    \hline
    Setting & Steps & Cost & CSI$\uparrow$ (all 16 frames) (of $8\times$) & CSI$\uparrow$ ($f=8$--$16$) (of $8\times$) & Member IoU ($f=12$) \\
    \hline
    $s=2$, $N=8$ & $66$ & $6.6\times$ & $0.1743$ ($97\%$) & $0.1366$ ($91\%$) & $0.151$ \\
    $s=3$, $N=8$ & $59$ & $5.9\times$ & $0.1753$ ($85\%$) & $0.1379$ ($87\%$) & $0.174$ \\
    $\mathbf{s=4}$, $\mathbf{N=8}$ (default) & $52$ & $5.2\times$ & $0.1740$ ($100\%$) & $0.1355$ ($94\%$) & $0.166$ \\
    $s=5$, $N=8$ & $45$ & $4.5\times$ & $0.1699$ ($102\%$) & $0.1339$ ($103\%$) & $0.190$ \\
    $s=6$, $N=8$ & $38$ & $3.8\times$ & $0.1688$ ($93\%$) & $0.1275$ ($82\%$) & $0.220$ \\
    $s=7$, $N=8$ & $31$ & $3.1\times$ & $0.1632$ ($79\%$) & $0.1244$ ($73\%$) & $0.228$ \\
    $s=8$, $N=8$ & $24$ & $2.4\times$ & $0.1559$ ($66\%$) & $0.1145$ ($70\%$) & $0.297$ \\
    \hline
    ens8 (independent sampling) & $80$ & $8.0\times$ & $0.1757$ ($100\%$) & $0.1412$ ($100\%$) & $0.165$ \\
    \hline
  \end{tabular*}%
  }
\end{table}

\subsection{Number of Votes $k$}
\label{app:k-sweep}
\begin{table}[!t]
  \centering
  \caption{Sweep of the number of votes $k$ (Swedish precipitation held-out set, $500$ samples, $s=4$, $N=8$, $\lambda=1$). CSI is at the highest tier ($\geq 6.3~\mathrm{mm/h}$). FB is the frequency bias, the number of forecast pixels above the threshold divided by the observed number, so $1$ is unbiased, $>1$ overestimates and $<1$ underestimates the rain area. $k=1$ takes the union of all member supports, and $k=2$ is the smallest vote that forms a consensus.}
  \label{tab:k-sweep}
  \resizebox{\ifdim\width>\linewidth\linewidth\else\width\fi}{!}{%
  \begin{tabular*}{\textwidth}{@{\extracolsep{\fill}}cccc}
    \hline
    $k$ & CSI (all 16 frames)$\uparrow$ & CSI ($f=8$--$16$)$\uparrow$ & FB (closer to $1$ is better) \\
    \hline
    $1$ & $\mathbf{0.1760}$ & $\mathbf{0.1449}$ & $1.92$ \\
    $\mathbf{2}$ & $0.1740$ & $0.1355$ & $\mathbf{1.05}$ \\
    $3$ & $0.1518$ & $0.1099$ & $0.70$ \\
    $4$ & $0.1266$ & $0.0862$ & $0.50$ \\
    $5$ & $0.1036$ & $0.0652$ & $0.37$ \\
    $6$ & $0.0821$ & $0.0455$ & $0.27$ \\
    $7$ & $0.0608$ & $0.0287$ & $0.19$ \\
    $8$ & $0.0420$ & $0.0172$ & $0.12$ \\
    \hline
  \end{tabular*}%
  }
\end{table}

$k$ is the number of members that must exceed the threshold for a pixel to count as wet, so a smaller $k$ gives a larger support. We sweep $k=1$--$8$ on the held-out set (Table~\ref{tab:k-sweep}) and do not simply pick the $k$ with the highest CSI. $k=1$ reaches the highest CSI, but it takes the union of all member supports and forms no consensus. Its FB of $1.92$ means that the forecast rain area is about twice the observed one, and its gain comes from this inflated area. $k=2$ is the smallest vote in which two members must agree. Its CSI is close to that of $k=1$ ($0.1740$ \vs $0.1760$ over all $16$ frames), and its area overestimate almost disappears (FB $1.05$). From $k=3$ on the FB falls below $1$ and the area is underestimated, and the CSI falls with $k$, to $0.1518$ at $k=3$ and $0.0420$ at $k=8$, where all members must agree. A strict vote thus deletes much real precipitation. We therefore use $k=2$.

\subsection{Number of Members $N$}
\label{app:n-sweep}
\begin{table}[!t]
  \centering
  \caption{Sweep of the number of members $N$ with the vote fixed at $k=2$ (Swedish precipitation held-out set, $500$ samples, $s=4$, $\lambda=1$). CSI is at the highest tier ($\geq 6.3~\mathrm{mm/h}$). Cost is $s+N(T-s)$ network evaluations relative to a single sample ($10$ steps). FB is the frequency bias, and $1$ is unbiased.}
  \label{tab:n-sweep}
  \resizebox{\ifdim\width>\linewidth\linewidth\else\width\fi}{!}{%
  \begin{tabular*}{\textwidth}{@{\extracolsep{\fill}}cccccc}
    \hline
    $N$ & Cost & CSI ($f=1$--$4$)$\uparrow$ & CSI ($f=8$--$16$)$\uparrow$ & CSI (all 16 frames)$\uparrow$ & FB (closer to $1$ is better) \\
    \hline
    $1$ (anchor, no vote) & $1.0\times$ & $0.1986$ & $0.0892$ & $0.1239$ & $0.66$ \\
    $2$ & $1.6\times$ & $0.1641$ & $0.0553$ & $0.0893$ & $0.33$ \\
    $4$ & $2.8\times$ & $0.2186$ & $0.1009$ & $0.1399$ & $0.65$ \\
    $6$ & $4.0\times$ & $0.2429$ & $0.1245$ & $0.1649$ & $0.88$ \\
    $\mathbf{8}$ & $5.2\times$ & $0.2481$ & $0.1355$ & $0.1740$ & $\mathbf{1.05}$ \\
    $16$ & $10.0\times$ & $\mathbf{0.2547}$ & $\mathbf{0.1538}$ & $\mathbf{0.1867}$ & $1.53$ \\
    \hline
  \end{tabular*}%
  }
\end{table}

Table~\ref{tab:n-sweep} varies the number of members on the held-out set with the vote fixed at $k=2$. $N=1$ is the anchor alone without voting. At $N=2$ the vote requires both members to agree, which removes real precipitation and drops the CSI below the anchor (FB $0.33$). From $N=4$ on, the CSI rises with $N$ and the frequency bias moves from an underestimate to an overestimate. $N=8$ gives the bias closest to $1$ ($1.05$). $N=16$ adds $0.013$ CSI at nearly twice the cost, and its bias of $1.53$ shows that the extra gain comes largely from an inflated rain area, the same effect as a looser vote (Appendix~\ref{app:k-sweep}). We therefore use $N=8$.

\section{Comparison on Radar Extrapolation Benchmarks}
\label{app:radar-benchmarks}

Beyond the Swedish and MRMS datasets, we compare with existing methods on four public radar extrapolation benchmarks, SEVIR, MeteoNet, Shanghai and CIKM. CSI-M averages over thresholds and hides where a gain occurs. The main text (Table~\ref{tab:radar-main}) therefore reports CSI at the three highest thresholds together with HSS, SSIM and LPIPS, and Table~\ref{tab:app-csi} lists CSI-M and CSI at every threshold of each benchmark. As in \S\ref{subsec:setup}, we follow the protocol of DiffCast~\cite{yu2024diffcast}, AlphaPre~\cite{lin2025alphapre} and PercpCast~\cite{fengperceptually} and evaluate all metrics at $128\times128$ (Appendix~\ref{app:resolution}). Best results are in bold and second best are \underline{underlined}. Table~\ref{tab:app-data} gives the test split, sample count and forecast setting of each benchmark.

\begin{table*}[htbp]
  \centering
  \small
  \caption{Evaluation settings of the four radar extrapolation benchmarks. The sample count is the number of test samples, and each sample has $5$ input frames. Thresholds follow the convention of each benchmark, in VIL counts ($0$--$255$) for SEVIR and in dBZ for the others.}
  \label{tab:app-data}
  \begin{tabular*}{\textwidth}{@{\extracolsep{\fill}}lcccc}
    \hline
    Dataset & Evaluation split & Samples & Condition frames $\to$ forecast frames & Intensity thresholds \\
    \hline
    SEVIR (VIL) & official test & $8096$ & $5 \to 20$ & $16/74/133/160/181/219$ \\
    MeteoNet    & test      & $1310$ & $5 \to 20$ & $12/18/24/32$ \\
    Shanghai    & official test & $526$  & $5 \to 20$ & $20/30/35/40$ \\
    CIKM        & test      & $4000$ & $5 \to 10$ & $20/30/35/40$ \\
    \hline
  \end{tabular*}

\end{table*}

\begin{table*}[htbp]
  \centering
  \caption{CSI-M and CSI at every intensity threshold on the four radar extrapolation benchmarks (official test sets, $128\times128$ protocol, sample counts in parentheses). Best in bold, second best underlined. CSI-M is the CSI averaged over thresholds. The subscripts are the thresholds of each benchmark, in VIL ($0$--$255$) for SEVIR and in dBZ for MeteoNet, Shanghai and CIKM.}
  \label{tab:app-csi}

  \resizebox{\ifdim\width>\textwidth\textwidth\else\width\fi}{!}{%
  \begin{tabular}{lccccccc|ccccc|ccccc|ccccc}
    \hline
     & \multicolumn{7}{c|}{\textbf{SEVIR} ($8096$)} & \multicolumn{5}{c|}{\textbf{MeteoNet} ($1310$)} & \multicolumn{5}{c|}{\textbf{Shanghai} ($526$)} & \multicolumn{5}{c}{\textbf{CIKM} ($4000$)} \\
    Method & CSI-M$\uparrow$ & CSI-16$\uparrow$ & CSI-74$\uparrow$ & CSI-133$\uparrow$ & CSI-160$\uparrow$ & CSI-181$\uparrow$ & CSI-219$\uparrow$
           & CSI-M$\uparrow$ & CSI-12$\uparrow$ & CSI-18$\uparrow$ & CSI-24$\uparrow$ & CSI-32$\uparrow$
           & CSI-M$\uparrow$ & CSI-20$\uparrow$ & CSI-30$\uparrow$ & CSI-35$\uparrow$ & CSI-40$\uparrow$
           & CSI-M$\uparrow$ & CSI-20$\uparrow$ & CSI-30$\uparrow$ & CSI-35$\uparrow$ & CSI-40$\uparrow$ \\
    \hline
    MIMO-VP~\cite{ning2023mimo} & 0.2931 & 0.6643 & 0.5562 & 0.2509 & 0.1492 & 0.1102 & 0.0280
                 & 0.2878 & 0.4701 & 0.3838 & 0.2469 & 0.0505
                 & 0.3108 & 0.4954 & 0.3722 & 0.2586 & 0.1171
                 & 0.2492 & 0.5932 & 0.1933 & 0.1351 & 0.0752 \\
    SwinLSTM~\cite{tang2023swinlstm} & 0.3110 & 0.6761 & 0.6015 & 0.2738 & 0.1585 & 0.1188 & 0.0373
                 & 0.3778 & 0.5161 & 0.4558 & 0.3631 & 0.1762
                 & 0.3839 & 0.5347 & 0.4409 & 0.3460 & 0.2141
                 & 0.3002 & 0.6217 & \sbest{0.2990} & 0.2002 & 0.0799 \\
    DiffCast~\cite{yu2024diffcast} & 0.3083 & 0.6586 & 0.5410 & 0.2578 & 0.1740 & 0.1435 & 0.0750
                 & 0.3693 & 0.5084 & 0.4427 & 0.3496 & 0.1767
                 & 0.3655 & 0.5010 & 0.4072 & 0.3254 & 0.2286
                 & 0.3067 & 0.6053 & 0.2792 & 0.2161 & 0.1261 \\
    AlphaPre~\cite{lin2025alphapre} & 0.3249 & 0.6984 & 0.6007 & 0.2883 & 0.1754 & 0.1332 & 0.0535
                 & 0.3627 & 0.4926 & 0.4315 & 0.3461 & 0.1807
                 & 0.3972 & 0.5497 & 0.4514 & 0.3591 & 0.2287
                 & 0.2853 & 0.5928 & 0.2604 & 0.1810 & 0.1070 \\
    PercpCast~\cite{fengperceptually} & 0.3492 & 0.6856 & 0.5844 & 0.3101 & 0.2203 & 0.1929 & 0.1022
                 & 0.4039 & 0.5279 & 0.4737 & 0.3933 & 0.2207
                 & 0.3995 & 0.5316 & 0.4459 & 0.3628 & 0.2577
                 & 0.3163 & 0.6108 & 0.2766 & 0.2223 & 0.1553 \\
    exPreCast~\cite{song2026extreme} & 0.3401 & 0.6636 & 0.6004 & 0.3266 & 0.2133 & 0.1649 & 0.0716
                 & 0.3973 & 0.5265 & 0.4704 & 0.3833 & 0.2090
                 & 0.3901 & 0.5412 & 0.4399 & 0.3471 & 0.2323
                 & 0.2927 & 0.6221 & 0.2729 & 0.1873 & 0.0886 \\
    FlowCast~\cite{ribeiro2026flowcast} & 0.3506 & \sbest{0.7083} & \textbf{0.6230} & 0.3276 & 0.2149 & 0.1661 & 0.0638
                 & 0.3898 & 0.5367 & 0.4705 & 0.3715 & 0.1803
                 & 0.4272 & 0.5725 & 0.4815 & 0.3927 & 0.2622
                 & 0.3162 & 0.6177 & 0.2954 & 0.2193 & 0.1324 \\
    SDIR~\cite{zhou2026sdir} & 0.3606 & \textbf{0.7108} & \sbest{0.6200} & \textbf{0.3363} & 0.2201 & 0.1828 & 0.0934
                 & 0.4270 & 0.5472 & 0.5037 & 0.4220 & 0.2349
                 & 0.4393 & \sbest{0.5817} & \sbest{0.4919} & 0.4040 & 0.2797
                 & 0.3288 & \textbf{0.6432} & 0.2988 & 0.2210 & 0.1524 \\
    PDRF~\cite{luo2026pdrf} & \sbest{0.3623} & 0.6818 & 0.5912 & 0.3289 & \sbest{0.2408} & \sbest{0.2110} & \sbest{0.1199}
                 & \sbest{0.4385} & \textbf{0.5525} & \sbest{0.5053} & \sbest{0.4285} & \sbest{0.2679}
                 & \sbest{0.4409} & 0.5681 & 0.4856 & \sbest{0.4066} & \sbest{0.3032}
                 & \sbest{0.3329} & 0.6277 & 0.2938 & \sbest{0.2381} & \sbest{0.1720} \\
    \rowcolor{oursrow}Ours$^{\dagger}$ & \textbf{0.3706} & 0.6815 & 0.5974 & \sbest{0.3341} & \textbf{0.2477} & \textbf{0.2173} & \textbf{0.1458}
                 & \textbf{0.4470} & \sbest{0.5482} & \textbf{0.5058} & \textbf{0.4369} & \textbf{0.2971}
                 & \textbf{0.4714} & \textbf{0.5933} & \textbf{0.5166} & \textbf{0.4369} & \textbf{0.3387}
                 & \textbf{0.3507} & \sbest{0.6317} & \textbf{0.3172} & \textbf{0.2598} & \textbf{0.1941} \\
    \hline
  \end{tabular}%
  }

  \vspace{1mm}
  {\footnotesize $^{\dagger}$Ours is the decoded ensemble with $\mathrm{thr}$ at the highest tier of each benchmark (SEVIR: VIL $\geq219$, MeteoNet: $\geq32$~dBZ, Shanghai and CIKM: $\geq40$~dBZ).}
\end{table*}

\section{On the Evaluation Resolution}
\label{app:resolution}

We evaluate at two resolutions, the native $256\times256$ output and the $128\times128$ protocol of prior radar benchmarks.

\noindent\textbf{Use of the two resolutions.} Our model outputs $256\times256$ fields on the four radar benchmarks. All metrics in Table~\ref{tab:radar-main} of the main text and Table~\ref{tab:app-csi} (CSI, HSS, SSIM, LPIPS) are evaluated at $128\times128$, the resolution at which DiffCast~\cite{yu2024diffcast}, AlphaPre~\cite{lin2025alphapre}, PercpCast~\cite{fengperceptually} and others report results on these benchmarks. Our forecasts and the ground truth are mapped to $[0,1]$, bilinearly downsampled to $128\times128$ and then evaluated. The baselines output $128\times128$ fields and are not resampled. On CIKM, whose native grid is $101\times101$, all methods including ours are evaluated on $101\times101$ fields zero-padded to $128\times128$. The CSI and HSS in Table~\ref{tab:abl-branch} follow the same $128\times128$ protocol. Its MAE, RMSE, CRPS, SSIM and LPIPS are computed on our native $256\times256$ output, with MAE and RMSE in the original units of each dataset. That table contains no baseline, so these metrics need not follow the $128\times128$ protocol.

\noindent\textbf{Numerical comparison under the two resolutions.} Table~\ref{tab:resolution} gives our results under both protocols. The conclusions do not depend on the resolution. On all four datasets decoding improves CSI-M and HSS, and the gain concentrates in the highest tier. The absolute values shift systematically. Compared with $256\times256$, the $128\times128$ protocol gives a higher HSS on all four datasets, a higher CSI-M on all but CIKM and a lower LPIPS, while SSIM rises on MeteoNet and falls on the other three, most on CIKM ($0.706 \to 0.675$). Values from the two protocols are therefore not interchangeable.

\begin{table*}[htbp]
  \centering
  \small
  \caption{Our results under the two evaluation resolutions (single model $\to$ ensemble). The ensemble setting is that of Table~\ref{tab:abl-branch}, and the samples and forecast setting are those of Table~\ref{tab:app-data}. $256\times256$ is the native output and $128\times128$ the protocol of prior work.}
  \label{tab:resolution}
  \resizebox{\ifdim\width>\textwidth\textwidth\else\width\fi}{!}{%
  \begin{tabular*}{\textwidth}{@{\extracolsep{\fill}}llcccc}
    \hline
    Metric & Protocol & \textbf{SEVIR} & \textbf{MeteoNet} & \textbf{Shanghai} & \textbf{CIKM} \\
    \hline
    \multirow{2}{*}{CSI-M$\uparrow$}
      & $256\times256$ & $0.3484 \to 0.3579$ & $0.4081 \to 0.4153$ & $0.4456 \to 0.4509$ & $0.3482 \to 0.3525$ \\
      & $128\times128$ & $0.3604 \to 0.3706$ & $0.4407 \to 0.4470$ & $0.4668 \to 0.4714$ & $0.3451 \to 0.3507$ \\
    \hline
    \multirow{2}{*}{HSS$\uparrow$}
      & $256\times256$ & $0.4531 \to 0.4659$ & $0.5454 \to 0.5550$ & $0.5835 \to 0.5901$ & $0.4268 \to 0.4329$ \\
      & $128\times128$ & $0.4664 \to 0.4800$ & $0.5779 \to 0.5860$ & $0.6041 \to 0.6098$ & $0.4468 \to 0.4546$ \\
    \hline
    \multirow{2}{*}{SSIM$\uparrow$}
      & $256\times256$ & $0.7019 \to 0.7061$ & $0.8365 \to 0.8385$ & $0.8346 \to 0.8350$ & $0.7066 \to 0.7057$ \\
      & $128\times128$ & $0.6926 \to 0.6978$ & $0.8439 \to 0.8451$ & $0.8218 \to 0.8222$ & $0.6757 \to 0.6754$ \\
    \hline
    \multirow{2}{*}{LPIPS$\downarrow$}
      & $256\times256$ & $0.2361 \to 0.2346$ & $0.1528 \to 0.1527$ & $0.1723 \to 0.1722$ & $0.3548 \to 0.3544$ \\
      & $128\times128$ & $0.1671 \to 0.1670$ & $0.0993 \to 0.1000$ & $0.1300 \to 0.1302$ & $0.2703 \to 0.2691$ \\
    \hline
  \end{tabular*}%
  }
\end{table*}
\section{Anchor versus Decoded Ensemble on All Six Datasets}
\label{subsec:abl-branch}
\begin{table}[htbp]
  \centering
  \small
  \caption{Anchor (single model, $1.0\times$) $\to$ decoded ensemble ($5.2\times$) on the Swedish precipitation dataset and MRMS, paired on the same anchor ($16$ frames, $80$-minute lead time). Thresholds in mm/h. The better value of each pair is in bold. This table gives the exact values behind Tab.~\ref{tab:decoding-gain} of the main text.}
  \label{tab:abl-main2}
  \setlength{\tabcolsep}{3pt}
  \resizebox{\ifdim\width>\linewidth\linewidth\else\width\fi}{!}{%
  \begin{tabular*}{\textwidth}{@{\extracolsep{\fill}}lcc}
    \hline
     & \textbf{Swedish} & \textbf{MRMS} \\
    \hline
    CSI$\uparrow$, highest & ($6.3$)~$0.0984 \to \mathbf{0.1357}$ & ($8$)~$0.0278 \to \mathbf{0.0360}$ \\
    CSI$\uparrow$, weak & ($0.06$)~$0.5926 \to \mathbf{0.5927}$ & ($1$)~$0.4346 \to \mathbf{0.4347}$ \\
    FSS$\uparrow$ & $\mathbf{0.9149} \to 0.9148$ & $\mathbf{0.7052} \to 0.7034$ \\
    SSIM$\uparrow$ & $\mathbf{0.7467} \to 0.7466$ & $\mathbf{0.8985} \to 0.8983$ \\
    CRPS$\downarrow$ & $\mathbf{0.1159} \to 0.1238$ & $\mathbf{0.1360} \to 0.1364$ \\
    LPIPS$\downarrow$ & $0.1897 \to \mathbf{0.1896}$ & $\mathbf{0.1306} \to 0.1308$ \\
    \hline
  \end{tabular*}%
  }
\end{table}

Table~\ref{tab:abl-main2} compares the anchor alone with the decoded ensemble on the Swedish precipitation dataset and MRMS, and Table~\ref{tab:abl-branch} does so on the official test sets of SEVIR, MeteoNet, Shanghai and CIKM. The ensemble uses $s=4$, $N=8$, $\lambda=1$, $52$ network evaluations in total ($5.2\times$), with $\mathrm{thr}$ at the highest tier of each dataset. The single-model baseline is the anchor of that ensemble, the field that support-consensus decoding starts from. It costs the $4$ trunk steps plus $6$ branch steps, $10$ steps in total ($1.0\times$). Both share the same anchor field and differ only in whether the $k$-of-$N$ vote corrects it, so the comparison is paired and the anchor's own sampling noise cancels. Table~\ref{tab:abl-branch} gives the full results. Since the thresholds differ across datasets, its CSI rows are ordered by intensity tier from highest to lowest.

\begin{table*}[htbp]
  \centering
  \small
  \caption{Anchor alone versus decoded ensemble on the official test sets of the four radar benchmarks. Each cell reads ``anchor ($1.0\times$) $\to$ ensemble ($5.2\times$)'', with the better value in bold. CSI and HSS follow the $128\times128$ protocol of Table~\ref{tab:radar-main} of the main text and Table~\ref{tab:app-csi}. The threshold of each tier is given in parentheses (VIL $0$--$255$ for SEVIR, dBZ for the others), and HSS is averaged over thresholds. MAE, RMSE, CRPS, SSIM and LPIPS are computed on the native $256\times256$ output, since the table contains no baseline. MAE and RMSE are in the original units of each dataset and are not comparable across datasets. Appendix~\ref{app:resolution} compares the two resolutions.}
  \label{tab:abl-branch}
  \resizebox{\textwidth}{!}{%
  \begin{tabular}{lcccc}
    \hline
     & \textbf{SEVIR} ($8096$) & \textbf{MeteoNet} ($1310$) & \textbf{Shanghai} ($526$) & \textbf{CIKM} ($4000$) \\
    \hline
    \multicolumn{5}{l}{\textit{CSI$\uparrow$ by intensity level, highest to lowest ($128\times128$ protocol)}} \\
    Highest & (219)~$0.1202 \to \mathbf{0.1458}$ & (32)~$0.2747 \to \mathbf{0.2971}$ & (40)~$0.3217 \to \mathbf{0.3387}$ & (40)~$0.1744 \to \mathbf{0.1941}$ \\
    2nd highest & (181)~$0.2087 \to \mathbf{0.2173}$ & (24)~$0.4354 \to \mathbf{0.4369}$ & (35)~$0.4351 \to \mathbf{0.4369}$ & (35)~$0.2574 \to \mathbf{0.2598}$ \\
    3rd & (160)~$0.2405 \to \mathbf{0.2477}$ & (18)~$0.5049 \to \mathbf{0.5058}$ & (30)~$\mathbf{0.5171} \to 0.5166$ & (30)~$0.3166 \to \mathbf{0.3172}$ \\
    4th & (133)~$0.3258 \to \mathbf{0.3341}$ & (12)~$0.5479 \to \mathbf{0.5482}$ & (20)~$0.5933 \to 0.5933$ & (20)~$\mathbf{0.6321} \to 0.6317$ \\
    5th & ~(74)~$0.5901 \to \mathbf{0.5974}$ & --- & --- & --- \\
    6th & ~(16)~$0.6774 \to \mathbf{0.6815}$ & --- & --- & --- \\
    CSI-M (multi-threshold avg.) & $0.3604 \to \mathbf{0.3706}$ & $0.4407 \to \mathbf{0.4470}$ & $0.4668 \to \mathbf{0.4714}$ & $0.3451 \to \mathbf{0.3507}$ \\
    \hline
    \multicolumn{5}{l}{\textit{Skill score ($128\times128$ protocol)}} \\
    HSS$\uparrow$ & $0.4664 \to \mathbf{0.4800}$ & $0.5779 \to \mathbf{0.5860}$ & $0.6041 \to \mathbf{0.6098}$ & $0.4468 \to \mathbf{0.4546}$ \\
    \hline
    \multicolumn{5}{l}{\textit{Pixel-level and perceptual metrics ($256\times256$)}} \\
    MAE$\downarrow$   & $9.158 \to \mathbf{8.976}$ & $1.183 \to \mathbf{1.177}$ & $\mathbf{1.6220} \to 1.6230$ & $\mathbf{4.7930} \to 4.8038$ \\
    RMSE$\downarrow$  & $18.374 \to \mathbf{18.014}$ & $\mathbf{3.460} \to 3.466$ & $\mathbf{5.1052} \to 5.1131$ & $\mathbf{7.3979} \to 7.4176$ \\
    CRPS$\downarrow$  & $0.0359 \to \mathbf{0.0352}$ & $0.01314 \to \mathbf{0.01307}$ & $\mathbf{0.01802} \to 0.01803$ & $\mathbf{0.05990} \to 0.06005$ \\
    SSIM$\uparrow$    & $0.7019 \to \mathbf{0.7061}$ & $0.8365 \to \mathbf{0.8385}$ & $0.8346 \to \mathbf{0.8350}$ & $\mathbf{0.7066} \to 0.7057$ \\
    LPIPS$\downarrow$ & $0.2361 \to \mathbf{0.2346}$ & $0.1528 \to \mathbf{0.1527}$ & $0.1723 \to \mathbf{0.1722}$ & $0.3548 \to \mathbf{0.3544}$ \\
    \hline
  \end{tabular}%
  }
\end{table*}

\noindent\textbf{Breadth of the gain.} The improvements are spread across intensity tiers. Of the $22$ CSI entries over the four datasets ($18$ tiers and $4$ CSI-M values), $19$ improve, $1$ ties and $2$ regress, each by at most $0.0005$ (Shanghai at $30$~dBZ and CIKM at $20$~dBZ). HSS improves on all four datasets. Decoding thus causes no material loss at any tier.

\noindent\textbf{Concentration of the gain.} The size of the gain concentrates in the highest tier. Gains in the lower tiers are mostly $0.0003$ to $0.0086$, while the highest tier gains $0.0170$ (Shanghai) to $0.0256$ (SEVIR). Since CSI-M is the mean of the per-threshold CSIs, the share of each tier can be read off directly. On MeteoNet, Shanghai and CIKM the highest tier gains $0.017$ to $0.022$ while every other tier changes by at most $0.0024$, so the highest tier accounts for $88\%$ to $92\%$ of the CSI-M increment. SEVIR also improves by $0.004$ to $0.009$ at all five lower tiers, and its highest tier accounts for $42\%$.

\noindent\textbf{Pixel-level cost.} At the pixel level the anchor and the ensemble differ little and not in a consistent direction. Both deliver one field, so CRPS equals the MAE on fields normalized to $[0,1]$ and the two always move together. On SEVIR MAE and RMSE improve by about $2\%$, on MeteoNet MAE improves and RMSE is flat, and on Shanghai and CIKM both worsen slightly. No regression exceeds $0.27\%$ (the largest is RMSE on CIKM, $7.3979 \to 7.4176$). Over the same experiments the highest-tier CSI improves by $5.3\%$ (Shanghai) to $21.3\%$ (SEVIR), $20$ to $80$ times more. The correction is thus local. It raises the heavy-echo hit rate and leaves the rest of the forecast field nearly unchanged.

\noindent\textbf{Perceptual quality.} At the native resolution, LPIPS improves on all four datasets, and SSIM improves everywhere except a $0.0009$ drop on CIKM. This follows from the construction of the decoder. Added pixels take values from an actual member and removed pixels are only pushed below the threshold, so raising the heavy-echo hit rate introduces no blur. On the four radar benchmarks decoding therefore raises the heavy-echo hit rate at a small pixel-level cost and without loss of perceptual quality. On the Swedish precipitation dataset the pixel-level cost is larger (CRPS $+6.8\%$), as analyzed in Sec.~\ref{sec:results} of the main text.

\section{Per-Dataset Results}
\label{app:sweden-extra}
\label{app:radar-leadtime}
This section reports, for each of the six datasets, the skill versus lead time and two forecast examples, one subsection per dataset. The two precipitation datasets (Swedish and MRMS) also get the full comparison table. Both forecast $16$ frames ($5$ to $80$~min) from $4$ input frames and are evaluated at the native $256\times256$ resolution. Ours is the decoded ensemble ($5.2\times$).

On the four radar benchmarks, Figures~\ref{fig:lt-sevir}--\ref{fig:lt-cikm} plot each metric against the forecast frame, under the same $128\times128$ protocol as Table~\ref{tab:app-csi}. CSI is shown at every intensity threshold of each dataset. HSS, POD and FAR are shown averaged over the thresholds and at the highest threshold. For MeteoNet, Shanghai and CIKM, which have fewer thresholds, MAE and RMSE are added. Ours is the decoded ensemble. CRPS, MAE and RMSE are outside the benchmark protocol, and for the baselines they appear only in these figures. For the other metrics, the frame means equal the corresponding entries of Tables~\ref{tab:radar-main} and~\ref{tab:app-csi}.

\newcommand{\ltpanel}[2]{%
  \begin{subfigure}[t]{0.2\textwidth}
    \centering
    \includegraphics[width=\linewidth]{#1}
    \caption{#2}
  \end{subfigure}}

Across the four benchmarks two patterns hold. At the highest threshold our CSI is above all baselines at every frame, with a FAR that never rises above the range spanned by the baselines, and on SEVIR and CIKM the decoder keeps forecasting heavy echoes after most baselines have stopped forecasting them. At the lowest thresholds the strong methods stay within a few hundredths of each other over the whole horizon, so the differences between methods sit in the heavy tail, which is where the decoder acts. Each benchmark is discussed in its own subsection below, with its curves and two forecast examples.

\subsection{Swedish Precipitation}
\label{app:sweden}
\begin{table}[H]
  \centering
  \small
  \caption{Comparison on the Swedish precipitation dataset ($1000$ test samples, $4\to16$ frames, native $256\times256$, mean over the $16$ forecast frames). CSI thresholds are in mm/h. The CSI, SSIM and LPIPS columns repeat Table~\ref{tab:AvG}. Best in bold, second best underlined.}
  \label{tab:sw-full}
  \footnotesize\renewcommand{\arraystretch}{0.9}
  \begin{tabular}{l|cccccc}
    \hline
    Method & CSI$_{0.06}\uparrow$ & CSI$_{6.3}\uparrow$ & FSS$\uparrow$ & CRPS$\downarrow$ & SSIM$\uparrow$ & LPIPS$\downarrow$ \\
    \hline
    PySteps~\cite{pulkkinen2019pysteps} & 0.473 & 0.049 & 0.872 & 0.144 & 0.589 & 0.284 \\
    DGMR~\cite{ravuri2021skilful} & 0.433 & 0.039 & 0.882 & 0.143 & 0.702 & 0.204 \\
    DiffCast~\cite{yu2024diffcast} & 0.439 & 0.029 & 0.903 & 0.149 & 0.660 & 0.292 \\
    SSLDM-ISI~\cite{ling2024spacetime} & 0.558 & 0.040 & 0.906 & 0.122 & 0.718 & 0.191 \\
    AlphaPre~\cite{lin2025alphapre} & 0.541 & 0.025 & 0.894 & 0.149 & 0.647 & 0.333 \\
    DTCA~\cite{li2024precipitation} & 0.572 & 0.046 & 0.911 & 0.115 & \sbest{0.732} & \textbf{0.180} \\
    RainHCNet~\cite{wang2025rainhcnet} & 0.499 & 0.025 & 0.386 & 0.153 & 0.332 & 0.320 \\
    SSRF-Net~\cite{luo2025ssrf} & 0.485 & 0.045 & 0.402 & 0.222 & 0.572 & 0.310 \\
    exPreCast~\cite{song2026extreme} & 0.489 & 0.028 & 0.917 & 0.132 & 0.520 & 0.348 \\
    FlowCast~\cite{ribeiro2026flowcast} & 0.482 & 0.009 & \textbf{0.927} & \textbf{0.109} & 0.680 & 0.324 \\
    SDIR~\cite{zhou2026sdir} & 0.424 & 0.019 & 0.922 & \sbest{0.112} & 0.718 & 0.307 \\
    PDRF~\cite{luo2026pdrf} & 0.474 & 0.037 & 0.905 & 0.149 & 0.571 & 0.314 \\
    LangPrecip~\cite{ling2026langprecip} & \sbest{0.586} & \sbest{0.074} & \sbest{0.923} & 0.132 & \sbest{0.732} & 0.193 \\
    \rowcolor{oursrow}Ours & \textbf{0.593} & \textbf{0.136} & 0.915 & 0.124 & \textbf{0.747} & \sbest{0.190} \\
    \hline
  \end{tabular}
\end{table}

Table~\ref{tab:sw-full} completes Table~\ref{tab:AvG} with FSS and CRPS. Ours is best in both CSI columns and in SSIM and second in LPIPS. FlowCast is best in FSS and CRPS, where ours ranks fifth. Sec.~\ref{sec:results} of the main text analyzes this gap.

\subsubsection{Temporal Analysis}
\label{app:sweden-temporal}

\begin{figure}[H]
  \centering
  \captionsetup[sub]{skip=1pt}
  \includegraphics[width=0.88\textwidth]{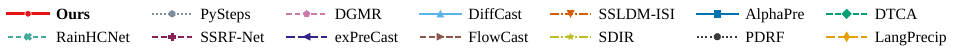}\par\vspace{2mm}
  \begin{subfigure}[t]{0.2\textwidth}\centering\includegraphics[width=\linewidth]{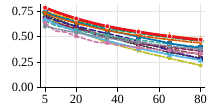}\caption{CSI ($\geq 0.06$~mm/h) $\uparrow$}\label{fig:sw_csi_light}\end{subfigure}\hspace{0.025\textwidth}
  \begin{subfigure}[t]{0.2\textwidth}\centering\includegraphics[width=\linewidth]{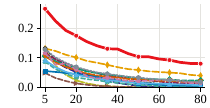}\caption{CSI ($\geq 6.3$~mm/h) $\uparrow$}\label{fig:sw_csi_heavy}\end{subfigure}\hspace{0.025\textwidth}
  \begin{subfigure}[t]{0.2\textwidth}\centering\includegraphics[width=\linewidth]{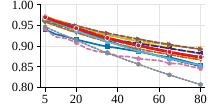}\caption{FSS $\uparrow$}\label{fig:sw_fss}\end{subfigure}

  \vspace{1mm}
  \begin{subfigure}[t]{0.2\textwidth}\centering\includegraphics[width=\linewidth]{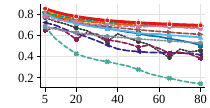}\caption{SSIM $\uparrow$}\label{fig:sw_ssim}\end{subfigure}\hspace{0.025\textwidth}
  \begin{subfigure}[t]{0.2\textwidth}\centering\includegraphics[width=\linewidth]{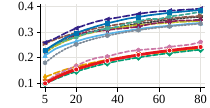}\caption{LPIPS $\downarrow$}\label{fig:sw_lpips}\end{subfigure}\hspace{0.025\textwidth}
  \begin{subfigure}[t]{0.2\textwidth}\centering\includegraphics[width=\linewidth]{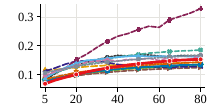}\caption{CRPS $\downarrow$}\label{fig:sw_crps}\end{subfigure}

  \caption{Skill versus lead time on the Swedish precipitation dataset ($4\to16$ frames, $1000$ test samples, native $256\times256$). The horizontal axis of every panel is the lead time in minutes. Ours is the decoded ensemble ($5.2\times$). The FSS panel is cut at $0.8$, so RainHCNet and SSRF-Net, whose FSS stays below this value, are not visible there.}
  \label{fig:sw_temporal}
\end{figure}

At $6.3$~mm/h our CSI is above every baseline at every lead time. It falls from $0.266$ at $5$~min to $0.080$ at $80$~min, against $0.132$ and $0.040$ for LangPrecip, the strongest baseline at every lead time. The ratio between the two stays between $1.7$ and $2.0$ over the whole horizon, so the margin does not narrow with lead time. Seven of the thirteen baselines fall below CSI $0.01$ at this threshold, first FlowCast at $25$~min and SDIR at $30$~min, last DiffCast at $75$~min. At $0.06$~mm/h our CSI leads LangPrecip by $0.034$ at $5$~min. The two are level at $45$~min, ours trails by at most $0.007$ afterwards, and both stay above all other baselines at every lead time.

FSS is the metric where ours trails most. It starts within $0.003$ of the best value (SDIR), and the gap to the best baseline grows to about $0.02$ at $80$~min, where FlowCast and LangPrecip lead. Our SSIM is the highest of all methods up to $35$~min and stays within $0.005$ of the highest afterwards. Our LPIPS is second to DTCA up to $45$~min and stays within $0.011$ of it over the whole horizon. Our CRPS is the lowest at $5$ and $10$~min and then grows by $0.084$ over the horizon, against $0.043$ for FlowCast and $0.056$ for SDIR. FlowCast has the lowest CRPS from $20$~min on ($0.125$ against our $0.153$ at $80$~min). As on CIKM (\S\ref{app:lt-cikm-temporal}), this is the double penalty that pixel errors assign to sharp heavy cores once their position drifts.

\clearpage
\subsubsection{Visualization}
\label{app:sweden-vis}
\begin{figure}[H]
  \centering
  \begin{subfigure}[t]{0.495\textwidth}\centering\includegraphics[width=0.88\linewidth]{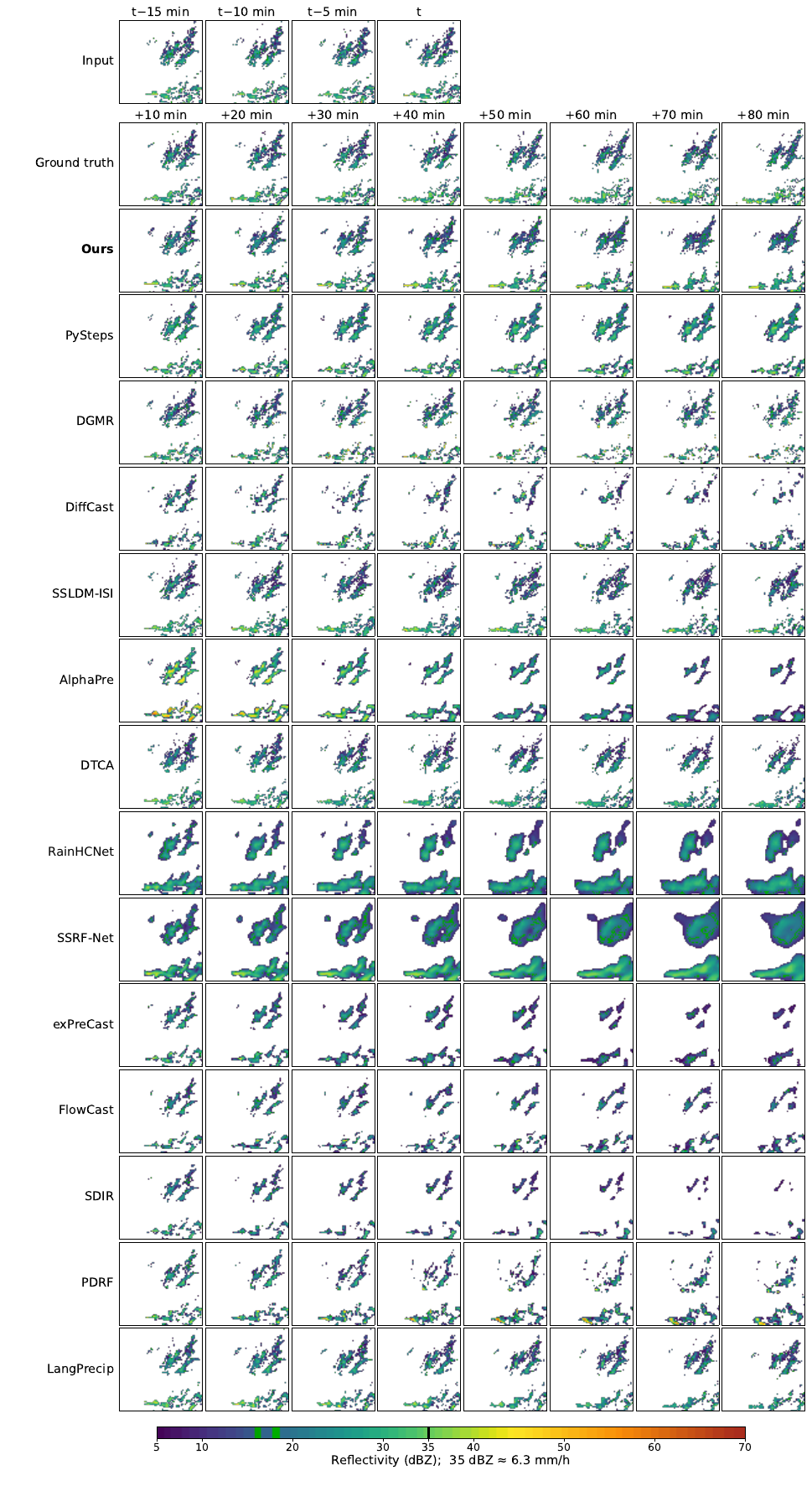}\caption{Case 1}
    \vspace{2pt}\par{\scriptsize\raggedright\textbf{Predicted scenario $\hat{c}$:} \emph{The radar echoes drift steadily rightward with a subtle downward tilt, tracing a smooth trajectory across the frame as fragmented, irregular blobs---concentrated in the upper right and lower left---slide cohesively without breaking apart or merging. Their diffuse, filamentous edges remain unchanged as the entire pattern glides uniformly, maintaining internal voids and structural integrity while advancing at a constant pace. No deformation or reorientation emerges; the motion is purely translational, with all elements locked in synchronized drift, preserving their relative positions and shapes throughout the sequence. The system evolves not through transformation but through persistent, linear displacement---right and slightly down---its form frozen in motion, its dynamics reduced to unwavering advection across the screen.}\par}
  \end{subfigure}\hfill
  \begin{subfigure}[t]{0.495\textwidth}\centering\includegraphics[width=0.88\linewidth]{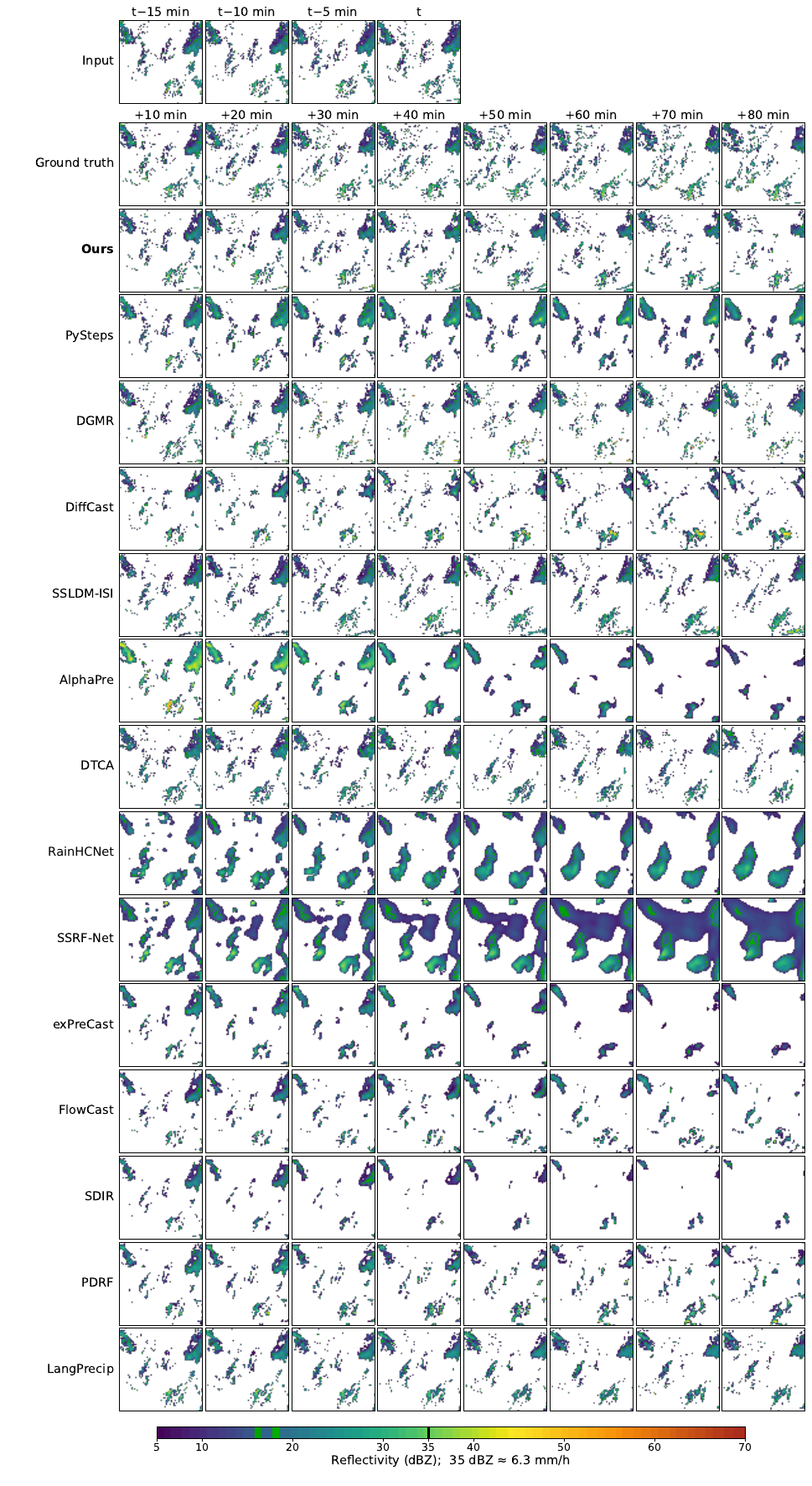}\caption{Case 2}
    \vspace{2pt}\par{\scriptsize\raggedright\textbf{Predicted scenario $\hat{c}$:} \emph{The radar echoes drift steadily rightward across the frame in a near-straight trajectory, maintaining their fragmented, blob-like forms with soft, undefined edges and scattered internal voids. As they slide uniformly to the right, the entire field undergoes subtle deformation, with no curvature or reorientation emerging. The motion remains consistent in speed and direction, with no acceleration or deviation, while the shapes neither merge nor break apart, preserving their irregular, disorganized layout throughout. No new structures form, and no existing ones dissolve; the system simply translates intact, its evolution defined purely by lateral displacement and gentle contour change, devoid of dynamic reorganization or structural transformation.}\par}
  \end{subfigure}
  \caption{Forecast examples on the Swedish precipitation dataset. Each panel shows the four input frames, the ground truth and the forecast of every compared method at every second forecast frame ($+10$ to $+80$~min). Below each panel is the scenario $\hat{c}$ that the fine-tuned VLM predicts from the input frames alone.}
  \label{fig:qv-sweden}
\end{figure}

Case~1 (Fig.~\ref{fig:qv-sweden}a) contains two systems. A group of fragmented cells in the north-east, mostly below $30$~dBZ, drifts north-east, and a band along the southern edge, whose western end holds cores above $35$~dBZ, drifts east while its heavy cores spread along it. Ours keeps a heavy core at the western end of the southern band through $+80$~min, together with fewer cores above $35$~dBZ further east than observed. The northern group keeps its position and orientation and from $+50$~min merges into one patch with fewer internal gaps than observed. DTCA and LangPrecip keep the fine texture of both systems, and the cores of their southern band fall mostly below $35$~dBZ after about $+50$~min. PySteps and AlphaPre forecast smooth blobs, AlphaPre with cores above $45$~dBZ in the first frames that decay below $30$~dBZ by $+50$~min. exPreCast and SDIR shrink the echo area over the horizon, and SDIR keeps only a few weak fragments at $+80$~min. RainHCNet and SSRF-Net over-smooth the echoes into broad, weak areas, while PDRF forecasts a speckled field with isolated cores above $45$~dBZ.

Case~2 consists of scattered showers over the whole domain, with stratiform areas in the north-west and north-east and a cluster of convective cells in the south whose cores stay above $35$~dBZ through $+80$~min. Over the horizon the small cells in the middle of the domain multiply. Ours keeps the southern cluster at the observed position with cores above $35$~dBZ through $+80$~min, and it keeps the north-western area. It under-predicts the convective cells in the middle after about $+40$~min, missing the initiation of new cells, and its north-eastern area contracts faster than observed. LangPrecip and DTCA keep the texture of the field, with southern cores mostly below $35$~dBZ after about $+40$~min. DiffCast intensifies the southern cluster above $45$~dBZ from $+50$~min on. PySteps and AlphaPre forecast smooth blobs, PySteps with a core that grows in the north-eastern area. exPreCast and SDIR lose most echoes in the middle of the domain, and RainHCNet and SSRF-Net over-smooth the cells into broad, contiguous areas.

\begin{figure}[H]
  \centering
  \includegraphics[width=0.95\textwidth]{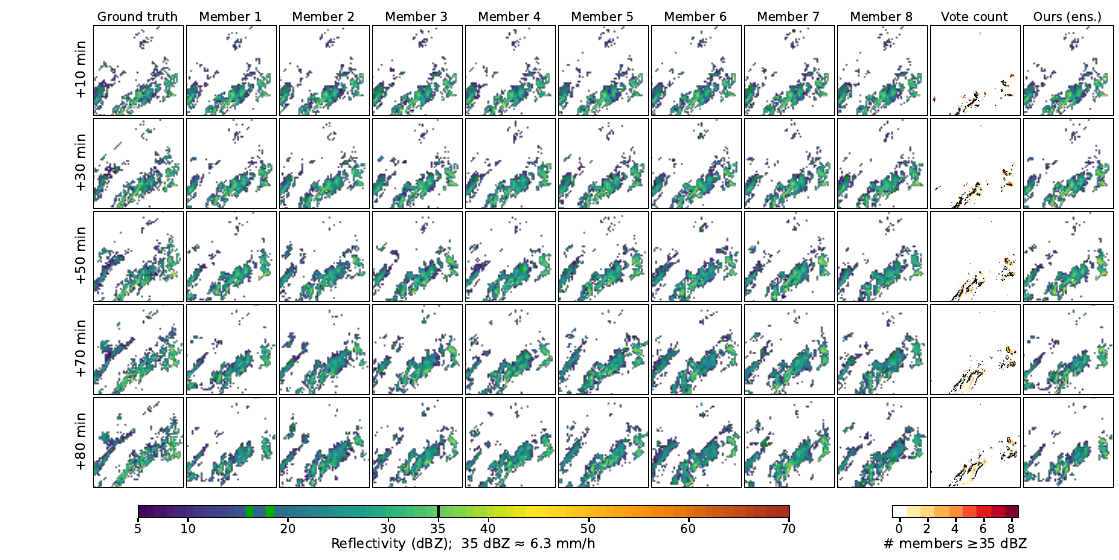}
  \caption{\textbf{All eight members of the branched ensemble} in Fig.~\ref{fig:members} of the main text at five lead times ($s^*=4$, $N=8$). The members share the position, orientation and motion of the rain band and differ in the location and extent of the heavy cores. \emph{Vote count}: number of members at or above $35$~dBZ ($\approx6.3$~mm/h) at each pixel. \emph{Ours (ens.)}: the field delivered by support-consensus decoding.}
  \label{fig:members-all}
\end{figure}

Figure~\ref{fig:members-all} shows all eight members behind the forecast of a third Swedish case, the case of Fig.~\ref{fig:members} in the main text.

\clearpage
\subsection{MRMS}
\label{app:mrms}
\begin{table}[H]
  \centering
  \small
  \caption{Comparison on MRMS ($4\to16$ frames, native $256\times256$, mean over the $16$ forecast frames). CSI thresholds are in mm/h. LangPrecip uses CFG scale $4$. The CSI, SSIM and LPIPS columns repeat Table~\ref{tab:AvG}. Best in bold, second best underlined.}
  \label{tab:mrms-full}
  \footnotesize\renewcommand{\arraystretch}{0.9}
  \begin{tabular}{l|cccccc}
    \hline
    Method & CSI$_{1}\uparrow$ & CSI$_{8}\uparrow$ & FSS$\uparrow$ & CRPS$\downarrow$ & SSIM$\uparrow$ & LPIPS$\downarrow$ \\
    \hline
    PySteps~\cite{pulkkinen2019pysteps} & 0.338 & 0.024 & 0.579 & 0.167 & 0.871 & 0.187 \\
    DGMR~\cite{ravuri2021skilful} & 0.367 & 0.022 & 0.633 & 0.154 & 0.883 & 0.137 \\
    DiffCast~\cite{yu2024diffcast} & 0.328 & 0.018 & 0.613 & 0.168 & 0.869 & 0.183 \\
    SSLDM-ISI~\cite{ling2024spacetime} & 0.416 & 0.024 & 0.704 & 0.135 & 0.897 & \sbest{0.120} \\
    AlphaPre~\cite{lin2025alphapre} & 0.360 & 0.024 & 0.699 & 0.144 & 0.891 & 0.197 \\
    DTCA~\cite{li2024precipitation} & \sbest{0.423} & 0.026 & \sbest{0.711} & 0.131 & 0.899 & \textbf{0.118} \\
    RainHCNet~\cite{wang2025rainhcnet} & 0.320 & 0.002 & 0.606 & 0.153 & 0.877 & 0.207 \\
    SSRF-Net~\cite{luo2025ssrf} & 0.336 & 0.009 & 0.573 & 0.219 & 0.844 & 0.213 \\
    exPreCast~\cite{song2026extreme} & 0.394 & 0.015 & 0.693 & 0.190 & 0.828 & 0.203 \\
    FlowCast~\cite{ribeiro2026flowcast} & 0.393 & 0.013 & 0.699 & 0.133 & 0.900 & 0.192 \\
    SDIR~\cite{zhou2026sdir} & 0.349 & 0.013 & 0.644 & \sbest{0.126} & \textbf{0.907} & 0.189 \\
    PDRF~\cite{luo2026pdrf} & 0.383 & 0.020 & 0.670 & 0.166 & 0.875 & 0.190 \\
    LangPrecip~\cite{ling2026langprecip} & \textbf{0.435} & \sbest{0.031} & \textbf{0.725} & \textbf{0.125} & \sbest{0.905} & 0.142 \\
    \rowcolor{oursrow}Ours & \textbf{0.435} & \textbf{0.036} & 0.703 & 0.136 & 0.898 & 0.131 \\
    \hline
  \end{tabular}
\end{table}

Table~\ref{tab:mrms-full} completes Table~\ref{tab:AvG} with FSS and CRPS. Ours ties LangPrecip for the best CSI at $1$~mm/h and has the best CSI at $8$~mm/h. In FSS, CRPS, SSIM and LPIPS it ranks fourth, sixth, fifth and third.

\subsubsection{Temporal Analysis}
\label{app:mrms-temporal}

\begin{figure}[H]
  \centering
  \captionsetup[sub]{skip=1pt}
  \includegraphics[width=0.88\textwidth]{img/leadtime_sweden/legend.pdf}\par\vspace{2mm}
  \begin{subfigure}[t]{0.2\textwidth}\centering\includegraphics[width=\linewidth]{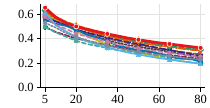}\caption{CSI ($\geq 1$~mm/h) $\uparrow$}\label{fig:mrms_csi_light}\end{subfigure}\hspace{0.025\textwidth}
  \begin{subfigure}[t]{0.2\textwidth}\centering\includegraphics[width=\linewidth]{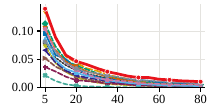}\caption{CSI ($\geq 8$~mm/h) $\uparrow$}\label{fig:mrms_csi_heavy}\end{subfigure}\hspace{0.025\textwidth}
  \begin{subfigure}[t]{0.2\textwidth}\centering\includegraphics[width=\linewidth]{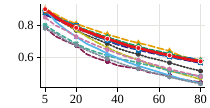}\caption{FSS $\uparrow$}\label{fig:mrms_fss}\end{subfigure}

  \vspace{1mm}
  \begin{subfigure}[t]{0.2\textwidth}\centering\includegraphics[width=\linewidth]{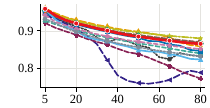}\caption{SSIM $\uparrow$}\label{fig:mrms_ssim}\end{subfigure}\hspace{0.025\textwidth}
  \begin{subfigure}[t]{0.2\textwidth}\centering\includegraphics[width=\linewidth]{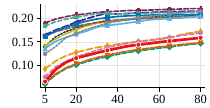}\caption{LPIPS $\downarrow$}\label{fig:mrms_lpips}\end{subfigure}\hspace{0.025\textwidth}
  \begin{subfigure}[t]{0.2\textwidth}\centering\includegraphics[width=\linewidth]{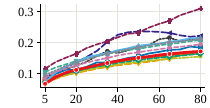}\caption{CRPS $\downarrow$}\label{fig:mrms_crps}\end{subfigure}

  \caption{Skill versus lead time on MRMS ($4\to16$ frames, native $256\times256$). The horizontal axis of every panel is the lead time in minutes. Ours is the decoded ensemble ($5.2\times$).}
  \label{fig:mrms_temporal}
\end{figure}

MRMS is the dataset where our absolute margin is smallest. At $8$~mm/h our CSI is the highest at every lead time, by $0.001$ to $0.007$ over the strongest baseline. It falls from $0.139$ at $5$~min to $0.010$ at $80$~min, against $0.136$ (LangPrecip) and $0.007$ (AlphaPre) for the strongest baseline at the two ends. The relative gain thus grows from $2\%$ at the first frame to between $27\%$ and $44\%$ from $55$~min on, at very small absolute values. Every baseline falls below CSI $0.01$ at this threshold, between $15$~min (RainHCNet) and $65$~min (PySteps), while ours stays at or above $0.010$. Heavy echoes are very sparse on MRMS, so the vote corrects few pixels. The absolute gain is small ($+0.008$) although the relative gain is large ($+29.5\%$, Tab.~\ref{tab:decoding-gain}). At $1$~mm/h ours leads at $5$~min ($0.654$ against $0.639$ for LangPrecip), trails LangPrecip by up to $0.009$ between $10$ and $30$~min, and stays within $0.004$ of it from $35$~min on.

LangPrecip has a higher FSS than ours at every lead time. The gap reaches $0.033$ at $55$~min and narrows to $0.008$ at $80$~min, as the FSS of LangPrecip drops faster in the last frames. In SSIM and CRPS ours is level with LangPrecip at $5$~min and falls behind the best baseline afterwards, by at most $0.015$ and $0.017$. SDIR is best in both from about $60$~min on. Our LPIPS is third at every lead time, behind DTCA and SSLDM-ISI and ahead of LangPrecip. On MRMS the decoder thus leaves the forecast close to LangPrecip in the area-based and continuous metrics and adds a small but consistent gain at the heavy threshold.

\clearpage
\subsubsection{Visualization}
\label{app:mrms-vis}
\begin{figure}[H]
  \centering
  \begin{subfigure}[t]{0.495\textwidth}\centering\includegraphics[width=0.88\linewidth]{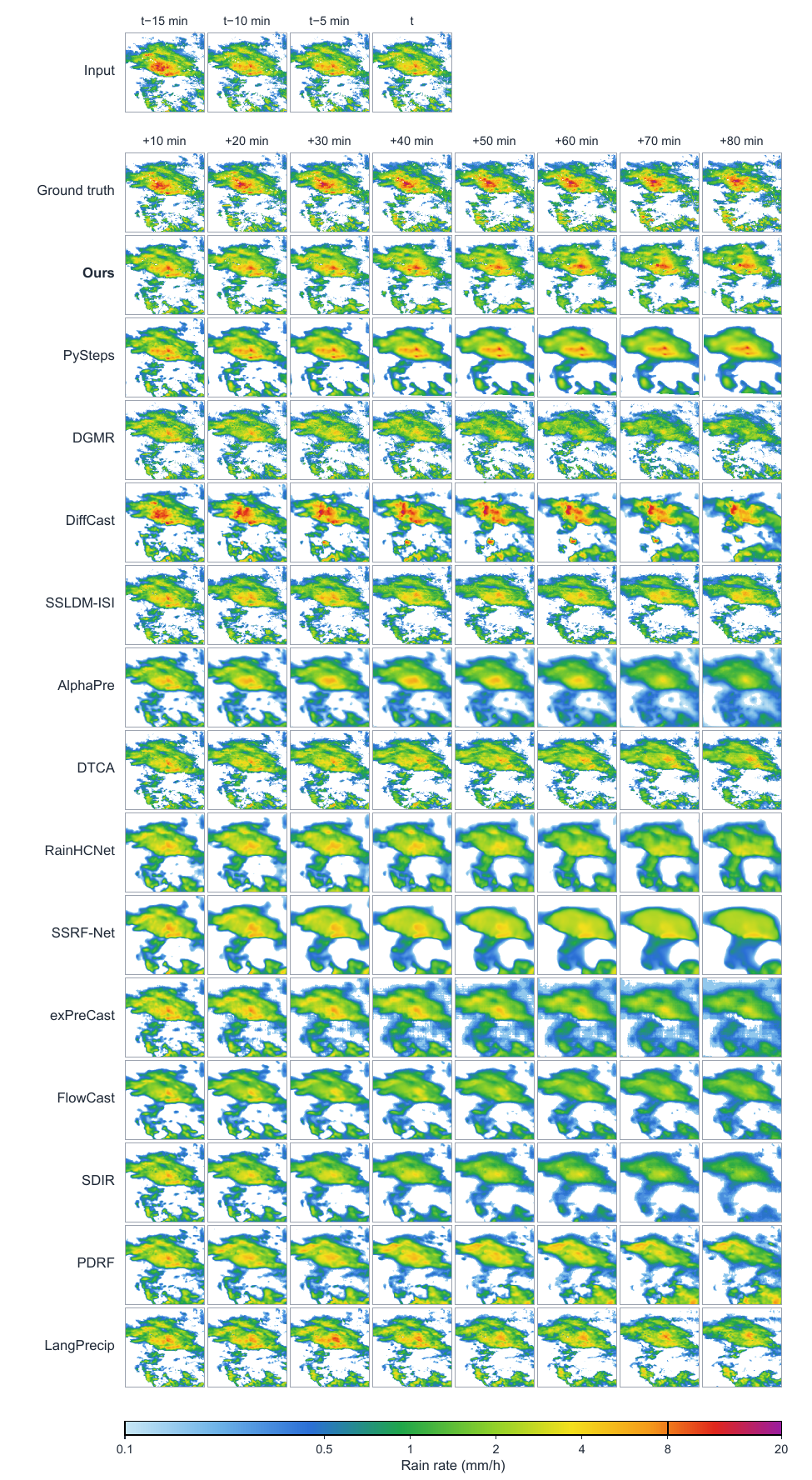}\caption{Case 1}
    \vspace{2pt}\par{\scriptsize\raggedright\textbf{Predicted scenario $\hat{c}$:} \emph{The radar echo structure drifts steadily rightward with a subtle upward tilt, tracing a near-straight trajectory across the frame as it maintains remarkable directional consistency. As it moves, the entire mass undergoes subtle curvature, its internal bright cores and dark voids reorienting in unison, while the whole system deforms---edges drawing closer together without fragmentation. The leading edge, once sharply defined, softens slightly as it advances, yet the core remains cohesive, resisting dispersion despite rapid translation. Brightness gradients shift subtly within, hinting at internal reorganization, but no new features emerge or break away; instead, the structure evolves through synchronized motion---sliding right, deforming, and compressing---all while preserving its elongated, irregular silhouette. This is a unified, dynamic entity: moving swiftly to the right, turning gently, and holding its form against the flow.}\par}
  \end{subfigure}\hfill
  \begin{subfigure}[t]{0.495\textwidth}\centering\includegraphics[width=0.88\linewidth]{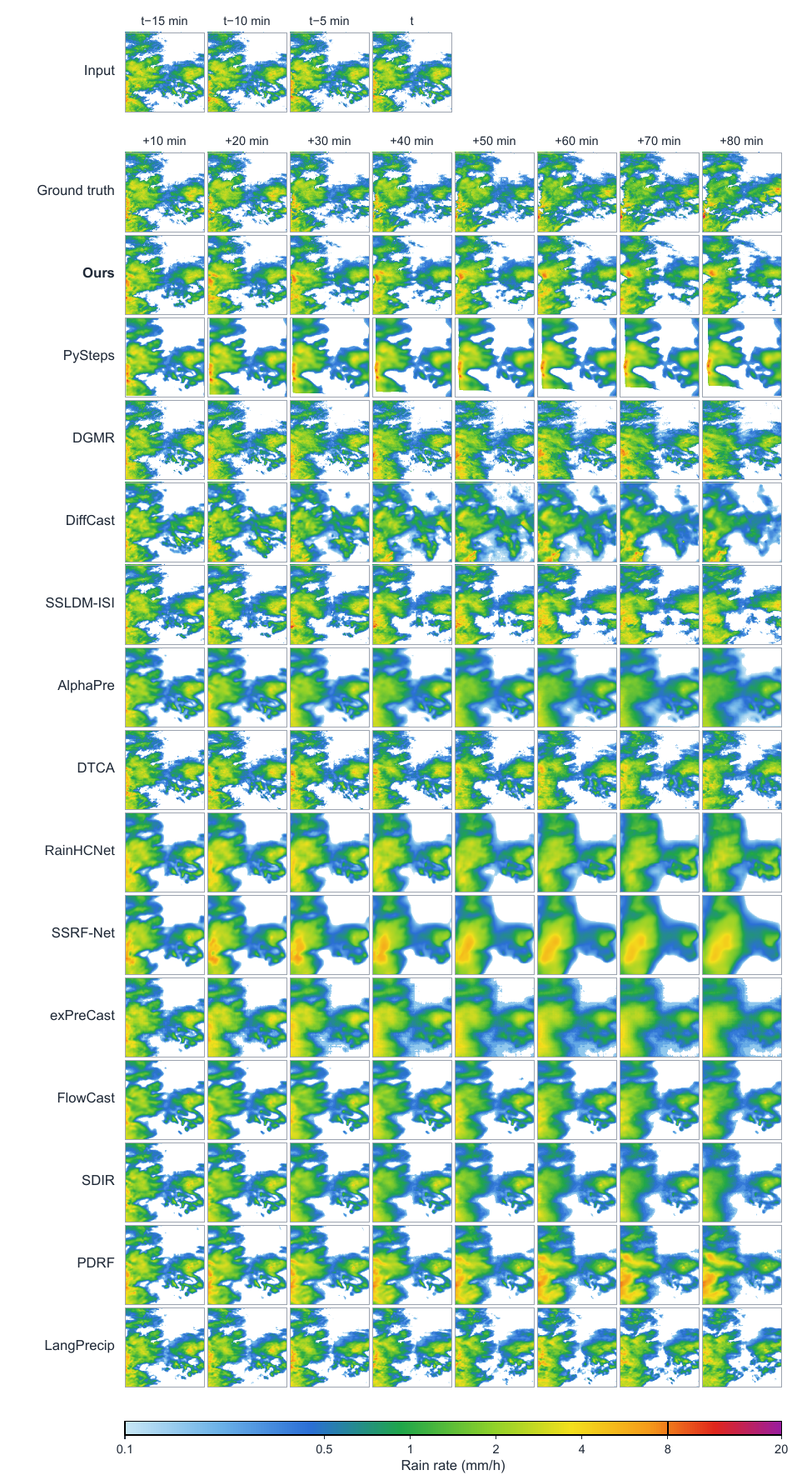}\caption{Case 2}
    \vspace{2pt}\par{\scriptsize\raggedright\textbf{Predicted scenario $\hat{c}$:} \emph{The radar echoes drift steadily rightward across the frame in a near-straight trajectory, maintaining their fragmented, blob-like form without coalescing or breaking apart. Scattered patches, concentrated in the upper-left and trailing toward the lower-right, slide uniformly as a loose cluster, their fuzzy edges never sharpening or organizing into lines or cores. Over time, the entire field glides smoothly to the right with no curvature or deformation---just pure translation, frame after frame, at a constant pace. Internal structure remains inert: no pulses, no splits, no surges---only passive advection carries the disorganized mass horizontally, preserving its shapeless, low-contrast character from start to finish. The motion is linear, unidirectional, and mechanically consistent, as if the whole scene were locked to an invisible conveyor belt moving rightward without turbulence or transformation.}\par}
  \end{subfigure}
  \caption{Forecast examples on MRMS. Each panel shows the four input frames, the ground truth and the forecast of every compared method at every second forecast frame ($+10$ to $+80$~min). Below each panel is the scenario $\hat{c}$ that the fine-tuned VLM predicts from the input frames alone.}
  \label{fig:qv-mrms}
\end{figure}

Case~1 (Fig.~\ref{fig:qv-mrms}a) is a rain shield in the northern half of the domain with an embedded core above $8$~mm/h near its centre, which persists over the whole horizon with a slight eastward drift, and scattered showers in the south. Ours keeps the outline of the shield and the textured showers in the south. Its core stays near the observed position, with a smaller area above $8$~mm/h than observed over the whole horizon. LangPrecip also keeps a textured shield, with a core around $8$~mm/h. DiffCast overestimates the core, with a broad area above $8$~mm/h and a spurious isolated cell in the south. DGMR, SSLDM-ISI and DTCA keep the texture and weaken the core below $8$~mm/h after the first frames. PySteps, AlphaPre, RainHCNet, SSRF-Net, FlowCast and SDIR forecast smooth fields, and the core of AlphaPre, FlowCast and SDIR decays below $4$~mm/h. exPreCast shows block artifacts and a spreading weak fringe, and PDRF forecasts a spurious intensification in the south-east.

Case~2 is widespread moderate rain of $1$ to $4$~mm/h over the western half of the domain with an eastern lobe. Rain above $8$~mm/h appears only in a narrow strip along the western edge, which intensifies after $+50$~min. Ours keeps the layout and texture of the field and forecasts a heavy patch near the western edge from about $+20$~min on, slightly displaced from the observed strip. The light echoes that fill the gap between the western area and the eastern lobe in the observation are partly missed. PySteps keeps a heavy strip at the western edge and leaves an empty region at the south-western boundary as the field is advected. LangPrecip, DGMR, SSLDM-ISI and DTCA keep the texture with few pixels above $8$~mm/h. DiffCast and exPreCast cover the south with a veil of light rain below $0.5$~mm/h. AlphaPre, RainHCNet and SSRF-Net blur the field, and SSRF-Net adds a spurious core of $4$ to $8$~mm/h. PDRF over-intensifies a broad western area to $4$ to $8$~mm/h after $+40$~min, and FlowCast and SDIR preserve the layout as an overly smooth field.

\subsection{SEVIR}
\label{app:lt-sevir}
\subsubsection{Temporal Analysis}
\label{app:lt-sevir-temporal}

\begin{figure}[H]
  \centering
  \captionsetup[sub]{skip=1pt}
  \includegraphics[width=0.88\textwidth]{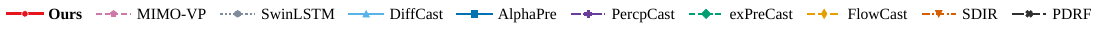}\par\vspace{0.5mm}
  \begin{subfigure}[t]{0.185\textwidth}\centering\includegraphics[width=\linewidth]{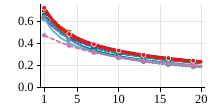}\caption{CSI-M $\uparrow$}\end{subfigure}\hspace{0.025\textwidth}
  \begin{subfigure}[t]{0.185\textwidth}\centering\includegraphics[width=\linewidth]{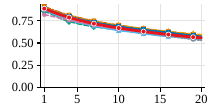}\caption{CSI ($\geq 16$) $\uparrow$}\end{subfigure}\hspace{0.025\textwidth}
  \begin{subfigure}[t]{0.185\textwidth}\centering\includegraphics[width=\linewidth]{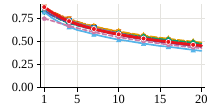}\caption{CSI ($\geq 74$) $\uparrow$}\end{subfigure}\hspace{0.025\textwidth}
  \begin{subfigure}[t]{0.185\textwidth}\centering\includegraphics[width=\linewidth]{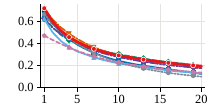}\caption{CSI ($\geq 133$) $\uparrow$}\end{subfigure}

  \vspace{0.3mm}
  \begin{subfigure}[t]{0.185\textwidth}\centering\includegraphics[width=\linewidth]{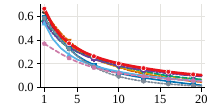}\caption{CSI ($\geq 160$) $\uparrow$}\end{subfigure}\hspace{0.025\textwidth}
  \begin{subfigure}[t]{0.185\textwidth}\centering\includegraphics[width=\linewidth]{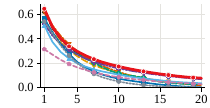}\caption{CSI ($\geq 181$) $\uparrow$}\end{subfigure}\hspace{0.025\textwidth}
  \begin{subfigure}[t]{0.185\textwidth}\centering\includegraphics[width=\linewidth]{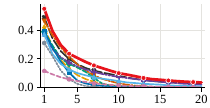}\caption{CSI ($\geq 219$) $\uparrow$}\end{subfigure}\hspace{0.025\textwidth}
  \begin{subfigure}[t]{0.185\textwidth}\centering\includegraphics[width=\linewidth]{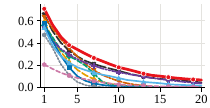}\caption{HSS ($\geq 219$) $\uparrow$}\end{subfigure}

  \vspace{0.3mm}
  \begin{subfigure}[t]{0.185\textwidth}\centering\includegraphics[width=\linewidth]{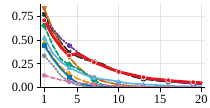}\caption{POD ($\geq 219$) $\uparrow$}\end{subfigure}\hspace{0.025\textwidth}
  \begin{subfigure}[t]{0.185\textwidth}\centering\includegraphics[width=\linewidth]{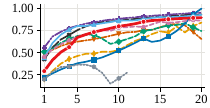}\caption{FAR ($\geq 219$) $\downarrow$}\end{subfigure}\hspace{0.025\textwidth}
  \begin{subfigure}[t]{0.185\textwidth}\centering\includegraphics[width=\linewidth]{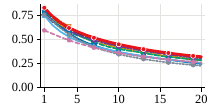}\caption{HSS (thr.\ avg.) $\uparrow$}\end{subfigure}\hspace{0.025\textwidth}
  \begin{subfigure}[t]{0.185\textwidth}\centering\includegraphics[width=\linewidth]{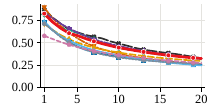}\caption{POD (thr.\ avg.) $\uparrow$}\end{subfigure}

  \vspace{0.3mm}
  \begin{subfigure}[t]{0.185\textwidth}\centering\includegraphics[width=\linewidth]{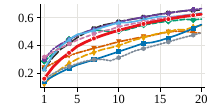}\caption{FAR (thr.\ avg.) $\downarrow$}\end{subfigure}\hspace{0.025\textwidth}
  \begin{subfigure}[t]{0.185\textwidth}\centering\includegraphics[width=\linewidth]{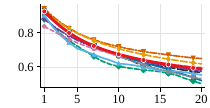}\caption{SSIM $\uparrow$}\end{subfigure}\hspace{0.025\textwidth}
  \begin{subfigure}[t]{0.185\textwidth}\centering\includegraphics[width=\linewidth]{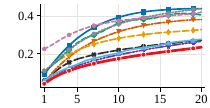}\caption{LPIPS $\downarrow$}\end{subfigure}\hspace{0.025\textwidth}
  \begin{subfigure}[t]{0.185\textwidth}\centering\includegraphics[width=\linewidth]{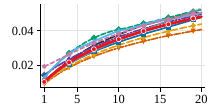}\caption{CRPS $\downarrow$}\end{subfigure}

  \caption{Skill versus forecast frame on SEVIR ($5\to20$ frames, $128\times128$ protocol). The horizontal axis of every panel is the forecast frame. Thresholds are VIL levels ($0$--$255$). FAR is undefined where a method forecasts no pixel above a threshold. Such points are left blank, and the threshold average of FAR skips them. After frame $6$ SwinLSTM forecasts almost no pixel above $219$ (POD $<0.01$), so its FAR at this threshold rests on very few pixels.}
  \label{fig:lt-sevir}
\end{figure}

At VIL $\geq 219$ our CSI falls from $0.549$ at the first frame to $0.035$ at frame $20$, while PDRF, the strongest baseline at both ends, goes from $0.489$ to $0.023$. The relative gain thus rises from $12\%$ to $51\%$. Seven of the nine baselines fall below CSI $0.01$ at this threshold, between frame $7$ (SwinLSTM) and frame $19$ (DiffCast), whereas ours stays above $0.03$. Our POD at this threshold decays from $0.709$ to $0.051$ and is the highest of all methods from frame $14$ on (PercpCast and PDRF lead before, at a higher mean FAR), while our FAR rises from $0.29$ to $0.90$. At VIL $\geq 16$ the best baseline stays $0.02$ to $0.04$ above ours at every frame. Our LPIPS is the lowest at every frame and grows by $0.19$ over the horizon, against $0.33$ for SDIR. SDIR and FlowCast have a higher mean SSIM and a lower mean CRPS.

At VIL $16$ and $74$ SDIR and FlowCast reach a higher CSI than ours, and at VIL $133$ SDIR does ($0.3363$ \vs $0.3341$). The gap is already present in the anchor, since the decoder acts only at the highest tier (Tab.~\ref{tab:abl-branch}). Both methods forecast smoother fields, as their higher SSIM and higher (worse) LPIPS show (Tab.~\ref{tab:radar-main}) and as the forecast examples in Fig.~\ref{fig:qv-sevir} illustrate. A smooth field covers the broad weak-echo area well and loses the heavy cores, where their CSI falls far below ours.

\clearpage
\subsubsection{Visualization}
\label{app:lt-sevir-vis}
\begin{figure}[H]
  \centering
  \begin{subfigure}[t]{0.495\textwidth}\centering\includegraphics[width=\linewidth]{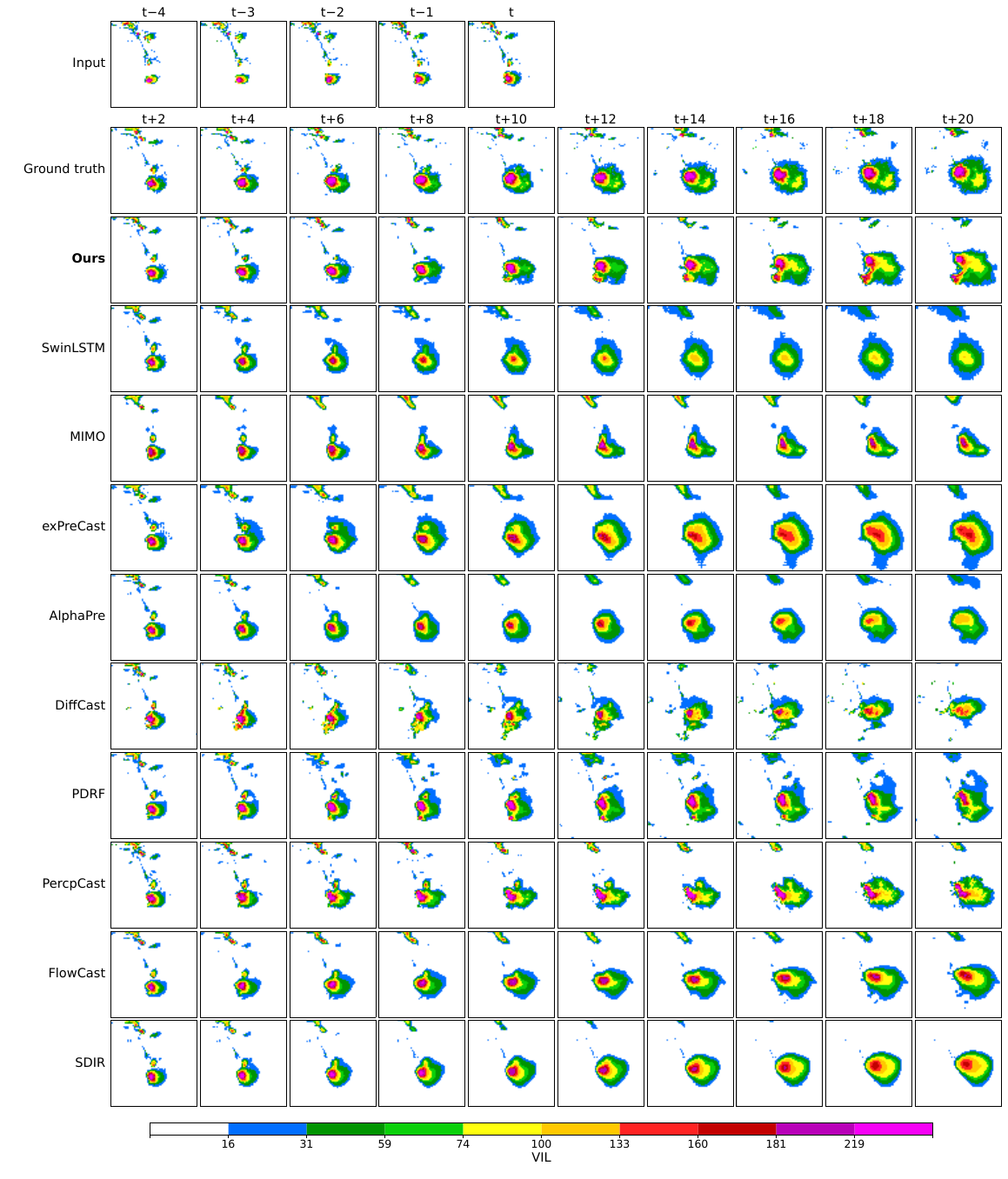}\caption{Case 1}
    \vspace{2pt}\par{\scriptsize\raggedright\textbf{Predicted scenario $\hat{c}$:} \emph{The radar echo complex, anchored in the lower-left quadrant as a bright, irregular core with a sharp leading edge and diffuse trailing tail, surges steadily rightward across the frame with unwavering directionality, propelled by rapid, coherent motion. As it glides horizontally, the entire structure undergoes subtle curvature, its internal gradients and edges reorienting in response to a dominant deformation that twists its form without breaking cohesion. Simultaneously, the system deforms, its boundary sharpness increasing as if drawn by an invisible force, compressing its mass while preserving its elongated silhouette. Though minor ripples flicker along its periphery, the echo resists fragmentation, maintaining structural integrity as it drifts rightward, deforms, and reorients in unison---a dynamically locked entity evolving through synchronized translation, deformation, and contour change, all unfolding smoothly over time without disruption or decay.}\par}
  \end{subfigure}\hfill
  \begin{subfigure}[t]{0.495\textwidth}\centering\includegraphics[width=\linewidth]{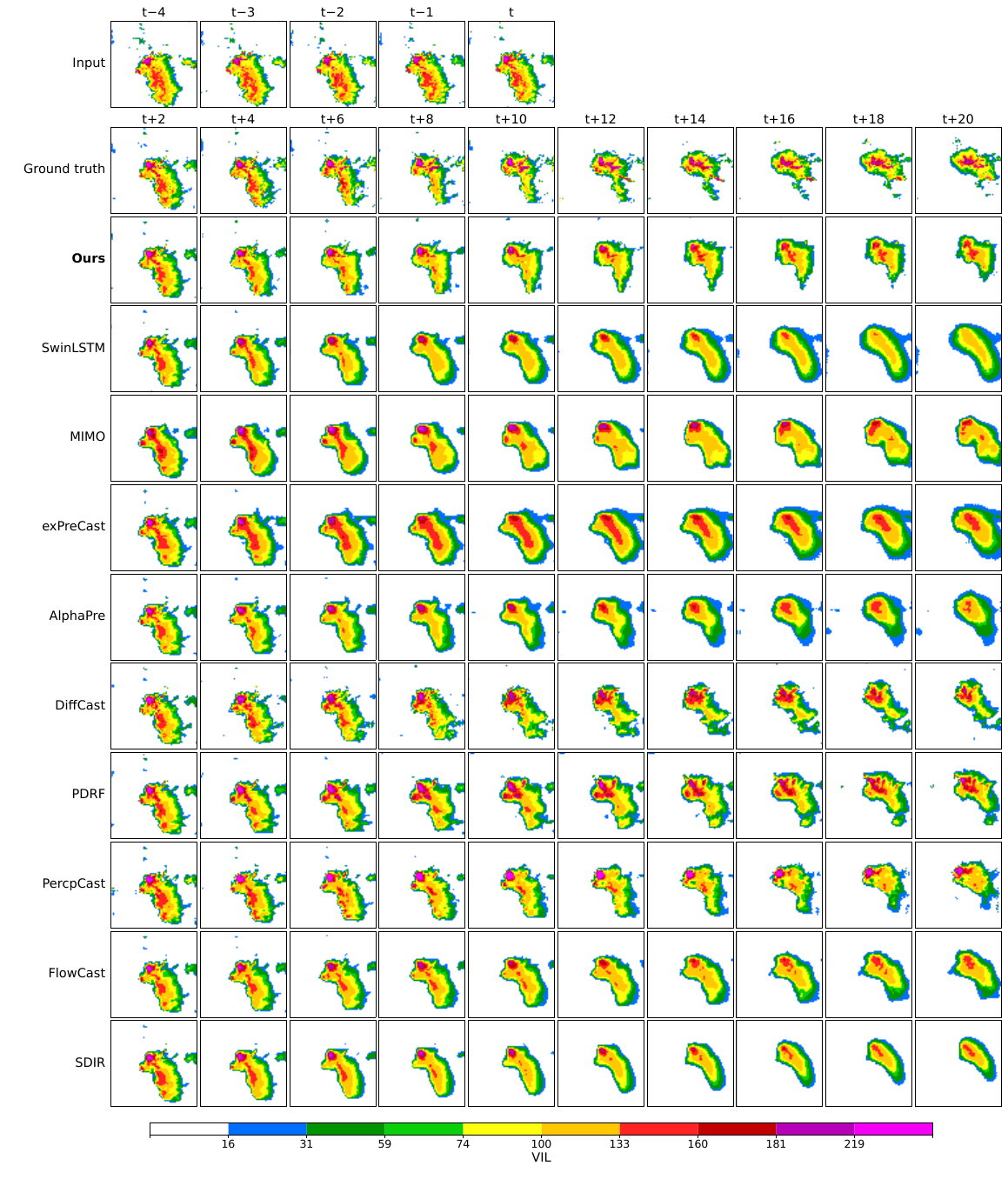}\caption{Case 2}
    \vspace{2pt}\par{\scriptsize\raggedright\textbf{Predicted scenario $\hat{c}$:} \emph{The radar echo drifts steadily rightward and slightly upward across the frame, propelled by a consistent motion vector, while its irregular, blob-like form---soft-edged and internally mottled---slowly deforms, with faint voids and brighter cores shifting within as the entire structure reorients. As it glides, the echo's edges subtly stretch and fray, with internal-core shift and boundary sharpness changes, yet it holds its coherence, never fragmenting or collapsing, maintaining a stable, organized evolution. The deformation gently warps its shape over time, while the reorientation nudges its mass, creating a dynamic but controlled transformation---no sharp breaks, no sudden bursts---just a smooth, persistent drift to the right, with subtle curvature and contour change, all unfolding in harmony as the system evolves without losing its identity.}\par}
  \end{subfigure}
  \caption{Forecast examples on SEVIR. Each panel shows the five input frames, the ground truth and the forecast of every compared method at every second forecast frame. Below each panel is the scenario $\hat{c}$ that the fine-tuned VLM predicts from the input frames alone.}
  \label{fig:qv-sevir}
\end{figure}

Case~1 (Fig.~\ref{fig:qv-sevir}a) is an isolated cell that keeps a core above VIL $219$ over the whole horizon while its envelope above VIL $16$ widens after $t+12$. Our forecast keeps the core above $219$ until $t+14$ and patches above $181$ until $t+20$, so the heavy core survives in the frames where the top-tier CSI of most baselines approaches zero (Fig.~\ref{fig:lt-sevir}g). Two errors are visible. From $t+12$ on ours develops a spurious heavy appendage south-west of the core, and its interior at VIL $74$ to $100$ is larger than observed. SwinLSTM, AlphaPre, FlowCast and SDIR lose the core above $181$ by about $t+8$, leaving a diffuse blob. exPreCast keeps an intense but oversized patch, and DiffCast breaks the cell into fragments between $t+4$ and $t+10$. MIMO-VP, PDRF and PercpCast also keep a core above $219$, with an envelope that is too small for MIMO-VP and irregular for PDRF.

Case~2 is a decaying system that drifts north-east and reorganizes into an east--west band with thin streaks above VIL $181$. Ours keeps a core above $219$ until $t+10$ and pockets of $160$ to $181$ afterwards, inside a compact envelope that keeps the original north-west to south-east orientation. It thus captures the intensity of the streaks better than their arrangement. SwinLSTM, exPreCast and MIMO-VP widen the envelope along the old orientation, AlphaPre, FlowCast and SDIR keep its size with a weaker core, and DiffCast and PDRF reproduce the streaky interior in a fragmented form.

\clearpage
\subsection{MeteoNet}
\label{app:lt-meteo}
\subsubsection{Temporal Analysis}
\label{app:lt-meteo-temporal}

\begin{figure}[H]
  \centering
  \captionsetup[sub]{skip=1pt}
  \includegraphics[width=0.88\textwidth]{img/leadtime_4ds/legend.pdf}\par\vspace{2mm}
  \begin{subfigure}[t]{0.2\textwidth}\centering\includegraphics[width=\linewidth]{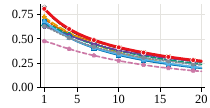}\caption{CSI-M $\uparrow$}\end{subfigure}\hspace{0.025\textwidth}
  \begin{subfigure}[t]{0.2\textwidth}\centering\includegraphics[width=\linewidth]{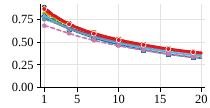}\caption{CSI ($\geq 12$~dBZ) $\uparrow$}\end{subfigure}\hspace{0.025\textwidth}
  \begin{subfigure}[t]{0.2\textwidth}\centering\includegraphics[width=\linewidth]{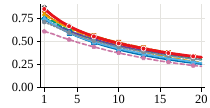}\caption{CSI ($\geq 18$~dBZ) $\uparrow$}\end{subfigure}\hspace{0.025\textwidth}
  \begin{subfigure}[t]{0.2\textwidth}\centering\includegraphics[width=\linewidth]{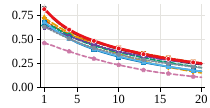}\caption{CSI ($\geq 24$~dBZ) $\uparrow$}\end{subfigure}

  \vspace{1mm}
  \begin{subfigure}[t]{0.2\textwidth}\centering\includegraphics[width=\linewidth]{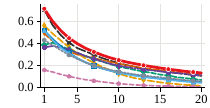}\caption{CSI ($\geq 32$~dBZ) $\uparrow$}\end{subfigure}\hspace{0.025\textwidth}
  \begin{subfigure}[t]{0.2\textwidth}\centering\includegraphics[width=\linewidth]{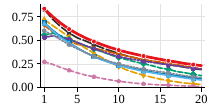}\caption{HSS ($\geq 32$~dBZ) $\uparrow$}\end{subfigure}\hspace{0.025\textwidth}
  \begin{subfigure}[t]{0.2\textwidth}\centering\includegraphics[width=\linewidth]{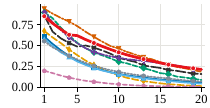}\caption{POD ($\geq 32$~dBZ) $\uparrow$}\end{subfigure}\hspace{0.025\textwidth}
  \begin{subfigure}[t]{0.2\textwidth}\centering\includegraphics[width=\linewidth]{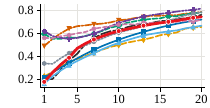}\caption{FAR ($\geq 32$~dBZ) $\downarrow$}\end{subfigure}

  \vspace{1mm}
  \begin{subfigure}[t]{0.2\textwidth}\centering\includegraphics[width=\linewidth]{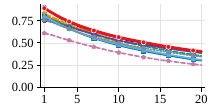}\caption{HSS (thr.\ avg.) $\uparrow$}\end{subfigure}\hspace{0.025\textwidth}
  \begin{subfigure}[t]{0.2\textwidth}\centering\includegraphics[width=\linewidth]{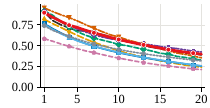}\caption{POD (thr.\ avg.) $\uparrow$}\end{subfigure}\hspace{0.025\textwidth}
  \begin{subfigure}[t]{0.2\textwidth}\centering\includegraphics[width=\linewidth]{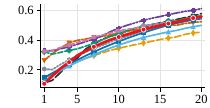}\caption{FAR (thr.\ avg.) $\downarrow$}\end{subfigure}\hspace{0.025\textwidth}
  \begin{subfigure}[t]{0.2\textwidth}\centering\includegraphics[width=\linewidth]{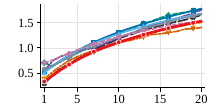}\caption{MAE $\downarrow$}\end{subfigure}

  \vspace{1mm}
  \begin{subfigure}[t]{0.2\textwidth}\centering\includegraphics[width=\linewidth]{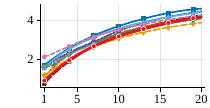}\caption{RMSE $\downarrow$}\end{subfigure}\hspace{0.025\textwidth}
  \begin{subfigure}[t]{0.2\textwidth}\centering\includegraphics[width=\linewidth]{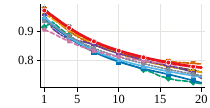}\caption{SSIM $\uparrow$}\end{subfigure}\hspace{0.025\textwidth}
  \begin{subfigure}[t]{0.2\textwidth}\centering\includegraphics[width=\linewidth]{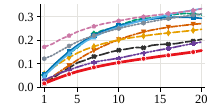}\caption{LPIPS $\downarrow$}\end{subfigure}\hspace{0.025\textwidth}
  \begin{subfigure}[t]{0.2\textwidth}\centering\includegraphics[width=\linewidth]{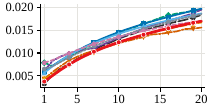}\caption{CRPS $\downarrow$}\end{subfigure}

  \caption{Skill versus forecast frame on MeteoNet ($5\to20$ frames, $128\times128$ protocol). The horizontal axis of every panel is the forecast frame.}
  \label{fig:lt-meteo}
\end{figure}

On MeteoNet our margin at the highest threshold is small in the first frames and grows with lead time. At $32$~dBZ our CSI is $0.717$ against $0.708$ (PDRF) at the first frame and $0.132$ against $0.111$ (SDIR) at frame $20$, so the relative gain grows from $1\%$ to $19\%$. SDIR has a higher POD at this threshold up to frame $15$, together with a clearly higher FAR. At $12$~dBZ our CSI stays within $0.021$ of the best baseline over the whole horizon. Our LPIPS is the lowest from frame $2$ on and shows the smallest increase of all methods ($0.14$). Ours also has the best mean SSIM, while SDIR and FlowCast reach a slightly lower CRPS.

\clearpage
\subsubsection{Visualization}
\label{app:lt-meteo-vis}
\begin{figure}[H]
  \centering
  \begin{subfigure}[t]{0.495\textwidth}\centering\includegraphics[width=\linewidth]{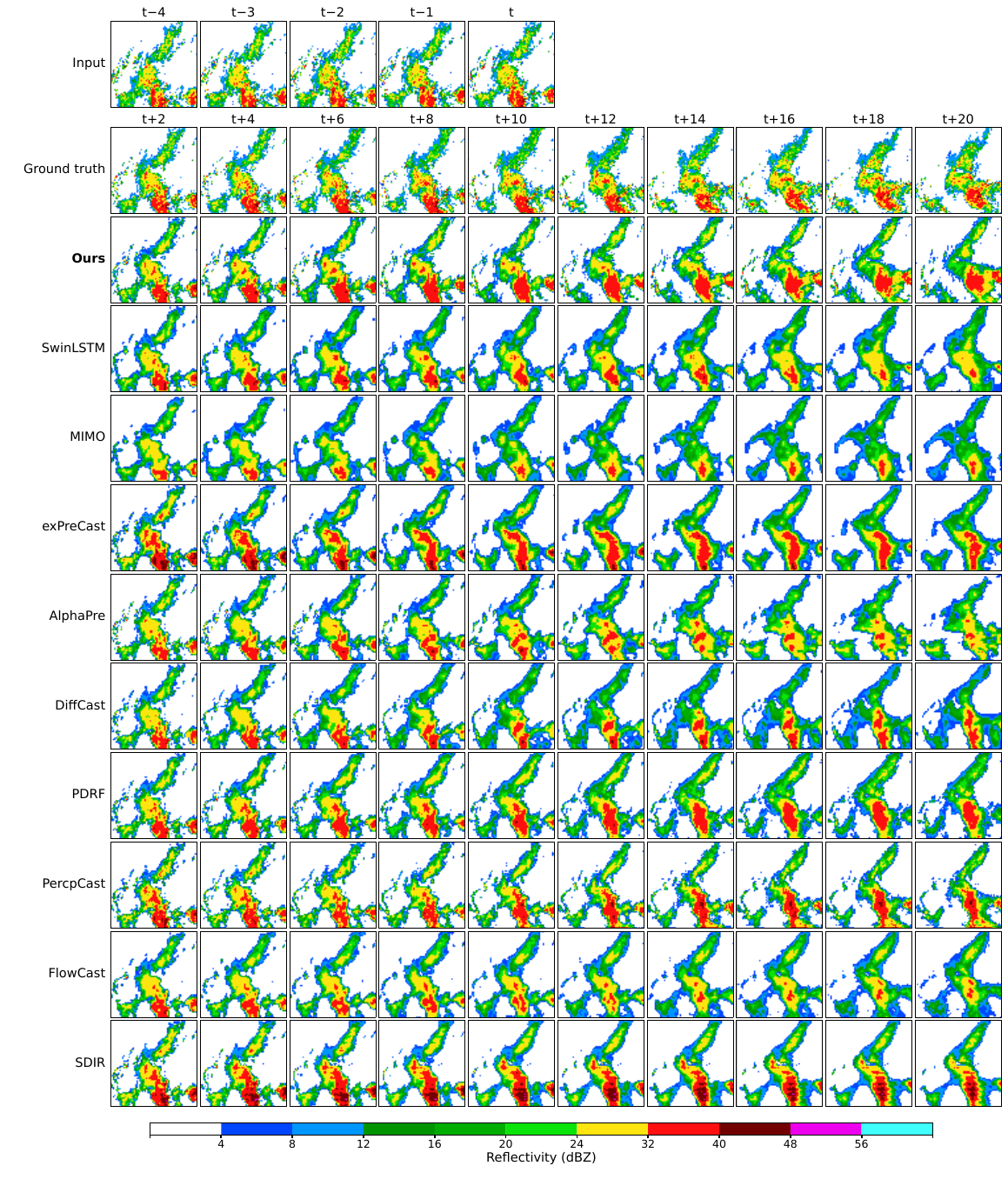}\caption{Case 1}
    \vspace{2pt}\par{\scriptsize\raggedright\textbf{Predicted scenario $\hat{c}$:} \emph{The radar echo structure drifts steadily rightward and slightly upward across the frame, maintaining a coherent, elongated form with a bright, dense core that softens into fragmented, filamentous edges as it extends toward the upper-right. As it moves, the entire system undergoes subtle curvature, its internal features reorienting in unison, while simultaneously deforming---its boundaries becoming less sharp, its mass diffusing without breaking apart. The motion is smooth and persistent, with no abrupt shifts or fragmentation; instead, the echo evolves through gradual deformation, its shape subtly warping as it glides diagonally across the screen. This is a unified, dynamic entity---translating, deforming, and reorienting in harmony---as it sweeps from lower-left to upper-right, leaving no voids or sharp breaks in its wake.}\par}
  \end{subfigure}\hfill
  \begin{subfigure}[t]{0.495\textwidth}\centering\includegraphics[width=\linewidth]{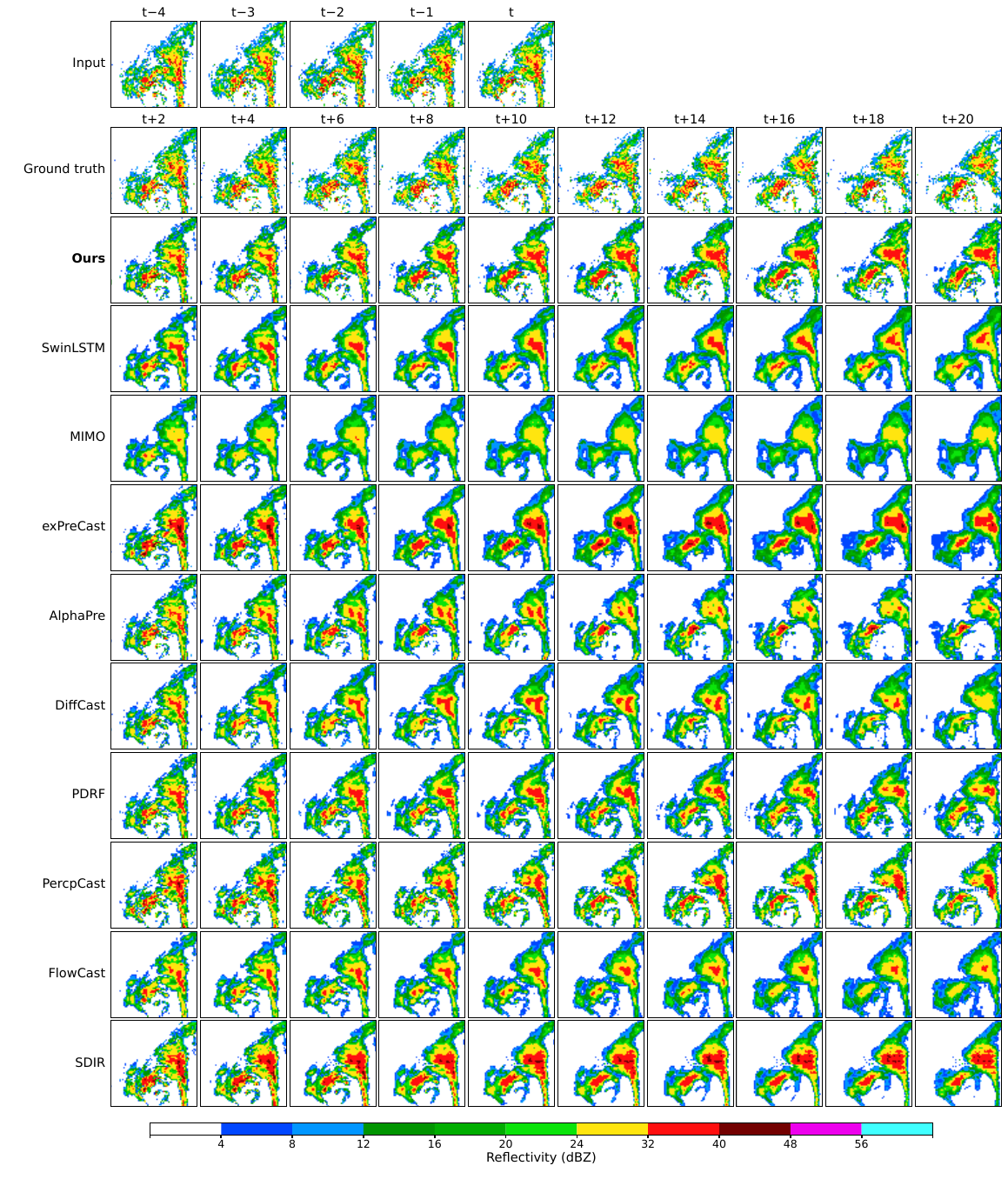}\caption{Case 2}
    \vspace{2pt}\par{\scriptsize\raggedright\textbf{Predicted scenario $\hat{c}$:} \emph{The radar echo structure drifts steadily upward and slightly rightward across the frame, maintaining a coherent, elongated form with a dense core near center-left that gradually tapers toward the upper-right. As it moves, the entire system undergoes subtle curvature, its filamentous arms---especially along the right flank---twisting into curved contours that evolve smoothly without breaking apart. These arms show sharp-edged gradients, particularly on the left where abrupt transitions suggest strong shear, yet the whole structure holds its shape, resisting fragmentation as it glides. Internal voids flicker faintly within the core but do not disrupt the flow, while the motion remains unwavering in direction and speed, propelled by a stable, large-scale force. The deformation is subtle but persistent, curling the edges without distorting the whole, as the system advances diagonally upward-right, its form locked in dynamic equilibrium---translating, reorienting, but never unraveling.}\par}
  \end{subfigure}
  \caption{Forecast examples on MeteoNet. Each panel shows the five input frames, the ground truth and the forecast of every compared method at every second forecast frame. Below each panel is the scenario $\hat{c}$ that the fine-tuned VLM predicts from the input frames alone.}
  \label{fig:qv-meteo}
\end{figure}

Case~1 (Fig.~\ref{fig:qv-meteo}a) is a convective band whose heavy cells above $32$~dBZ are concentrated in its southern part and drift slowly east. Ours keeps a heavy region at the observed location through $t+20$ and preserves fine texture in the weaker part of the band. Its heavy region is a contiguous patch that grows from $t+10$ on, larger and smoother than the observed group of scattered cells, which is consistent with the rise of our FAR at $32$~dBZ to $0.74$ at frame $20$ (Fig.~\ref{fig:lt-meteo}h). SwinLSTM, MIMO-VP and FlowCast keep the band outline with few cells above $32$~dBZ at the end. exPreCast, PDRF and SDIR keep heavy cores, and SDIR forecasts a large area above $40$~dBZ that is almost absent from the observation.

Case~2 has two heavy clusters made of fine-grained cells, one along the north-eastern band and one in the south-west. Ours keeps both clusters through $t+20$ with a textured interior and merges the north-eastern cells into a single patch larger than observed from about $t+12$. MIMO-VP loses all echoes above $32$~dBZ after $t+4$, and AlphaPre breaks both clusters into scattered, noisy fragments. exPreCast and SDIR keep the clusters with cores above $40$~dBZ that are larger than observed. SwinLSTM, DiffCast and FlowCast keep the north-eastern cluster and weaken the south-western one.

\clearpage
\subsection{Shanghai}
\label{app:lt-shanghai}
\subsubsection{Temporal Analysis}
\label{app:lt-shanghai-temporal}

\begin{figure}[H]
  \centering
  \captionsetup[sub]{skip=1pt}
  \includegraphics[width=0.88\textwidth]{img/leadtime_4ds/legend.pdf}\par\vspace{2mm}
  \begin{subfigure}[t]{0.2\textwidth}\centering\includegraphics[width=\linewidth]{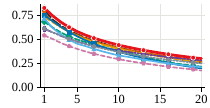}\caption{CSI-M $\uparrow$}\end{subfigure}\hspace{0.025\textwidth}
  \begin{subfigure}[t]{0.2\textwidth}\centering\includegraphics[width=\linewidth]{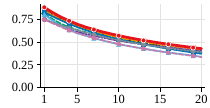}\caption{CSI ($\geq 20$~dBZ) $\uparrow$}\end{subfigure}\hspace{0.025\textwidth}
  \begin{subfigure}[t]{0.2\textwidth}\centering\includegraphics[width=\linewidth]{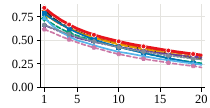}\caption{CSI ($\geq 30$~dBZ) $\uparrow$}\end{subfigure}\hspace{0.025\textwidth}
  \begin{subfigure}[t]{0.2\textwidth}\centering\includegraphics[width=\linewidth]{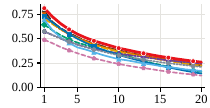}\caption{CSI ($\geq 35$~dBZ) $\uparrow$}\end{subfigure}

  \vspace{1mm}
  \begin{subfigure}[t]{0.2\textwidth}\centering\includegraphics[width=\linewidth]{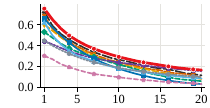}\caption{CSI ($\geq 40$~dBZ) $\uparrow$}\end{subfigure}\hspace{0.025\textwidth}
  \begin{subfigure}[t]{0.2\textwidth}\centering\includegraphics[width=\linewidth]{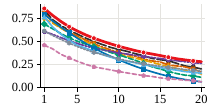}\caption{HSS ($\geq 40$~dBZ) $\uparrow$}\end{subfigure}\hspace{0.025\textwidth}
  \begin{subfigure}[t]{0.2\textwidth}\centering\includegraphics[width=\linewidth]{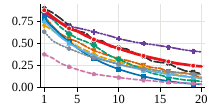}\caption{POD ($\geq 40$~dBZ) $\uparrow$}\end{subfigure}\hspace{0.025\textwidth}
  \begin{subfigure}[t]{0.2\textwidth}\centering\includegraphics[width=\linewidth]{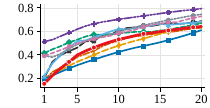}\caption{FAR ($\geq 40$~dBZ) $\downarrow$}\end{subfigure}

  \vspace{1mm}
  \begin{subfigure}[t]{0.2\textwidth}\centering\includegraphics[width=\linewidth]{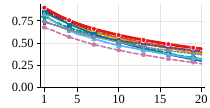}\caption{HSS (thr.\ avg.) $\uparrow$}\end{subfigure}\hspace{0.025\textwidth}
  \begin{subfigure}[t]{0.2\textwidth}\centering\includegraphics[width=\linewidth]{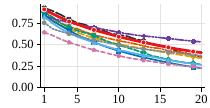}\caption{POD (thr.\ avg.) $\uparrow$}\end{subfigure}\hspace{0.025\textwidth}
  \begin{subfigure}[t]{0.2\textwidth}\centering\includegraphics[width=\linewidth]{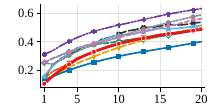}\caption{FAR (thr.\ avg.) $\downarrow$}\end{subfigure}\hspace{0.025\textwidth}
  \begin{subfigure}[t]{0.2\textwidth}\centering\includegraphics[width=\linewidth]{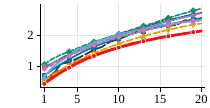}\caption{MAE $\downarrow$}\end{subfigure}

  \vspace{1mm}
  \begin{subfigure}[t]{0.2\textwidth}\centering\includegraphics[width=\linewidth]{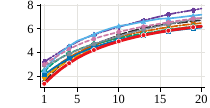}\caption{RMSE $\downarrow$}\end{subfigure}\hspace{0.025\textwidth}
  \begin{subfigure}[t]{0.2\textwidth}\centering\includegraphics[width=\linewidth]{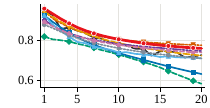}\caption{SSIM $\uparrow$}\end{subfigure}\hspace{0.025\textwidth}
  \begin{subfigure}[t]{0.2\textwidth}\centering\includegraphics[width=\linewidth]{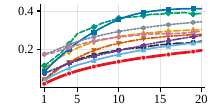}\caption{LPIPS $\downarrow$}\end{subfigure}\hspace{0.025\textwidth}
  \begin{subfigure}[t]{0.2\textwidth}\centering\includegraphics[width=\linewidth]{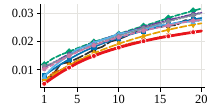}\caption{CRPS $\downarrow$}\end{subfigure}

  \caption{Skill versus forecast frame on Shanghai ($5\to20$ frames, $128\times128$ protocol). The horizontal axis of every panel is the forecast frame.}
  \label{fig:lt-shanghai}
\end{figure}

On Shanghai the margin at the highest threshold is small at both ends of the horizon. At $40$~dBZ our CSI goes from $0.756$ to $0.167$, against $0.728$ (PDRF) and $0.158$ (PercpCast) for the strongest baseline at the two ends, a relative gain of $4\%$ at the first frame and $6\%$ at the last. PDRF and then PercpCast have a higher POD at this threshold, at a higher mean FAR, and our FAR is below the baseline range in the first two frames and within it afterwards. At $20$~dBZ ours leads the best baseline at every frame, by at most $0.018$. Shanghai is the only benchmark where ours has the best mean value of every continuous metric shown, namely SSIM, LPIPS, CRPS, MAE and RMSE.

\clearpage
\subsubsection{Visualization}
\label{app:lt-shanghai-vis}
\begin{figure}[H]
  \centering
  \begin{subfigure}[t]{0.495\textwidth}\centering\includegraphics[width=\linewidth]{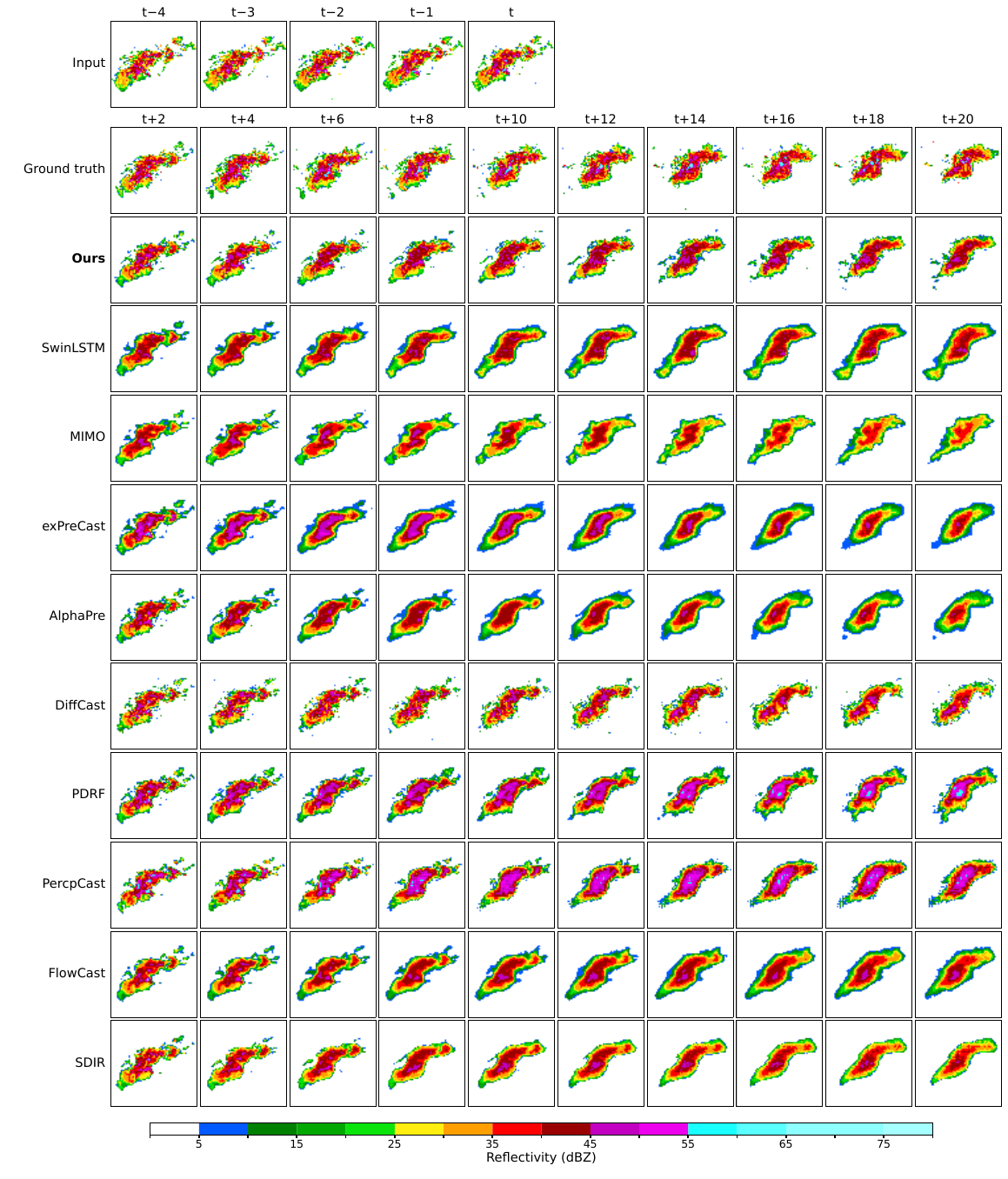}\caption{Case 1}
    \vspace{2pt}\par{\scriptsize\raggedright\textbf{Predicted scenario $\hat{c}$:} \emph{The radar echo structure drifts steadily rightward with a subtle upward tilt, tracing a near-straight trajectory across the frame as it maintains its elongated, filamentous form---its brightest core gliding smoothly while fainter, fragmented edges trail and ripple without breaking apart. As it advances, the entire band undergoes subtle curvature, its internal gradients shifting in unison like a slow deformation, preserving cohesion despite minor undulations along its length. No fragmentation or merging occurs; instead, the system flows as a unified entity, its motion dominated by advection rather than internal change, gliding with remarkable directional consistency---each frame revealing only incremental displacement, no sudden shifts or structural ruptures, just a persistent, coherent slide toward the upper-right, propelled by stable, large-scale forces that leave its shape intact from start to finish.}\par}
  \end{subfigure}\hfill
  \begin{subfigure}[t]{0.495\textwidth}\centering\includegraphics[width=\linewidth]{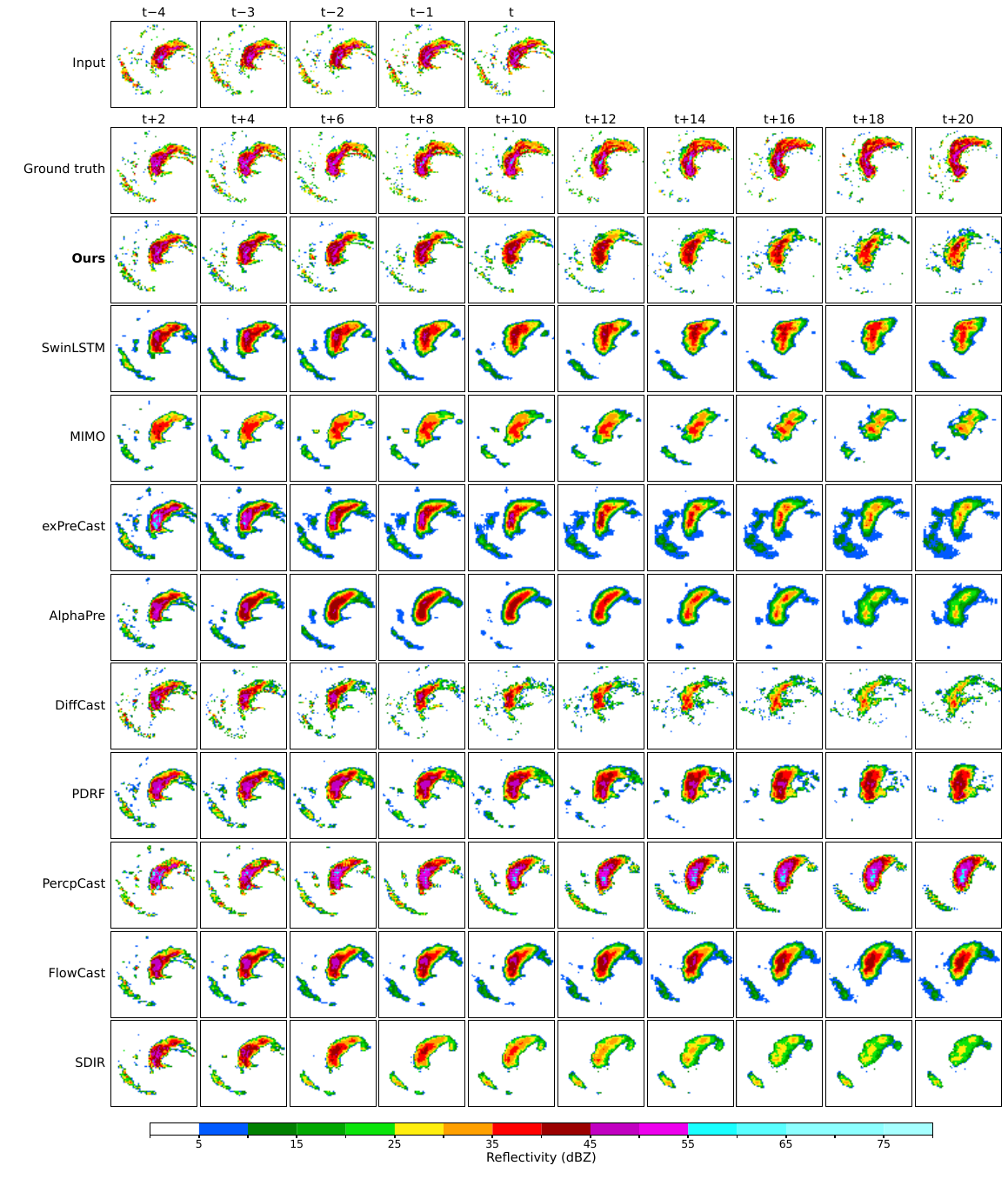}\caption{Case 2}
    \vspace{2pt}\par{\scriptsize\raggedright\textbf{Predicted scenario $\hat{c}$:} \emph{The radar echo structure drifts steadily leftward and slightly downward across the frame, maintaining a smooth, unwavering trajectory as it moves with consistent speed. Its dominant arc-shaped band---brightest along the leading edge---curves from upper right to lower left, with the curvature subtly reorienting as it moves, preserving its form without fragmentation. The core remains dense and coherent, while the trailing edge softens and diffuses, yet no voids or internal disruptions emerge; the entire system holds its shape with remarkable stability. As it translates, the arc's curvature gently reorients in place, like a rigid band pivoting mid-motion, reinforcing its organized, mesoscale nature. No deformation or fragmentation occurs---just pure, fluid advection: a fast-sliding, reorienting crescent of energy, locked in motion, gliding left-downward with quiet, relentless momentum.}\par}
  \end{subfigure}
  \caption{Forecast examples on Shanghai. Each panel shows the five input frames, the ground truth and the forecast of every compared method at every second forecast frame. Below each panel is the scenario $\hat{c}$ that the fine-tuned VLM predicts from the input frames alone.}
  \label{fig:qv-shanghai}
\end{figure}

Case~1 (Fig.~\ref{fig:qv-shanghai}a) is an elongated line whose interior stays between $35$ and $45$~dBZ, with scattered cores above $45$~dBZ and a narrow rim. Ours keeps the elongated outline, the narrow rim and the textured interior with scattered cores above $45$~dBZ through $t+20$, in line with its low LPIPS (Fig.~\ref{fig:lt-shanghai}o). It misses the few pixels above $55$~dBZ that appear between $t+10$ and $t+18$. SwinLSTM, exPreCast, AlphaPre, FlowCast and SDIR forecast a smooth band with a wide weak rim, and MIMO-VP weakens the interior below $45$~dBZ. PDRF and PercpCast fill most of the interior above $45$~dBZ and partly above $55$~dBZ, which overestimates the heavy area. DiffCast keeps the fine texture but breaks the line into fragments.

Case~2 is a bow-shaped echo whose core above $45$~dBZ persists and widens to the north over the horizon. Ours keeps the curvature over all frames and the core above $45$~dBZ until about $t+10$, after which the core falls to $40$ to $45$~dBZ and the outline frays. SDIR loses most of the intensity by $t+10$ and AlphaPre by $t+18$. SwinLSTM and FlowCast keep a heavy core inside a shrunken or bloated bow, and PDRF turns the bow into a compact blob. PercpCast keeps both the curvature and a core above $45$~dBZ until $t+20$, with an interior above $55$~dBZ that is more intense than observed.

\clearpage
\subsection{CIKM}
\label{app:lt-cikm}
\subsubsection{Temporal Analysis}
\label{app:lt-cikm-temporal}

\begin{figure}[H]
  \centering
  \captionsetup[sub]{skip=1pt}
  \includegraphics[width=0.88\textwidth]{img/leadtime_4ds/legend.pdf}\par\vspace{2mm}
  \begin{subfigure}[t]{0.2\textwidth}\centering\includegraphics[width=\linewidth]{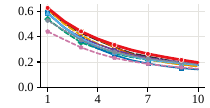}\caption{CSI-M $\uparrow$}\end{subfigure}\hspace{0.025\textwidth}
  \begin{subfigure}[t]{0.2\textwidth}\centering\includegraphics[width=\linewidth]{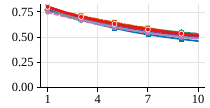}\caption{CSI ($\geq 20$~dBZ) $\uparrow$}\end{subfigure}\hspace{0.025\textwidth}
  \begin{subfigure}[t]{0.2\textwidth}\centering\includegraphics[width=\linewidth]{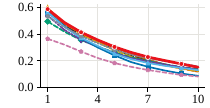}\caption{CSI ($\geq 30$~dBZ) $\uparrow$}\end{subfigure}\hspace{0.025\textwidth}
  \begin{subfigure}[t]{0.2\textwidth}\centering\includegraphics[width=\linewidth]{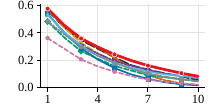}\caption{CSI ($\geq 35$~dBZ) $\uparrow$}\end{subfigure}

  \vspace{1mm}
  \begin{subfigure}[t]{0.2\textwidth}\centering\includegraphics[width=\linewidth]{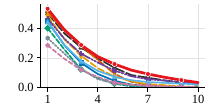}\caption{CSI ($\geq 40$~dBZ) $\uparrow$}\end{subfigure}\hspace{0.025\textwidth}
  \begin{subfigure}[t]{0.2\textwidth}\centering\includegraphics[width=\linewidth]{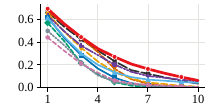}\caption{HSS ($\geq 40$~dBZ) $\uparrow$}\end{subfigure}\hspace{0.025\textwidth}
  \begin{subfigure}[t]{0.2\textwidth}\centering\includegraphics[width=\linewidth]{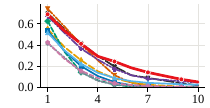}\caption{POD ($\geq 40$~dBZ) $\uparrow$}\end{subfigure}\hspace{0.025\textwidth}
  \begin{subfigure}[t]{0.2\textwidth}\centering\includegraphics[width=\linewidth]{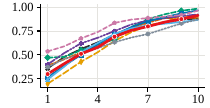}\caption{FAR ($\geq 40$~dBZ) $\downarrow$}\end{subfigure}

  \vspace{1mm}
  \begin{subfigure}[t]{0.2\textwidth}\centering\includegraphics[width=\linewidth]{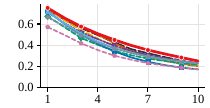}\caption{HSS (thr.\ avg.) $\uparrow$}\end{subfigure}\hspace{0.025\textwidth}
  \begin{subfigure}[t]{0.2\textwidth}\centering\includegraphics[width=\linewidth]{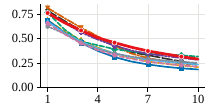}\caption{POD (thr.\ avg.) $\uparrow$}\end{subfigure}\hspace{0.025\textwidth}
  \begin{subfigure}[t]{0.2\textwidth}\centering\includegraphics[width=\linewidth]{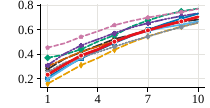}\caption{FAR (thr.\ avg.) $\downarrow$}\end{subfigure}\hspace{0.025\textwidth}
  \begin{subfigure}[t]{0.2\textwidth}\centering\includegraphics[width=\linewidth]{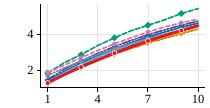}\caption{MAE $\downarrow$}\end{subfigure}

  \vspace{1mm}
  \begin{subfigure}[t]{0.2\textwidth}\centering\includegraphics[width=\linewidth]{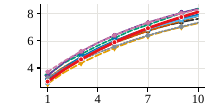}\caption{RMSE $\downarrow$}\end{subfigure}\hspace{0.025\textwidth}
  \begin{subfigure}[t]{0.2\textwidth}\centering\includegraphics[width=\linewidth]{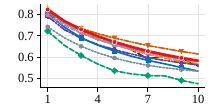}\caption{SSIM $\uparrow$}\end{subfigure}\hspace{0.025\textwidth}
  \begin{subfigure}[t]{0.2\textwidth}\centering\includegraphics[width=\linewidth]{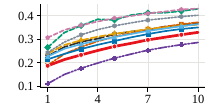}\caption{LPIPS $\downarrow$}\end{subfigure}\hspace{0.025\textwidth}
  \begin{subfigure}[t]{0.2\textwidth}\centering\includegraphics[width=\linewidth]{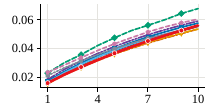}\caption{CRPS $\downarrow$}\end{subfigure}

  \caption{Skill versus forecast frame on CIKM ($5\to10$ frames, $128\times128$ protocol). The horizontal axis of every panel is the forecast frame.}
  \label{fig:lt-cikm}
\end{figure}

CIKM forecasts ten frames. At $40$~dBZ our CSI goes from $0.5331$ to $0.0337$, against $0.5050$ (SDIR) and $0.0294$ (PercpCast) for the strongest baseline at the two ends, and six of the nine baselines fall below $0.01$ by frame $9$. The relative gain grows from $6\%$ to $15\%$, and our POD at this threshold is the highest from frame $5$ on. At $20$~dBZ ours trails the best baseline at every frame, by $0.002$ at the first frame and $0.017$ at the end. CIKM is also the benchmark where the cost of sharp forecasts is most visible. PercpCast has the lowest LPIPS at every frame, and our MAE, RMSE and CRPS grow faster than those of FlowCast and SDIR, with RMSE rising by $5.14$ over the ten frames against $4.47$ and $4.58$. This matches the double penalty that pixel errors assign to sharp echoes once their position drifts.

\clearpage
\subsubsection{Visualization}
\label{app:lt-cikm-vis}
\begin{figure}[H]
  \centering
  \begin{subfigure}[t]{0.495\textwidth}\centering\includegraphics[width=\linewidth]{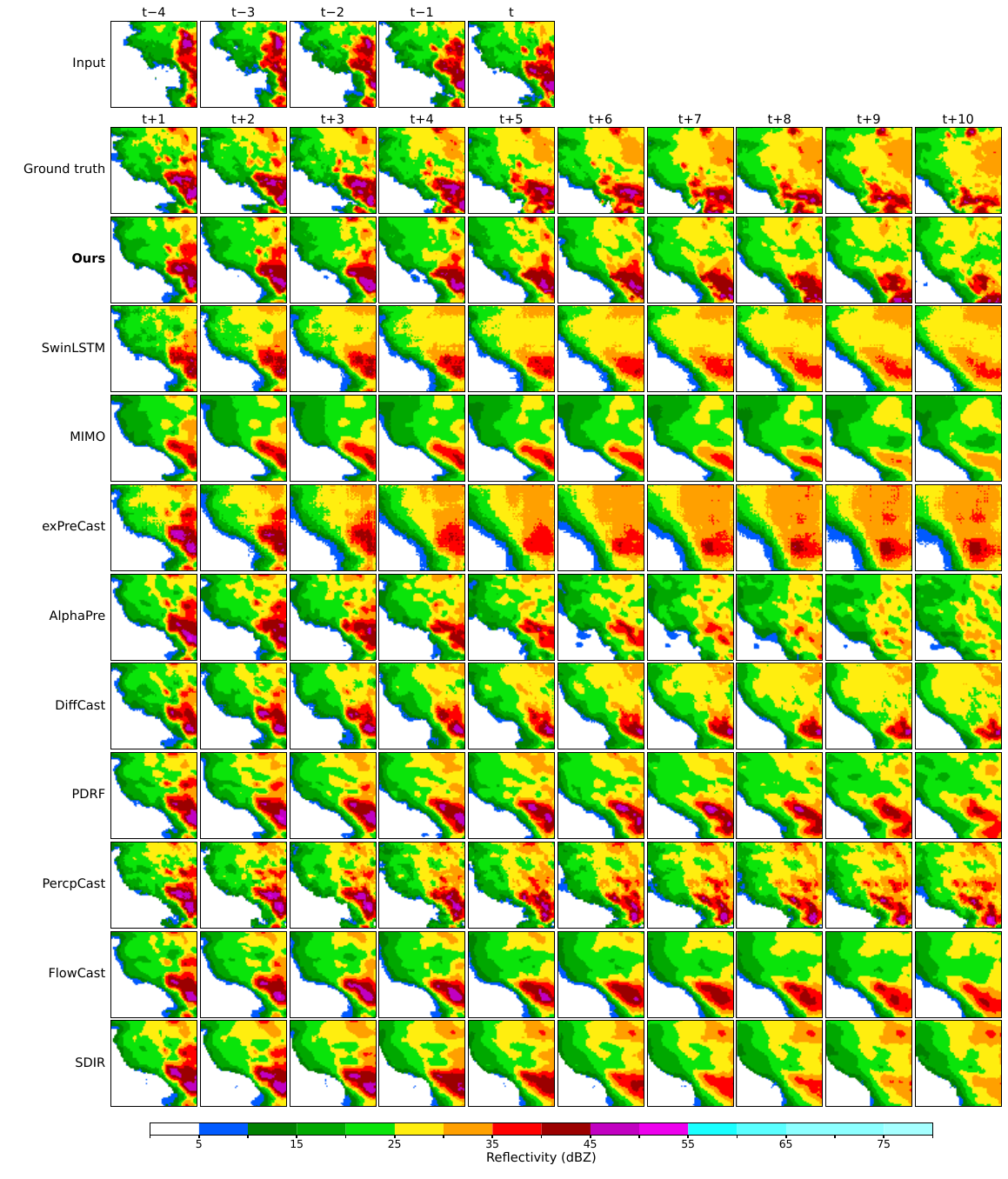}\caption{Case 1}
    \vspace{2pt}\par{\scriptsize\raggedright\textbf{Predicted scenario $\hat{c}$:} \emph{The echo complex initiates motion in the upper-right quadrant, rapidly translating rightward with a slight downward drift along a consistent trajectory, resulting in a noticeable displacement over time. As it advances, the elongated, irregular band undergoes deformation, with its leading edge sharpening while the trailing side diffuses, and internal bright cores shift amid dark voids, indicating dynamic reorganization. A distinct curvature emerges, particularly in the central region, where contour changes develop, reinforcing a coherent mesoscale structure. Throughout the sequence, the system maintains structural integrity, with no fragmentation or dissipation, as deformation and reorientation interact to sustain a stable, organized evolution. The overall motion combines rapid translation, persistent curvature, and gradual contour change, tracing a continuous path of coordinated movement and transformation across the frame.}\par}
  \end{subfigure}\hfill
  \begin{subfigure}[t]{0.495\textwidth}\centering\includegraphics[width=\linewidth]{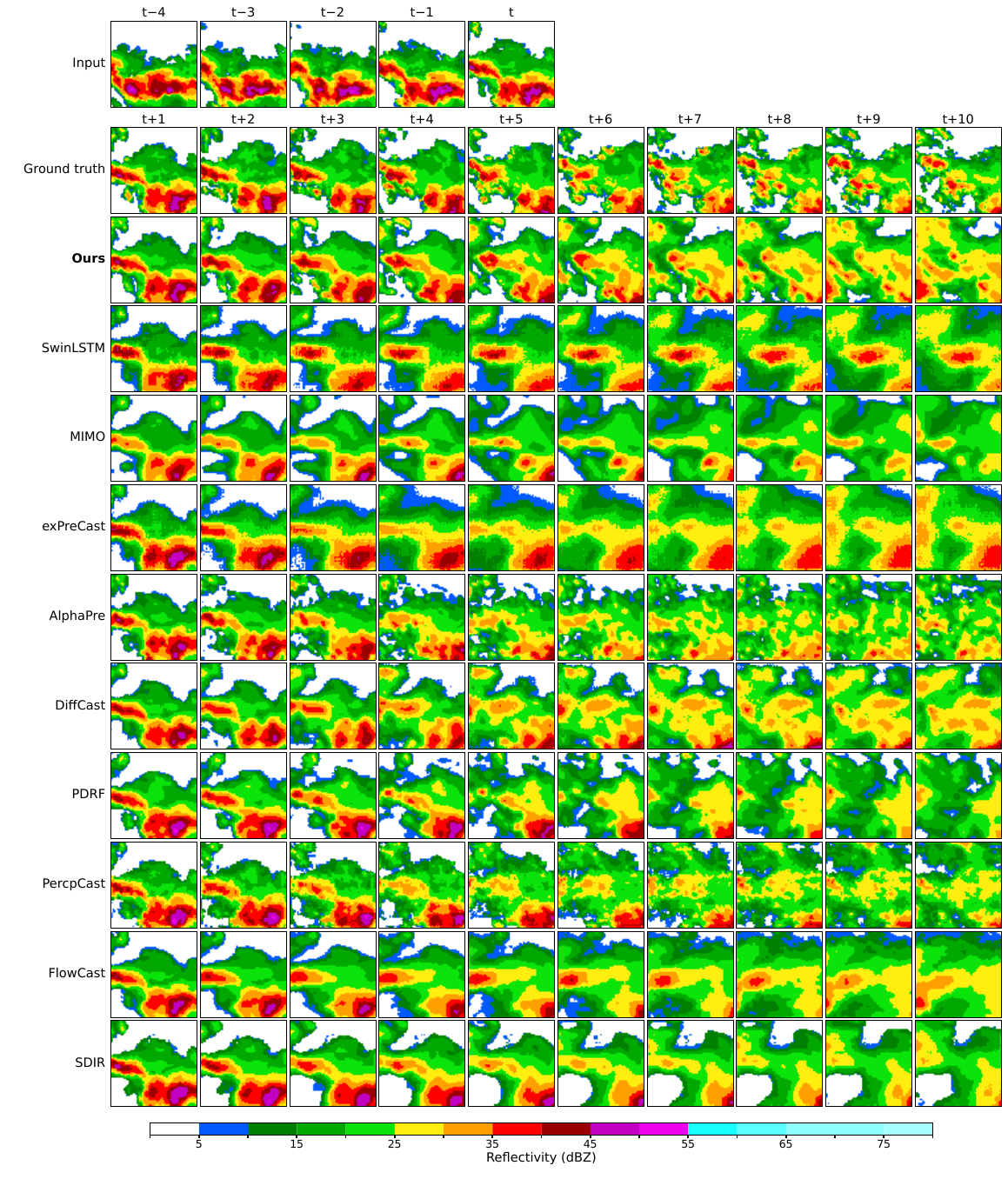}\caption{Case 2}
    \vspace{2pt}\par{\scriptsize\raggedright\textbf{Predicted scenario $\hat{c}$:} \emph{The radar echo structure drifts steadily rightward with a subtle downward tilt, tracing a smooth trajectory across the frame at a steady pace, while its dense core---anchored in the lower-left---pulls fragmented, wispy tendrils behind it as they stretch and dissolve toward the upper-right. As it glides, the entire mass undergoes subtle curvature, its leading edge curling like a slow contour change, while internal voids shift and reconfigure without breaking cohesion. The outer fringes continuously fray and re-knit, yet the system holds its form, resisting fragmentation as it slides unswervingly right-downward, its motion fluid and persistent, propelled by a balance of deformation and contour change that keeps its shape intact even as its edges ripple and reform in response to unseen forces.}\par}
  \end{subfigure}
  \caption{Forecast examples on CIKM. Each panel shows the five input frames, the ground truth and the forecast of every compared method at every forecast frame. Below each panel is the scenario $\hat{c}$ that the fine-tuned VLM predicts from the input frames alone.}
  \label{fig:qv-cikm}
\end{figure}

Case~1 (Fig.~\ref{fig:qv-cikm}a) is a band of echoes above $40$~dBZ that drifts toward the bottom edge and broadens eastward, under a widening area of $30$ to $35$~dBZ in the upper right. Ours keeps a core above $40$~dBZ near the observed position through $t+10$. The core stays more compact than the observed band, and the upper-right area is forecast at $25$ to $30$~dBZ, weaker than observed. SwinLSTM and exPreCast over-smooth the echoes into a broad $30$ to $40$~dBZ region, and exPreCast also shows block artifacts. AlphaPre and SDIR lose the core by $t+7$. DiffCast and PDRF keep a heavy core of similar quality to ours, FlowCast keeps an elongated and displaced core, and PercpCast keeps the core with a noisy texture.

Case~2 starts with a heavy band on the left and a core above $45$~dBZ at the bottom. After $t+4$ the observation breaks up into small scattered cells of $35$ to $45$~dBZ over the whole domain while the bottom core decays. No method reproduces this rapid break-up. Ours keeps the left band until $t+5$ and then forecasts a broad field of $25$ to $35$~dBZ with few heavy cells, which overestimates the area of moderate echoes. This case illustrates the growth of our pixel errors on CIKM (Fig.~\ref{fig:lt-cikm}l,~m). AlphaPre and PercpCast produce a speckled field closer in texture to the observation, with cells at the wrong positions, and exPreCast keeps the bottom core long after it has decayed.

\clearpage
\section{Implementation Details}
\label{app:impl}

Table~\ref{tab:impl} lists the settings of the generator, the autoencoder, the generator optimization and the student VLM. The $5.2\times$ cost in the main text counts network evaluations ($52$ against $10$). Measured on one RTX 5090, the wall-clock ratio is $4.4\times$, as the eight branches after the shared trunk are evaluated together in a single batch at each step.

\begin{table}[htbp]
  \centering
  \small
  \caption{Implementation details.}
  \label{tab:impl}
  \renewcommand{\arraystretch}{1.12}
  \begin{tabular}{ll}
    \hline
    \multicolumn{2}{l}{\emph{Generator}} \\
    Backbone & DiT, $28$ layers, width $1152$, $16$ heads, patch $(1,2,2)$ over (time, height, width) \\
    Parameters & $583.93$M ($435.29$M visual backbone $+$ $148.64$M text cross-attention) \\
    Frames (condition $+$ forecast) & Swedish precipitation / MRMS $4+16$, SEVIR / MeteoNet / Shanghai $5+20$, CIKM $5+10$ \\
    Tokens per sample & Swedish precipitation / MRMS $5120$, SEVIR / MeteoNet / Shanghai $6400$, CIKM $3840$ \\
    Objective & Rectified Flow \\
    Sampler & Euler, $T=10$ steps \\
    Average inference time & $2.0$~s single forecast, $8.8$~s $8$-member ensemble (one RTX 5090, batch size $1$) \\
    \hline
    \multicolumn{2}{l}{\emph{Autoencoder}} \\
    Architecture & FLUX.2 autoencoder~\cite{flux2}, base channels $128$, multipliers $[1,2,4,4]$, $2$ residual blocks per level \\
    Parameters & $84.05$M (encoder $34.43$M $+$ decoder $49.62$M) \\
    Latent per frame & $32\times32\times32$ for a $256\times256$ frame ($8\times$ spatial downsampling, $2\times$ fewer values) \\
    Input / output & range $[-1,1]$, first of the $3$ output channels kept \\
    \hline
    \multicolumn{2}{l}{\emph{Generator optimization}} \\
    Optimizer & AdamW, $\beta=(0.9,0.999)$, weight decay $0.01$, $\epsilon=10^{-8}$ \\
    Learning rate & $1\times10^{-5}$ \\
    Schedule & constant learning rate after $200$ warm-up steps \\
    Regularization and precision & EMA $0.9999$, gradient clipping $1.0$, bf16, gradient checkpointing \\
    Batch size & $32$--$36$ \\
    Hardware & $6\times$ NVIDIA RTX 4090 \\
    \hline
    \multicolumn{2}{l}{\emph{Student VLM}} \\
    Model & Qwen3.5-9B~\cite{qwen35}, base weights frozen, condition frames $h$ as input, maximum length $1024$ tokens \\
    LoRA & rank $8$, $\alpha=32$, dropout $0.05$, all linear layers of the language model, vision encoder and merger \\
    Optimizer & AdamW, $\beta=(0.9,0.95)$, weight decay $0.1$, $\epsilon=10^{-8}$ \\
    Learning rate & $2\times10^{-5}$, cosine schedule after $100$ warm-up steps \\
    Training & $2$ epochs, batch size $12$, bf16 \\
    \hline
  \end{tabular}
\end{table}